\documentclass[acmsmall]{acmart} 
 
\AtBeginDocument{%
  \providecommand\BibTeX{{%
    \normalfont B\kern-0.5em{\scshape i\kern-0.25em b}\kern-0.8em\TeX}}}

\usepackage{graphicx}
\usepackage[shortlabels]{enumitem}
\usepackage{multirow}
\usepackage{wrapfig}
\usepackage{hhline}
\usepackage{booktabs}
\usepackage{subcaption}
\usepackage{cancel}

\usepackage[english]{babel} % To obtain English text with the blindtext package
\usepackage{blindtext}

\usepackage{makecell}
\usepackage{bbm}
\usepackage{hyperref}
\hypersetup{
  colorlinks,
  citecolor=violet,
  linkcolor=red,
  urlcolor=blue}

\usepackage{url}
\def\cL{{\mathcal{L}}}

\usepackage[utf8]{inputenc}

\usepackage{amsmath}
\usepackage{algorithm}
\usepackage[noend]{algpseudocode}
\usepackage{algorithmicx}

\begin{document}

\title{Taxonomy of Machine Learning Safety: A Survey and Primer}
% \title{Practical Machine Learning Safety: A Survey and Primer}
% \title{Practical Solutions in Machine Learning Safety: A Primer and Survey}

\author{Sina Mohseni}
\email{smohseni@nvidia.com}
\author{Zhiding Yu}
\email{zhidingy@nvidia.com}
\author{Chaowei Xiao}
\email{chaoweix@nvidia.com}
\author{Jay Yadawa}
\email{jyadawa@nvidia.com}
\affiliation{%
  \institution{NVIDIA}
  \city{Santa Clara}
  \state{CA}
  \country{USA}
  \postcode{95051}
}

\author{Haotao Wang}
\email{htwang@utexas.edu}
\author{Zhangyang Wang}
\email{atlaswang@utexas.edu>}
\affiliation{
  \institution{The University of Texas at Austin}
  \city{Austin}
  \state{TX}
  \country{USA}
  \postcode{78712}
}

\renewcommand{\shortauthors}{Mohseni, et al.}
%
%
%% The abstract is a short summary of the work to be presented in the article.
%
%
\begin{abstract}
The open-world deployment of Machine Learning (ML) algorithms in safety-critical applications such as autonomous vehicles needs to address a variety of ML vulnerabilities such as interpretability, verifiability, and performance limitations. Research explores different approaches to improve ML dependability by proposing new models and training techniques to reduce generalization error, achieve domain adaptation, and detect outlier examples and adversarial attacks. 
However, there is a missing connection between ongoing ML research and well-established safety principles.
In this paper, we present a structured and comprehensive review of ML techniques to improve the dependability of ML algorithms in uncontrolled open-world settings. 
From this review, we propose the \textit{Taxonomy of  ML Safety} that maps state-of-the-art ML techniques to key engineering safety strategies. 
Our taxonomy of ML safety presents a safety-oriented categorization of ML techniques to provide guidance for improving dependability of the ML design and development. 
The proposed taxonomy can serve as a safety checklist to aid designers in improving coverage and diversity of safety strategies employed in any given ML system.
\end{abstract}




\maketitle

\section{Introduction}

Advancements in machine learning (ML) have been one of the most significant innovations of the last decade. 
Among different ML models, Deep Neural Networks (DNNs)~\cite{lecun2015deep} are well-known and widely used for their powerful representation learning from high-dimensional data such as images, texts, and speech. 
However, as ML algorithms enter sensitive real-world domains with trustworthiness, safety, and fairness prerequisites, the need for corresponding techniques and metrics for high-stake domains is more noticeable than before. 
Hence, researchers in different fields propose guidelines for \textit{Trustworthy AI}~\cite{shneiderman2020bridging}, \textit{Safe AI} \cite{amodei2016concrete}, and \textit{Explainable AI}~\cite{mohseni2018multidisciplinary} as stepping stones for next generation Responsible AI \cite{arrieta2020explainable}. % wiens2019no
Furthermore, government reports and regulations on AI accountability~\cite{goodman2017european}, trustworthiness \cite{smuha2019eu}, and safety \cite{cluzeau2020concepts} are gradually creating mandating laws to protect citizens' data privacy rights, fair data processing, and upholding safety for AI-based products.

The development and deployment of ML algorithms for open-world tasks come with reliability and dependability challenges rooting from model performance, robustness, and uncertainty limitations \cite{mohseni2019practical}.  % mcallister2017concrete 
Unlike traditional code-based software, ML models have fundamental safety drawbacks, including performance limitations on their training set and run-time robustness constraints in their operational domain.  
For example, ML models are fragile to unprecedented domain shift \cite{ganin2014unsupervised} that could easily occur in open-world scenarios. 
Data corruptions and natural perturbations~\cite{hendrycks2019benchmarking} are other factors affecting ML models. 
Moreover, from the security perspective, it has been shown that DNNs are susceptible to adversarial attacks that make small perturbations to the input sample (indistinguishable by the human eye) but can fool a DNN~\cite{goodfellow2014explaining}. 
Due to the lack of verification techniques for DNNs, validation of ML models is often bounded to performance measures on standardized test sets and end-to-end simulations on the operation design domain. 
Realizing that dependable ML models are required to achieve safety, we observe the need to investigate gaps and opportunities between conventional engineering safety standards and a set of ML safety-related techniques.

\begin{figure*}
\vspace{-0.5em}
    \centering
    \includegraphics[width=0.99\columnwidth]{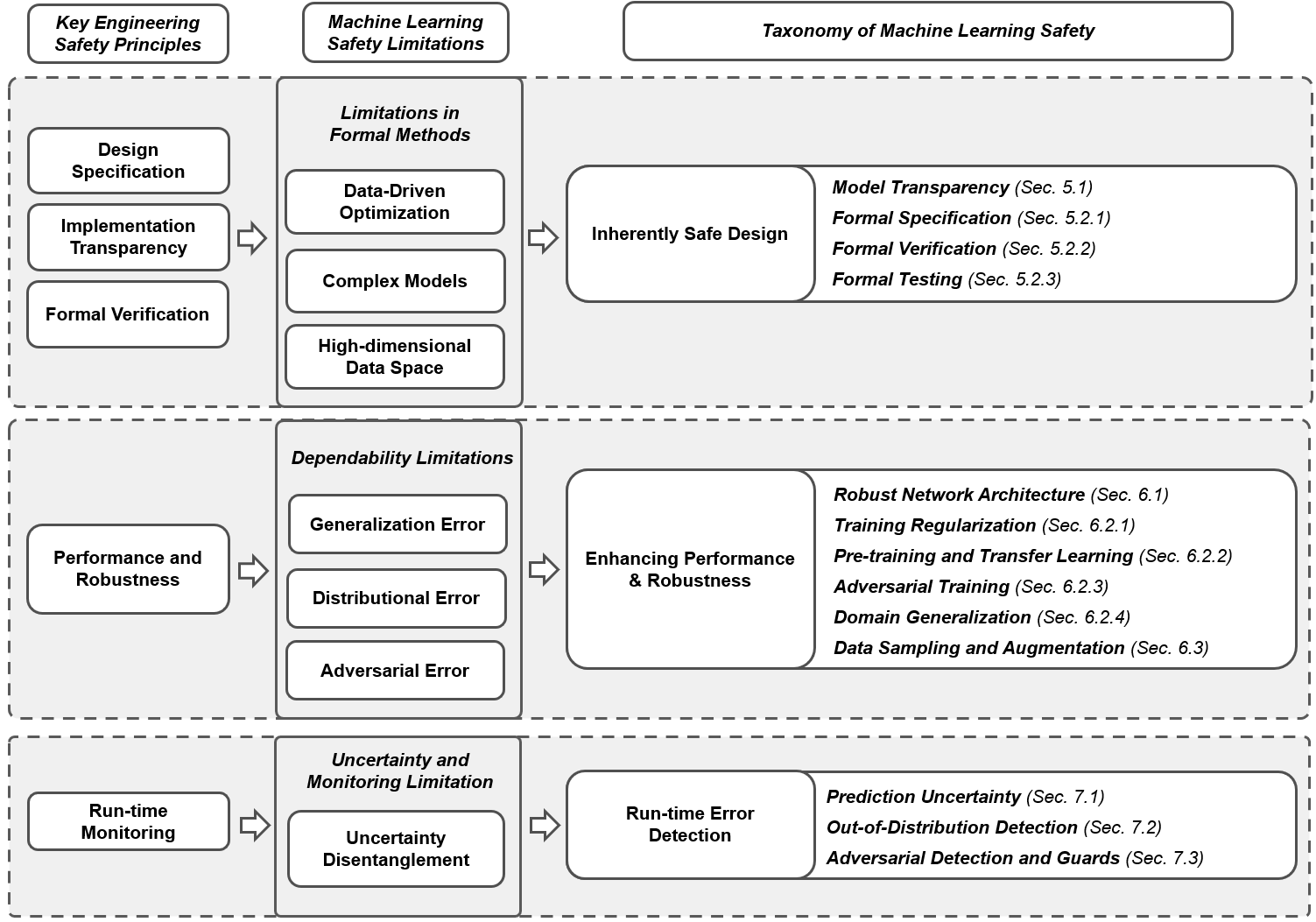}
    \vspace{-0.6em}
    \caption{Paper Roadmap: we first identify key engineering safety requirements (first column) that are limited or not readily applicable on complex ML algorithms (second column). From there, we present a review of safety-related ML research followed by their categorization (third column) into three strategies to achieve \textit{(1) Inherently Safe Models}, improving \textit{(2) Enhancing Model Performance and Robustness}, and incorporate \textit{(3) Run-time Error Detection} techniques.} % Note that safety strategies and their subsequent techniques complement and interact with each other to improve the reliability of any ML-based system.
    \label{fig:roadmap}
    \vspace{-1em}
\end{figure*}

\vspace{-0.5em}
\subsection{Scope, Organization, and Survey Method}
%%% scope 
ML safety includes diverse hardware and software techniques for the safe execution of algorithms in open-world applications \cite{koopman2017autonomous}. 
In this paper, we limit our scope to only ML algorithm design and not the execution of those algorithms in platforms. 
With that being said, we also mainly focus on ``in situ'' techniques to improve run-time dependability and not on further techniques for the efficiency of the network or training. 

%%% Survey Method
We used a structured and iterative methodology to find ML safety-related papers and categorize these research as summarized in Table \ref{tab:ref-table}. 
In our iterative paper selection process, we started with reviewing key research papers from AI and ML safety (e.g., \cite{amodei2016concrete,leike2017ai,varshney2016engineering}) and software safety literature and standards (e.g., \cite{ISO-26262,ISO/PAS-21448,salay2017analysis}) to identify mutual safety attributes between engineering safety and ML techniques. 
Next, we conducted an upward and downward literature investigation using top computer science conference proceedings, journal publications, and the \textit{Google Scholar} search engine to maintain reasonable literature coverage and balance the number of papers on each ML safety attribute.
% In order to do so, we selected papers across different fields of artificial intelligence, including machine learning and deep learning, reinforcement learning, and other related fields like software safety and Human-AI interactions. 

%%% organization
Figure \ref{fig:roadmap} presents the overall organization of this paper. 
We first review the background on common safety terminologies and situate ML safety limitations with reference to conventional engineering safety requirements in Section \ref{sec:background}. % for a better understand of the problem space. % software engineering
In Section \ref{sec:error-types} we discuss a unified ``big picture'' of different ML error types for real-world applications and common benchmark datasets to evaluate models for these errors. 
Next, we propose a ML safety taxonomy in Section \ref{sec:categorization} to organize ML techniques into safety strategies with Table \ref{tab:ref-table} as an illustration of the taxonomy with the summary of representative papers on each subcategory. 
% with Table \ref{tab:ref-table} as a hierarchical summary of reviewed papers. 
Sections \ref{sec:safe-design}, \ref{sec:robustness-performance}, and \ref{sec:error-detection} construct the main body of the reviewed papers organized into ML solutions and techniques for each safety strategy. 
Finally, Section \ref{sec:discussion} presents a summary of key takeaways and a discussion of open problems and research directions for ML safety.

\vspace{-0.5em}
\subsection{Objectives and Contributions}
In this paper, we review challenges and opportunities to achieve ML safety for open-world safety-critical applications. 
We first review dependability limitations and challenges for ML algorithms in comparison to engineering safety standard requirements. 
Then, we decompose ML dependability needs into three safety strategies for (1) achieving inherently safe ML design, (2) improving model performance and robustness, and (3) building run-time error detection solutions for ML. 
Following our categorization of safety strategies, we present a structured and comprehensive review of 300 papers from a broad spectrum of state-of-the-art ML research and safety literature. % and propose a ML safety Taxonomy with a representative reference table (Table \ref{tab:ref-table}). 
We propose a unifying taxonomy (Table \ref{tab:ref-table}) that serves ML researchers and designers as a collection of best practices and allows to checkup the coverage and diversity of safety strategies employed in any given ML system. 
Additionally, the taxonomy of ML safety lays down a road map to safety needs in ML and accommodates in assessing technology readiness for each safety strategy. % , and helps to identify needed future research directions. 
We review open challenges and opportunities for each strategy and present a summary of key takeaways in the end.

% ... decomposing engineering safety strategies into practical solution
% ... by avoiding common pitfalls
% ... can serve as a checklist for testers... 
% ... introduce new design approaches
% .... solutions to diversify safety strategies
% ... Our taxonomy allows to checkup the coverage and diversity of safety strategies employed in any given ML system. 
% .... from different aspects, and discuss research gaps as well as promising solutions.

\vspace{-0.5em}
\section{Background}
\label{sec:background}

In order to introduce and categorize ML safety techniques, we start with reviewing background on engineering safety strategies and investigate safety gaps between design and development of code-based software and ML algorithms. 
% In the following, we first decompose and clarify common terminologies in this space and emphasize the differences between \textit{Algorithmic Dependability} and \textit{Software Safety} for ML-based systems. 
% Next, we review safety limitations for ML algorithms w.r.t. key engineering safety principles that do not stand for the design and implementation of ML-based software. 
% Note that we identify and introduce model dependability limitations as a subgroup of safety limitations. 

\vspace{-0.3em}
\subsection{Related Surveys} 
\label{sec:terminology}

Related survey papers dive into ML and AI safety topics to analyze problem domain, review existing solutions and make suggestions on future directions \cite{amodei2016concrete,leike2017ai}.
Survey papers cover diverse topics including safety-relevant characteristics in reinforcement learning~\cite{hernandez2019surveying}, verification of ML components~\cite{huang2017safety,wang2018efficient}, adversarial robustness \cite{huang2020survey}, anomaly detection \cite{salehi2021unified}, ML uncertainty \cite{mcallister2017concrete}, and aim to connect the relation between well-established engineering safety principals and ML safety \cite{varshney2016engineering,mohseni2019practical} limitations.
\citet{hendrycks2021unsolved} introduce four major research problems to improve ML safety namely robustness, monitoring, alignment, and external safety for ML models.
\citet{shneiderman2020bridging} presents high level guidelines for teams, organizations, and industries to increase the reliability, safety, and trustworthiness of next generation Human-Centered AI systems.

More recently, multiple surveys present holistic review of ML promises and pitfalls for safety-critical autonomous systems.
For instance, \citet{ashmore2021assuring} demonstrate a systematic presentation of 4-stage ML lifecycle, including data management, model training, model verification, and deployment. 
Authors present itemized safety assurance requirements for each stage and review methods that support each requirement.
In a later work, \citet{hawkins2021guidance} add ML safety assurance scoping and the safety requirements elicitation stages to ML lifecycle to establish the fundamental link between system-level hazard and risk analysis and unit-level safety requirements. 
In a broader context, \citet{lu2021software} study challenges and limitations of existing ML system development tools and platforms (MLOps) in achieving \textit{Responsible AI} principles such as data privacy, transparency, and safety. % fairness, and human-centered values. 
Authors report their findings with a list of operationalized Responsible AI principles and their benefits and drawbacks.

Although prior work targeted different aspects and characteristics of ML safety and dependability, in this paper, we elaborate on  ML safety concept by situating open-world safety challenges with ongoing ML research.
Particularly, we combine ML safety concerns between engineering and research communities to uncover mutual goals and accelerate safety developments.
% we benefit from the body of research in prior work to 

\vspace{-0.3em}
\subsection{Related Terminologies} 
We introduce terminologies related to ML safety by clarifying the relationship between ML \textit{Safety}, \textit{Security}, and \textit{Dependability} that are often interchangeably used in the literature. 
% --- safety ---
\textit{Safety} is a \textit{System-Level} concept as a set of processes and strategies to minimize the risk of hazards due to malfunctioning of system components.
Safety standards such as IEC 61508 \cite{IEC-61508} and ISO 26262 \cite{ISO-26262} mandate complete analysis of hazards and risks, documentation for system architecture and design, detailed development process, and thorough verification strategies for each component, integration of components, and final system-level testing. 
% \vspace{-0.5em}
% --- dependability --- 
% \paragraph{Dependability} 
\textit{Dependability} is a \textit{Unit-Level} concept to ensure performance and robustness of the software in its operational domain. 
We define ML dependability as the model's ability to minimize test-time prediction error. % which is a software unit failure. 
Therefore, a highly dependable ML algorithm is expected to be robust to natural distribution shifts within their intended operation design domain. 
% \vspace{-0.5em}
% --- security ---
% \paragraph{Security}
\textit{Security} is both a \textit{System-Level} and a \textit{Unit-Level} concept to protected from harm or other non-desirable (e.g., data theft, privacy violation) outcomes caused by adversaries. 
Note that engineering guidelines distinguish safety hazards (e.g., due to natural perturbations) from security hazards (e.g., due to adversarial perturbations) as the latter intentionally exploits system vulnerabilities to cause harm. % \cite{carlini2017towards}
However, the term safety is often loosely used in ML literature to refer to the dependability of algorithms against adversaries \cite{huang2020survey}.

In this paper, we focus on unit-level strategies to maintain the dependability of ML algorithms in an intelligent system rather than the safety of a complex AI-based system as a whole.
We also cover adversarial training and detection techniques as a part of unit-level safety strategies regardless of the role of the adversary in generating the attack.

\vspace{-0.3em}
\subsection{Engineering Safety Limitations in ML} 
Engineering safety broadly refers to the management of operations and events in a system in order to protect its users by minimizing hazards, risks, and accidents.
Given the importance of dependability of the system's internal components (hardware and software), various engineering safety standards have been developed to ensure the system's functional safety based on two fundamental principles of safety life cycle and failure analysis. 
Built on collection of best practices, engineering safety processes discover and eliminate design errors followed by a probabilistic analysis of safety impact of possible system failures (i.e., failure analysis). 
Several efforts attempted to extend engineering safety standards to ML algorithms \cite{salay2017analysis,cluzeau2020concepts}.
For example, European Union Aviation Safety Agency released a report on concepts of design assurance for neural networks~\cite{cluzeau2020concepts} that introduces safety assurance and assessment for learning algorithms in safety-critical applications. 
% Their report focuses on using model generalization bounds for performance and safety assessment and proposes a development cycle for learning assurance. 
In another work, Siebert et al.~\cite{siebert2020towards} present a guideline to assess ML system quality from different aspects specific to ML algorithms including data, model, environment, system, and infrastructure in an industrial use case. % similar to software standards. They review evaluation objectives and metrics for
However, the main body of engineering standards do not account for the statistical nature of ML algorithms and errors occurring due to the inability of the components to comprehend the environment. 
In a recent review of automotive functional safety for ML-based software, Salay et al.~\cite{salay2017analysis} present an analysis that shows about 40\% of software safety methods do not apply to ML models. 
% of ISO-26262 part-6 methods with respect to the safety of ML models. 
% Their assessment of software safety methods' applicability on ML algorithms (as software unit design) 

Given the dependability limitations of ML algorithms and lack of adaptability for traditional software development standards, we identify 5 open safety challenges for ML and briefly review active research topics for closing these safety gaps in the following. % presented in Figure \ref{fig:limitations}.
We extensively review the techniques for each challenge later in Sections \ref{sec:safe-design}, \ref{sec:robustness-performance}, and \ref{sec:error-detection}.

\vspace{-0.3em}
\subsubsection{Design Specification}
Documenting and reviewing the software specification is a crucial step in engineering safety; however, formal design specification of ML models is generally not feasible, as the models learn patterns from large training sets to discriminate (or generate) their distributions for new unseen input. 
Therefore, ML algorithms learn the target classes through their training data (and regularization constraints) rather than formal specification. 
The lack of specifiability could cause a mismatch between ``designer objectives'' and ``what the model actually learned'', which could result in unintended functionality of the system. 
The data-driven optimization of model variables in ML training makes it challenging to define and pose specific safety constraints. 
Seshia et al.~\cite{seshia2018formal} surveyed the landscape of formal specification for DNNs to lay an initial foundation for formalizing and reasoning about properties of DNNs. 
To fill this gap, a common practice is to achieve partial design specification through training data specification and coverage.
Another practical way to overcome the design specification problem is to break ML components into smaller algorithms (with smaller tasks) to work in a hierarchical structure.
In the case of intelligent agents, safety-enforcing regularization terms \cite{amodei2016concrete}, and simulation environments \cite{brockman2016openai} are suggested to specify and verify training goals for the agent. %  have been proposed to specify and verify safety goals.

\vspace{-0.3em}
\subsubsection{Implementation Transparency}
Implementation transparency is an important requirement in engineering safety which gives the ability to trace back design requirements from the implementations. 
However, advanced ML models trained on high-dimensional data are not transparent. 
The very large number of variables in the models makes them incomprehensible or a so-called black-box for design review and inspection. 
In order to achieve traceability, significant research has been performed on interpretability methods for DNN to provide instance explanations of model prediction and DNN intermediate feature layers~\cite{zeiler2014visualizing}.
In autonomous vehicles application, \citet{bojarski2018visualbackprop} propose VisualBackProp technique and show that a DNN algorithm trained to control a steering wheel would in fact learn patterns of lanes, road edges, and parked vehicles to execute the targeted task. 
However, the completeness of interpretability methods to grant traceability is not proven yet~\cite{adebayo2018sanity}, and in practice, interpretability techniques are mainly used by designers to improve network structure and training process rather than support a safety assessment.

\vspace{-0.3em}
\subsubsection{Testing and Verification.}
Design and implementation verification is another demanding requirement for unit testing to meet engineering safety standards. 
For example, coding guidelines for software safety enforce the elimination of dead or unreachable functions.
Depending upon the safety integrity level, complete statement, branch coverage, or modified condition and decision coverage are required to confirm the adequacy of the unit tests. 
Coming to DNNs, formally verifying their correctness is challenging and in fact provably an NP-hard~\cite{seshia2016towards} problem due to the high dimensionality of the data. 
Therefore, reaching complete testing and verification of the operational design domain is not feasible for domains like image and video. 
As a result, researchers proposed new techniques such as searching for unknown-unknowns~\cite{bansal2018coverage} and predictor-verifier training~\cite{dvijotham2018training}, and simulation-based toolkits \cite{dreossi2019verifai} guided by formal models and specifications. 
Other techniques, including neuron coverage and fuzz testing \cite{wang2018efficient} in neural networks incorporate these aspects. 
Note that formal verification of shallow and linear models for low dimensional sensor data does not carry verification challenges of the image domain. 

\vspace{-0.3em}
\subsubsection{Performance and Robustness}
Engineering safety standards treat the ML models as a black box and suggest using methods to improve model performance and robustness.
However, improving model performance and robustness is still an open problem and a vast research topic. 
Unlike code-based algorithms, statistical learning algorithms typically contain a residual error rate (due to false positive and false negative predictions) on the test set. 
In addition to the error rate on the test set,  
\textit{operational error} is referred to as the model's error rate that commonly occurs in open-world deployment.
Section \ref{sec:robustness-performance} reviews various approaches like introducing larger networks, training regularization, active learning and data collection, and domain generalization techniques to increase the model's ability to learn generalizable representations for open-world applications.
% We will review more details and ML techniques to improve model robustness in Section \label{sec:robustness-performance}.

\vspace{-0.3em}
\subsubsection{Run-time Monitoring} 
Engineering safety standards suggest run-time monitoring functions as preventive solutions for various system errors, including less frequent transient errors.
Monitoring functions in code-based algorithms are based on a rule-set to detect hardware errors and software crashes in the target operational domain. 
However, designing monitoring functions to predict ML error (e.g., false positive and false negative errors) is different in nature. 
ML models generate prediction probability that could be used to predict uncertainty for run-time validation of predictions. 
However, research shows that prediction probability in complex models like DNN does not fully represent uncertainty and hence can not guarantee failure prediction~\cite{hein2019relu}.
Section \ref{sec:error-detection} reviews different approaches for run-time uncertainty estimation and detection of outlier samples and adversarial attacks.
% \input{sections/2.1-extended-background}

%% 1- Empirical Error:      generalization error on i.i.d test set
%% 2- Distributional Error: generalization error on reasonable shifts 
%% 3- Adversarial Error:    generalization error on adv. perturbed sample

\vspace{-0.5em}
\section{ML Dependability}
\label{sec:error-types}

We define ML dependability as the model's ability to minimize prediction risk on a given test set. 
Unlike code-based algorithms, the dependability of ML algorithms is bounded to the model's learning capacity and statistical assumptions such as independent and identically distribution (i.i.d) relation of source and target domains. 
However, maintaining data distribution assumptions when deployed in the open-world is challenging and results in different types of prediction errors.

In this section, we decompose model dependability limitations into three prediction error types: (i) Generalization Error, (ii) Distributional Error, and (iii) Adversarial Error as a unified ``big picture'' for dependable and robust ML models in open-world. 
Additionally, we review benchmark datasets commonly used for evaluating model dependability.

\vspace{-0.5em}
\subsection{Generalization Error} 
\label{sec:generalization_error}

The first and foremost goal of machine learning is to minimize the \emph{Generalization Error}.
Given a hypothesis $h$ (e.g., a model with learned parameters), the generalization error (also know as the \emph{true error} and denoted as $R(h)$) is defined as the expected error of $h$ on the data distribution $\mathcal{D}$~\cite{mohri2018foundations}:
$
    R(h) = \text{Pr}_{(x,y) \sim \mathcal{D}}[h(x)\neq y] = \mathbb{E}_{(x,y) \sim \mathcal{D}}[\mathbbm{1}_{h(x)\neq y}],
$
where $(x,y)$ is a pair of data and labels sampled from $\mathcal{D}$, and $\mathbbm{1}$ is the indicator function. 
However, the generalization error is not directly computable since $\mathcal{D}$ is usually unknown. 
The \textit {de facto} practical solution is to learn $h$ by empirical risk minimization (ERM) on the training set $\mathbb{S}_{S}=\{(x_i,y_i)\}_{i=1}^{N_S}$ and then estimate its generalization error by the \emph{empirical error} on the holdout test set $\mathbb{S}_{T}=\{(x_i,y_i)\}_{i=1}^{N_T}$. 
Formally, the empirical error $\hat{R}(h)$ is defined as the mean error on a finite set of data points $\mathbb{S} \sim \mathcal{D}^{m}$~\cite{mohri2018foundations}:
$
    \hat{R}(h) = \frac{1}{m} \sum_{i=1}^{m} \mathbbm{1}_{h(x_i)\neq y_i},
$
where $\mathbb{S} \sim \mathcal{D}^{m}$ means $\mathbb{S}=\{(x_i,y_i)\}_{i=1}^m \overset{i.i.d.}{\sim} \mathcal{D}$.
The training and test sets are all sampled from the same distribution $\mathcal{D}$ but are disjoint.

Recent years have witnessed the successful application of this holdout evaluation methodology to monitoring the progress of many ML fields, especially where large-scale labeled datasets are available. 
The generalization error can be affected by many factors, such as training set quality (e.g., imbalanced class distribution \cite{haixiang2017learning}, noisy labels \cite{bouguelia2018agreeing}), model learning capacity, and training method (e.g., using pre-training \cite{mahajan2018exploring} or regularization \cite{bansal2018can}).
% On the other hand, the training set quality can significantly affect model generalization error as many assumptions such as uniform distribution of training data may not be easy to achieve in open and uncontrolled environments. 
% This is because of the large number of model parameters in the network that may exceed the number of data points in training sets, resulting in relatively small generalization error on their test set, but a high error rate on slight domain shifts within the same label space (i.e., hyper-parameter over-fitting). 
% However, representation learning gets more complicated when aiming for more complex tasks and using data with higher dimensionality. 

\vspace{-0.5em}
\paragraph{Benchmark Datasets:} 
Model generalization is commonly evaluated on a separate i.i.d test set provided for the dataset. 
However, recent research has found limitations of this evaluation strategy. 
For example, \citet{wang2020going} showed that the fixed ImageNet \cite{deng2009imagenet}  test set is not sufficient to reliably evaluate the generalization ability of state-of-the-art image classifiers due to the insufficiency in representing the rich visual open-world.
% Other efforts have been made to collect new samples to improve test sets for image classification tasks. % as complements for the original ImageNet test set.
In another work, \citet{tsipras2020imagenet} observed that a noisy data collection pipeline could lead to a systematic misalignment between the training sets and the real-world tasks. 
% a brittleness that puts claims about human-level performance into context. 

\vspace{-0.6em}
\subsection{Distributional Error} 
\label{sec:distributional_shift}
We define \emph{Distributional Error} as the increase in model generalization error when the i.i.d assumption between the source training set $\mathbb{S}_S$ and target test set $\mathbb{S}_T$ is violated. 
% When measuring generalization error, we assume the source training set $\mathbb{S}_S$ and the target test set $\mathbb{S}_T$ are from the same data distribution. 
Eliminating distributional error is particularly important for real-world applications because the i.i.d assumption is frequently violated in uncontrolled settings. 
In other words, we will have $p_S(x,y) \neq p_T(x,y)$ where $p_S(x,y)$ and $p_T(x,y)$ are the joint probability density distributions of data $x$ and label $y$ on the training and test distributions, respectively.
Such mismatch between training and test data distribution is known as \textit{Distributional Shift} (also termed as \textit{Dataset Shift} \cite{quinonero2009dataset} or \textit{Domain Shift}).
In the following, we review the three most common roots of distribution shifts and their benchmark datasets. 

\vspace{-0.6em}
\paragraph{Covariate Shift}
refers to a change in the test distribution of the input covariate $x$ compared to training distribution so that $p_S(x) \neq p_T(x)$ while the labeling function remains the same $p_S(y|x) = p_T(y|x)$. 
Covariate shift may occur due to natural perturbations (e.g., weather and lighting changes), data changes over time (e.g., seasonal variations of data), and even more subtle digital corruptions on images (e.g., JPEG compression and low saturation). 

\vspace{-0.6em}
\paragraph{Label Distribution Shift} 
is the scenario when the marginal distributions of $y$ changes while the class-conditional distribution remains the same. 
Label distribution shift is also know as \textit{prior probability shift} and formally defined as $p_S(y) \neq p_T(y), p_S(x|y) = p_T(x|y)$.  
Label distribution shift is typically concerned in applications where the label $y$ is the casual variable for the observed feature $x$ \cite{lipton2018detecting}. 
For example, a trained model to predict pneumonia (i.e., label $y$) using chest X-ray data (i.e., features $x$) that was collected during summer time (when $p(y)$ is low), should still require it to be accurate on patients (i.e., new inputs) visiting in winter time (when $p(y)$ is high) regardless of label distribution shift. 
Long-tailed distribution \cite{van2017devil} is a special case for label distributional shift where the training set $p_S(y)$ follows a long-tailed distribution but the test set is balanced (i.e., $p_T(y)$ roughly follows a uniform distribution).

\vspace{-0.6em}
\paragraph{Out-of-Distribution Samples} are test time inputs that are outliers to the training set without any semantic content shared with the training distribution, which is considered beyond reasonably foreseeable domain shifts. 
For example, given a model trained to recognize handwritten characters in English, a Roman character with a completely disjoint label space is an Out-of-Distribution (OOD) test sample. 
OOD detection \cite{hendrycks2016baseline} is a common approach to detect such outlier samples, whose predictions should be abstained (see Section \ref{sec:ood-detection} for details).

\vspace{-0.6em}
\paragraph{Benchmark Datasets for Covariate Shift:} 
Several variants of the ImageNet dataset have been introduced to benchmark distributional error (i.e., evaluating robustness against distributional shifts) when the model is trained on the original ImageNet dataset.
\citet{hendrycks2019benchmarking} introduce two variants of the original ImageNet validation set: ImageNet-C benchmark for input corruption robustness and the ImageNet-P dataset for input perturbation robustness. 
ImageNet-A \cite{hendrycks2019natural} sorts out unmodified samples from ImageNet test set that falsifies state-of-the-art image classifiers. 
\citet{hendrycks2020many} present a series of benchmarks for measuring model robustness to variations on image renditions (ImageNet-R benchmark), imaging time or geographic location (StreetView benchmark), and objects size, occlusion, camera viewpoint, and zoom (DeepFashion Remixed benchmark). 
\citet{recht2019imagenet} collected ImageNet-V2 using the same data source and collection pipeline as the original ImageNet paper \cite{deng2009imagenet}.
This new benchmark leads to the observation that the prediction accuracy of even the best image classifiers are still highly sensitive to minutiae of the test set distribution and extensive hyperparameter tuning. 
Shifts \cite{malinin2021shifts} is another recent benchmark dataset for distributional shifts beyond the computer vision tasks.

\vspace{-0.6em}
\paragraph{Benchmark Datasets for Label Distribution Shift:} 
Synthetic label distribution shift is a common benchmarking method in which the test set is manually sampled according to a predefined target label distribution $p_T(y)$ that is different from the source label distribution $p_S(y)$ \cite{lipton2018detecting}. % ,wu2021online
\citet{wu2021online} is an example of a real-world label distribution shift benchmark for text domain.
% Specifically, the authors collected papers from 23 categories on arXiv and extracted the tf-idf vector from the abstract as the features.

\vspace{-0.6em}
\paragraph{Benchmark Datasets for OOD Detection:}
Test sets of natural images with disjoint label space are typically used for benchmarking OOD detection. 
For example, a model trained on the CIFAR10 dataset may use ImageNet (samples that do not overlap with CIFAR10 labels) as OOD test samples. 
ImageNet-O \cite{hendrycks2019natural}, containing 2000 images from 200 classes within ImageNet-22k and outside ImageNet is an example for the ImageNet-1k dataset.
Hendrycks et al. \cite{hendrycks2019scaling} presented three large-scale and high-resolution OOD detection benchmarks for multi-class and multi-label image classification, object detection, and semantic segmentation, respectively. 
\citet{yang2021semantically} presents semantically coherent OOD (SC-OOD) benchmarks for CIFAR10 and CIFAR100 datasets.
In another work, \citet{chan2021segmentmeifyoucan} presents benchmarks for anomalous object segmentation and road obstacle segmentation.

\vspace{-0.5em}
\subsection{Adversarial Error}
\label{sec:adversarial_attack}
\emph{Adversarial Error} is the model misprediction due to synthetic perturbations (termed as adversarial perturbations) added to the original clean sample. 
The adversarial attack is the act of generating adversarial perturbations to cause intentional model mispredictions while keeping the identical semantic meaning of the clean sample. 
Different forms of adversarial attacks have been studied on different types of data.
On image data, typical forms include the $\ell_p$ constrained additive perturbation \cite{goodfellow2014explaining}, spatial perturbation~\cite{xiao2018spatially}, and semantically meaningful perturbation~\cite{qiu2020semanticadv}. 
Beyond the image data, adversarial attacks can also be designed, such as by altering the shape of 3D surfaces \cite{xiao2019meshadv}, by replacing words with synonyms \cite{ren2019generating} or rephrasing the sentence \cite{iyyer2018adversarial} in natural language data, by applying adversarial printable 2D patches on real-world physical objects \cite{brown2017adversarial}.

\vspace{-0.5em}
\paragraph{Benchmark Datasets:} 
Evaluating adversarial error (also known as adversarial robustness) is usually done by measuring empirical performance (e.g., accuracy in image classification tasks) on a set of adversarial samples. 
However, the key requirement for a faithful evaluation is to use strong and diverse unseen attacks to break the model. % that the attackers should try their best to break the model.
There are two commonly used strategies to achieve this goal.
First, an ensemble of multiple strong and diverse adversarial attacks should be simultaneously used for evaluation. 
For instance, AutoAttack \cite{croce2020reliable} consists of four state-of-the-art white-box attacks and two state-of-the-art black-box attacks. 
\citet{kang2019testing} present another setting by creating evaluation benchmarks of ImageNet-UA and CIFAR-10-UA, which contain both $\ell_p$ constrained adversarial attacks and real-world adversarial attacks such as worst-case image fogging. 
Second, the attacks should be carefully designed to prevent the ``gradient obfuscation'' effect \cite{athalye2018obfuscated,carlini2017adversarial}.
Since the success of traditional white-box attacks depends on the accurate calculation of model gradients, they may fail if the model gradients are not easily accessible (e.g., the model has non-differential operations). 
As a result, evaluating model robustness on such attacks may provide a false sense of robustness. 
\citet{athalye2018obfuscated} proposed three enhancements for traditional white-box attacks as solutions for common causes of gradient obfuscation.
Other solutions include designing adaptive attacks for each specific defense strategy \cite{tramer2020adaptive} (see Section \ref{sec:adv-detection} for details).
% RobustBench \cite{croce2020robustbench} is another standardized benchmark of adversarial robustness, which as accurately as possible reflects the robustness of the considered models within a reasonable computational budget. 
% The authors evaluate model robustness against AutoAttack \cite{croce2020reliable} and also analyzed general trends in adversarial robustness and its impact on other tasks such as robustness to various distribution shifts and out-of-distribution detection. 

\vspace{-0.5em}
\section{ML Safety Taxonomy}
\label{sec:categorization}

Looking at fundamental limitations of code-based software safety for machine learning on one hand and research debt in AI safety on the other hand, we review and organize practical ML solutions into a taxonomy for ML safety. % improve ML dependability and safety
The proposed taxonomy unifies ML dependability objective with engineering safety strategies for their safe execution in open-world scenarios. 
% .... into three safety strategies for improving dependability of ML-based systems for their safe execution in open-world scenarios. 
% In order to find a common ground in ML safety needs, we connect ML techniques with relevant engineering safety strategies....
% After reviewing a broad spectrum of state-of-the-art ML techniques to achieve reasonable coverage and diversity of research papers. 
Our ML safety taxonomy is followed by a systematic and broad review of relevant ML techniques to serve as a way to checkup coverage and diversity of safety strategies employed in any given ML system.

As illustrated in Figure \ref{fig:roadmap} and Table \ref{tab:ref-table}, we propose categorizing ML techniques in three following safety strategies: 
\begin{itemize}
    \item \emph{(1) Inherently Safe Model:} refers to techniques for designing ML models that are intrinsically error-free or verifiable to be error-free in their intended target domain. 
    We review model transparency and formal methods such as model specification, verification, and formal testing as main pillars to achieve the inherently safe design. 
    However, there are many open challenges for these solutions to guarantee ML safety.
    
    \item \emph{(2) Enhancing Performance and Robustness:} refers to techniques to increase model performance (on the source domain) and robustness against distributional shifts. % close the generalization gap on target domain. 
    % \item \emph{(2) Model Robustification:} refers to techniques to increase model robustness against distributional shifts. 
    Perhaps the most commonly used in practice, these techniques contribute to safety by improving the operational performance of ML algorithms.
    % We review key approaches and techniques that advance model performance, domain shift robustness, and adversarial robustness. 
    We review key approaches and techniques such as training regularization, domain generalization, adversarial training, etc. 

    \item \emph{(3) Run-time Error Detection} refers to strategies to detect model mispredictions at the run-time (or test-time) to prevent model errors from becoming system failures. 
    This strategy can help mitigating hazards related to ML performance limitations in the operational domain. % at the time of fault by using internal or external run-time error detection techniques. %  and graceful degradation plans such as notifying the driver to take vehicle's control.
    We review key approaches and techniques for model uncertainty estimation, out-of-distribution detection, and adversarial attack detection. 
\end{itemize}

Additionally, we emphasize on safety-oriented \textit{Human-AI Interaction} design as a type of \textit{Procedural Safeguards} to prioritize end-user awareness and trust, and misuse prevention for non-experts end-users of ML-based products.
Also, we differentiate ML safety from Security because the external factors (i.e., attacker) which intentionally exploit system vulnerabilities are the security threats rather than a design limitation. %  (see Section~\ref{sec:security-threat}). 
Table~\ref{tab:ref-table} presents a summary of reviewed techniques and papers for each safety strategy. 
Reviewed papers are organized into different solutions (middle column) to group papers into individual research approaches. 
We go through details and describe complement each other in the following sections.

\begin{table*}[]
\centering
\caption{A taxonomy of techniques for ML safety with the left column identifying key ML safety strategies, middle column presenting relevant ML solutions, and right column listing machine learning techniques with representative research papers.}
\label{tab:ref-table}
\vspace{-0.3em}
\resizebox{\textwidth}{!}{%
\begin{tabular}{cll}
% \hline
\toprule 
\textbf{Safety Strategy} & \multicolumn{1}{c}{\textbf{ML Solutions}} & \multicolumn{1}{c}{\textbf{ML Techniques}} \\ \hline
\multirow{9}{*}{\textbf{\begin{tabular}[c]{@{}c@{}}Inherently Safe \\ Design\end{tabular}}} & \multirow{3}{*}{Model Transparency} & Visualization Tools (\cite{maaten2008visualizing,hohman2019summit,strobelt2018lstmvis,wongsuphasawat2017visualizing,kahng2018cti}) \\ \cline{3-3} 
&  & Global Explanations  (\cite{LageRGKD18,lakkaraju2016interpretable,guidotti2018local,lakkaraju2019faithful,lakkaraju2020robust,wu2018beyond,kim2018interpretability,NEURIPS2019_77d2afcb,yeh2020completeness}) \\ \cline{3-3} 
&  & Local Explanations (\cite{ribeiro2016should,ribeiro2018anchors,lundberg2017unified,shrikumar2017learning,zeiler2014visualizing,simonyan2013deep,smilkov2017smoothgrad,springenberg2014striving,bach2015pixel,selvaraju2017grad}) \\ \cline{2-3} 
& \multirow{2}{*}{Design Specification} & Model Specification (\cite{pei2017towards,seshia2016towards,seshia2018formal,bartocci2018specification,dreossi2019formalization,sadigh2016information}) \\ \cline{3-3} 
&  & Environment Specification (\cite{sadigh2016planning}) \\ \cline{2-3} 
& \multirow{4}{*}{\makecell[l]{Model Verification \\ and Testing}} & Formal Verification (\cite{wang2018efficient,narodytska2018verifying,katz2017reluplex,dutta2017output,huang2017safety}) \\ \cline{3-3} 
&  & Semi-Formal Verification (\cite{dvijotham2018training,dreossi2019compositional,chakraborty2014distribution,dreossi2019verifai}) \\ \cline{3-3} 
&  & Formal Testing (\cite{zhang2020machine,lakkaraju2017identifying,qin2018syneva,bansal2018coverage,sun2018testing,pei2017deepxplore}) \\  \cline{3-3}
&  & End-to-End Testing (\cite{yamaguchi2016combining,fremont2020formal,Dreossi2018aug,kim2020programmatic}) \\ 
\hline
\multirow{10}{*}{\textbf{\begin{tabular}[c]{@{}c@{}}\makecell[c]{Enhancing Performance \\ and Robustness}\end{tabular}}} 
& \multirow{2}{*}{\makecell[l]{Robust Network\\Architecture}} & Model Capacity (\cite{lin2018defensive,nakkiran2019adversarial,djolonga2020robustness,madry2017towards,gui2019model,hu2020triple,wang2020once,ye2019adversarial,sehwag2020pruning})  \\ \cline{3-3} 
&  & Model Structure and Operations (\cite{guo2020meets,chen2020anti,ning2020multi,xie2020smooth,tavakoli2021splash,vasconcelos2020effective,zhang2019making}) \\ \cline{2-3}
& \multirow{4}{*}{Robust Training} & Training Regularization  (\cite{zheng2016improving,zhang2017universum,yuan2020revisiting,muller2019does,pan2018two,wang2019learning,wang2019learning2,huang2020self}) \\ \cline{3-3} 
&  & Pretraining and Transfer Learning (\cite{yosinski2014transferable,you2020does,hendrycks2019using,chen2020adversarial,jiang2020robust,yue2019domain}) \\ \cline{3-3} 
&  & \begin{tabular}[c]{@{}l@{}}Adversarial Training (\cite{goodfellow2014explaining,zhang2019theoretically,ding2020mma,zhang2019you,shafahi2019adversarial,wong2018provable,gowal2018effectiveness,zhang2019towards})\end{tabular} \\ \cline{3-3} 
%   &  & \begin{tabular}[c]{@{}l@{}}Domain Generalization (\cite{wang2019learning,wang2019learning2,wang2020high,huang2020self,tobin2017domain,tremblay2018training,lee2020network,xu2021robust,zheng2019joint,li2018learning,balaji2018metareg,gong2019dlow,lambert2020mseg,wu2019delving,tang2019pamtri})\end{tabular} \\ \cline{3-3}
&  & \begin{tabular}[c]{@{}l@{}}Domain Generalization (\cite{tobin2017domain,tremblay2018training,lee2020network,wang2019learning,li2018learning,balaji2018metareg})\end{tabular} \\ \cline{2-3}
%     &  & \begin{tabular}[c]{@{}l@{}}Domain Randomization (\cite{tobin2017domain,tremblay2018training,lee2020network,xu2021robust,zheng2019joint})\end{tabular} \\ \cline{3-3}
%   &  & \begin{tabular}[c]{@{}l@{}}Robust Representation Learning (\cite{wang2019learning,wang2019learning2,wang2020high,huang2020self}) \end{tabular} \\ \cline{3-3} 
%   &  & \begin{tabular}[c]{@{}l@{}}Multi-source Training (\cite{li2018learning,balaji2018metareg,gong2019dlow,lambert2020mseg,wu2019delving,tang2019pamtri})\end{tabular} \\ \cline{2-3} 
% &  & \begin{tabular}[c]{@{}l@{}}Domain Generalization (\cite{tobin2017domain,wang2019learning,gong2019dlow})\end{tabular} \\ \cline{2-3} 
% & \multirow{3}{*}{Domain Generalization} & \begin{tabular}[c]{@{}l@{}}Domain Randomization (\cite{tobin2017domain,tremblay2018training,lee2020network,xu2021robust,zheng2019joint})\end{tabular} \\ \cline{3-3} 
% &  & \begin{tabular}[c]{@{}l@{}}Robust Representation Learning \cite{wang2019learning,wang2019learning2,wang2020high,huang2020self} \end{tabular} \\ \cline{3-3} 
% &  & \begin{tabular}[c]{@{}l@{}}Multi-source Training (\cite{li2018learning,balaji2018metareg,gong2019dlow,lambert2020mseg,wu2019delving,tang2019pamtri})\end{tabular} \\ \cline{2-3} 
& \multirow{4}{*}{\makecell[l]{Data Sampling\\and Augmentation}} & Active Learning (\cite{gal2017deep,beluch2018power,siddiqui2020viewal,haussmann2020scalable})\\ \cline{3-3} 
&  & Hardness Weighted Sampling (\cite{zhang2020geometry,fidon2020distributionally})\\ \cline{3-3} 
&  & Data Cleansing (\cite{beyer2020we,yun2021re,han2019deep}) \\ \cline{3-3} 
&  & Data Augmentation (\cite{cubuk2018autoaugment,zhong2017random,hendrycks2019augmix,geirhos2018imagenet,yun2019cutmix,hendrycks2020many,volpi2018generalizing,zhao2020maximum}) \\ \hline
\multirow{7}{*}{\textbf{\begin{tabular}[c]{@{}c@{}}Run-time Error Detection\end{tabular}}} & \multirow{2}{*}{\makecell[l]{Prediction Uncertainty}} & Model Calibration (\cite{guo2017calibration,szegedy2016rethinking,li2020improving,kumar2018trainable,zhang2019confidence}) \\ \cline{3-3} 
 &  & Uncertainty Estimation (\cite{der2009aleatory,lakshminarayanan2017simple,gal2016dropout,gal2017concrete,van2020uncertainty,meinke2019towards,ovadia2019can,mukhoti2021deterministic,liu2020simple,chen2020angular}) \\ \cline{2-3} 
  & \multirow{3}{*}{\begin{tabular}[c]{@{}l@{}}Out-of-distribution \\ Detection\end{tabular}} & Distance-based Detection (\cite{lee2018simple,techapanurak2020hyperparameter,ruff2018deep,ruff2019deep,bergman2020classification,sastry2019detecting,tack2020csi,sohn2020learning,vyas2018out}) \\ \cline{3-3} 
 &  & Classification-based Detection (\cite{hendrycks2018deep,Hendrycks2019sv,lee2017training,hsu2020generalized,yu2019unsupervised,pmlr-v119-goyal20c,golan2018deep,mohseni2020ood}) \\ \cline{3-3} 
 &  & Density-based Detection (\cite{schlegl2017unsupervised,zong2018deep,Choi2020Novelty,serra2019input,wang2020further,du2019implicit,grathwohl2019your,liu2020energy}) \\ \cline{2-3} 
  & \multirow{2}{*}{\begin{tabular}[c]{@{}l@{}}Adversarial Attack \\ Detection and Guard\end{tabular}} & Adversarial Detection (\cite{feinman2017detecting,li2017adversarial,xu2017feature,grosse2017statistical,meng2017magnet,ma2018characterizing,hendrycks2016early,gong2017adversarial})
  \\ \cline{3-3} 
 &  & Adversarial Guard (\cite{guo2017countering,das2017keeping,samangouei2018defense,shaham2018defending,xie2017mitigating}) \\ % \cline{3-3}
%  & & Gradient Masking and Defense Evaluation (\cite{carlini2017adversarial,athalye2018obfuscated}) \\ 
\bottomrule
\end{tabular}
}\vspace{-0.7em}
\end{table*}

\vspace{-0.5em}
\section{Inherently Safe Design} 
\label{sec:safe-design}

Achieving inherently safe ML algorithms that are provably error-less w.r.t. is still an open problem (NP-hard in high-dimensional domains) despite being trivial for code-based algorithms.
% The inherently safe design represents the goal of building ML systems that are provably error-less w.r.t. specified operational domain.  % and requirements
In this section, we review the three main requirements to achieve safe ML algorithms as being \textit{(1) model transparency}, \textit{(2) formal specification}, and \textit{(3) formal verification and testing}.
These three requirements aim to formulate high-level system design specifications into low-level task specifications, leading to transparent system design, and formal verification or testing for model specifications.

\vspace{-0.5em}
\subsection{Model Transparency}

Transparency and interpretability of the ML model is an essential requirement for trustworthy, fair, and safe ML-based systems in real-world applications \cite{mohseni2018multidisciplinary}. 
However, advanced ML models with high performance on high dimensional domains usually have very large parameter space, making them hard to interpret for humans.
In fact, the interpretability of a ML model is inversely proportional to its size and complexity. 
For example, a shallow interpretable model like a decisions tree becomes uninterpretable when a large number of trees are ensembled to create a random forest model. 
The inevitable trade-off between model interpretability and performance limits the transparency for deep models to ``explaining the black-box'' in a human understandable way w.r.t explanations complexity and length \cite{doshi2017towards}.
Regularizing the training for model interpretability is a way to improve model transparency for low-dimensional domains. 
For example, Lage et al. \cite{LageRGKD18} present a regularization to improve human interpretability by incorporate user feedback in model training by measuring users' mean response time to predict the label assigned to each data point at inference time.  
Lakkaraju et al. \cite{lakkaraju2016interpretable} build predictive models with sets of independent interpretable if-then rules. 
In the following, we review techniques for model transparency in three parts: model explanations (or global explanations), prediction explanations (or instance explanations), and evaluating the truthfulness of explanations.

\vspace{-0.3em}
\subsubsection{Model Explanations}

Model Explanations are techniques to estimate ML models for explaining the representation space or what the model has learned. 
Additionally, ML visualization and analytic systems provide various monitoring and inspection tools for training data \cite{wexler2017facets}, data flow graphs \cite{wongsuphasawat2017visualizing}, training process \cite{strobelt2018lstmvis} and inspecting a trained model \cite{hohman2019summit}.

%%%% model estimation 
\vspace{-0.3em}
\paragraph{Model Estimations.} One way to explain ML models is through estimation and approximation of deep models to generate simple and human understandable representations. 
The descriptive decision rule set is a common way to generate interpretable model explanations. 
For example, Guidotti et al. \cite{guidotti2018local} present a technique to train local decision tree (i.e., interpretable estimate) to explain any given black-box model. 
Their explanation consists of a decision rule for the prediction and a set of counterfactual rules for the reversed decision.
Lakkaraju et al. \cite{lakkaraju2019faithful} propose to explain deep models with a small number of compact decision sets through subspace explanations with user-selected features of interest.
Ribeiro et al. \cite{ribeiro2018anchors} introduce Anchors, a model-agnostic estimator that can explain the behavior of deep models with high-precision rules applicable for different domains and tasks.
To address this issue, Lakkaraju et al. \cite{lakkaraju2020robust} present a framework to optimize a minimax objective for constructing high-fidelity explanations in the presence of adversarial perturbations. 
In a different direction, Wu et al. \cite{wu2018beyond} present a tree regularization technique to estimate a complex model by learning tree-like decision boundaries. 
Their implementations on time-series deep models show that users could understand and trust decision trees trained to mimic deep model predictions.

%%%%  visual concepts
\vspace{-0.3em}
\paragraph{Visual Concepts.} 
Exploring a trained network by its learned semantic concepts is another way to inspect model rationale and efficiency in recognizing patterns. 
Kim et al. \cite{kim2018interpretability} introduce concept activation vectors as a solution to translate model representation vectors to human understanding of concepts. 
They create high-level user-defined concepts (e.g., texture patterns) by training auxiliary linear ``concept classifiers'' with samples from the training set to draw concepts that are important in model prediction.
Ghorbani et al. \cite{NEURIPS2019_77d2afcb} take another direction by clustering the super-pixel segmentation of image saliency maps to discover visual concepts.
% Their post-training explanation method provides meaningful concepts to create coherent and important examples for model prediction.
Along the same line, Zhou et al. \cite{zhou2018interpretable} propose a framework for generating visual concepts by decomposing the neural activations of the input image into semantically interpretable components pre-trained from a large concept corpus. 
Their technique is able to disentangle the features encoded in the activation feature vectors and quantify the contribution of different features to the final prediction.
In another work, Yeh et al. \cite{yeh2020completeness} study the completeness of visual concepts with a completeness score to quantify the sufficiency of a particular set of concepts in explaining the model's prediction. 

\vspace{-0.3em}
\subsubsection{Instance Explanations}
Instance or local explanations explain the model prediction for specific input instances regardless of overall model behavior. 
This type of explanation carries less holistic information about the model but informs about model behavior near the examples input, which is suitable for investigating the edge cases for model debugging. 
% Saliency map (also as known as attribution or sensitivity map) is a common instance explanation type to explain the reasoning behind a model prediction by identifying which features in the input is strongly influencing it. 
%%%%  counterfactual example
% Counterfactual explanation is another type of instance explanation that enables model-agnostic exploration to find a link between how the input should be changed (i.e., a hypothetical example) for the model to predict a specific class \cite{verma2020counterfactual}.  % removed wachter2017counterfactual
% For example, Goyal et al. \cite{goyal2019counterfactual} presents counterfactual explanation that compares two images (belonging to $c_1$ and $c_2$ classes) to generate visual explanations that shows which regions in the each image have made the model to predict the class $c_1$ instead of class $c_2$.  
% In another work, Mothilal et al. \cite{mothilal2020explaining} propose a framework for generating and evaluating a diverse set of counterfactual explanations. 
% Their framework compares counterfactual explanations with other instance explanation techniques for their effectiveness to  satisfy ``feasibility'' of the counterfactual actions given user context and constraints, and ``diversity'' among the counterfactuals. 

%%%% local approximation - model agnostic 
\vspace{-0.3em}
\paragraph{Local Approximations.}
Training shallow interpretable models to locally approximate the deep model's behavior can provide model explanations. 
A significant benefit of local approximation techniques is their model agnostic application and clarity of the saliency feature map.
However, the faithfulness of explanations is greatly limited to factors like the heuristic technique, input example, and training set quality. 
For instance, Ribeiro et al. \cite{ribeiro2016should} proposed LIME which trains a linear model to locally mimic the deep model's prediction. 
The linear model is trained on a small binary perturbed training set located near the input sample which the labels are generated by the deep model. 
Lundberg and Lee \cite{lundberg2017unified} present a model agnostic prediction explanation technique that uses the Shapley values of a conditional expectation function from the deep model as the measure of feature importance. 
DeepLIFT \cite{shrikumar2017learning} is another local approximation technique that decomposes the output prediction for the input by backpropagating the neuron contributions of the network w.r.t each input feature. They compare activations for the specific input to its ``reference activation'' to assign feature contribution scores. 
%Although DeepLift is designed only for DNNs, it explains the difference in output by approximating model behavior using a reference or anchor input. 

%%%% saliency map - model dependent
\vspace{-0.3em}
\paragraph{Saliency Map for DNNs.}
Various heuristic gradient-based, deconvolution-based, and perturbation-based techniques have been proposed to generate saliency maps for DNNs. 
%%  gradient-based
Gradient-based methods use backpropagation to compute the partial derivative of the class prediction score w.r.t. the input image \cite{simonyan2013deep}. 
Later, Smilkov et al. \cite{smilkov2017smoothgrad} proposed to improve the noisy visualization of saliency map by introducing noise to the input. 
% CAM \cite{zhou2016learning} and 
Grad-CAM \cite{selvaraju2017grad} technique combines feature maps from DNN's intermediate layers to generate saliency maps for the target class.
%%  deconvolution-based
Zeiler and Fergus \cite{zeiler2014visualizing} propose a deconvolution-based saliency map by adding a deconvnet on each layer which provides a continuous path from the prediction back to the image. 
Similarly, Springenbeg et al. \cite{springenberg2014striving} propose a guided backpropagation technique that modifies ReLu function gradients and uses class-dependent constraints in the backpropagation process. 
For real-time applications, Bojarski et al. \cite{bojarski2018visualbackprop} presents a variant of layer-wise relevance propagation for fast execution of saliency maps.  % \cite{bach2015pixel}
%%  perturbation-based / sensitivity-based
Perturbation-based or sensitivity-based techniques measure the sensitivity of the model output w.r.t. the input features. 
For example, Zeiler and Fergus \cite{zeiler2014visualizing} calculate the saliency map by sliding fixed size patches to occlude the input image and measure prediction probability. 

\vspace{-0.3em}
\subsubsection{Explanation Truthfulness}
Since model explanations are always incomplete estimation of the black-box models, there need to be mechanisms to evaluate for both correctness and completeness of model explanations w.r.t. the main model. 
Particularly, the fidelity of the ad-hoc explanation technique should be evaluated against the black-box model itself. 
Aside from the qualitative review of model explanations and their consistency compared to similar techniques \cite{olah2018the,selvaraju2017grad,lundberg2017unified}, we look into sanity check tests and human-grounded evaluations in the following.

%%%% qualitative and proxy evaluation 
\vspace{-0.3em}
\paragraph{Sanity Checks and Proxy Tasks} Examining model explanations with different heuristic tests is shown to be an effective way to evaluate explanation truthfulness for specific scenarios. 
For example, Samek et al. \cite{samek2017evaluating} proposed a framework for evaluating saliency explanations by their correlation between saliency map quality and network performance under input perturbation. 
In a similar work, Kindermans et al. \cite{kindermans2017reliability} present the inconsistencies in saliency maps due to simple image transformations. 
Adebayo et al. \cite{adebayo2018sanity} propose three tests to measure the fidelity of any interpretability technique in tasks that are either data sensitive or model sensitive. 
% Ribeiro et al. \cite{ribeiro2016should} evaluates explanations generated by LIME against gold standard explanations directly from a sparse logistic regression model. 
% A drawback of this approach is the limitation of shallow models on learning complex data. 
Additionally, presenting the usefulness of explanations with a proxy task has been used in some literature. 
For instance, Zeiler and Fergus \cite{zeiler2014visualizing} propose using visualization of features in different layers to improve network architecture and training by adjusting network layers.
In another example, Zhang et al. \cite{zhang2018examining} present cases of evaluating explanations' usefulness to find representation learning flaws caused by biases in the training dataset as a proxy task.

% %%%% human evaluation of explanations 
\vspace{-0.3em}
\paragraph{Human Evaluation and Ground-truth} 
Evaluating model explanations with human input is based on the assumption that good model explanations should be consistent with human reasoning and understanding of data. 
Multiple works \cite{ribeiro2016should,ribeiro2018anchors,lundberg2017unified} made use of human-subject studies to evaluate model explanations. % and tested a hypothesis on how users would identify and prefer better models and explanations by only looking at explanations. 
However, there are multiple human factors in user feedback on ML explanations such as average user understanding, the task dependency and usefulness of explanations, and user trust in explanations. 
Therefore, more concise evaluation metrics are required to reveal model behavior to users, justifying the predictions, and helping humans investigate uncertain predictions \cite{Lertvittayakumjorn2019human}. % schmidt2019quantifying
Another challenge in human subject evaluations is the time and cost to run human subject studies and collect user feedback. 
To eliminate the need for repeated user studies, human annotated benchmarks have been proposed containing annotation of important features w.r.t. the target class \cite{mohseni2018human,VQA-HAT}.
% for model explanations with multi-layer user annotation of important features w.r.t. the target class. 
% Another similar work introduced a human-attention benchmark \cite{VQA-HAT} for visual question answering task. 

\vspace{-0.5em}
\subsection{Formal Methods}
Formal methods require rigorous mathematical specification and verification of a system to obtain ``guarantees'' on system behavior.
A design cycle in formal methods involves two main steps of (1) listing design specifications to meet the system requirements and then (2) verify the system to prove the delivery of requirements in the target environment. % guarantee certify verify
Specifically, formal verification certifies that the system $S$ exhibits the specified property $P$ when operating in the environment $E$. 
Therefore, system specification and verification are two complementary components in formal methods.  % design environment
Unlike the common practice in data-driven algorithms which rely on available data samples to model the environment, formal methods require exact specification of the algorithm properties. 
In contrary to model validation in ML, formal methods require verification of the system given the environment space. 
Related to ML specification and verification, Huang et al. \cite{huang2020survey} reviews ML verification methods by the type of guarantees they can provide such as deterministic guarantees, approximate bounds, and converging bounds. %into deterministic guarantee, one-sided guarantee, 
However, due to challenges in model specification and verification in high-dimensional domains, many works such as Sheshia et al.~\cite{seshia2016towards} and Yamaguchi et al. \cite{yamaguchi2016combining} suggest end-to-end simulation-based validation of AI-based systems as a semi-formal verification of complex systems in which a realistic simulation of the environment and events is used to find counterexamples for system failures. 
In the following, we review different research on formal methods for ML algorithms. % specification and verification as the two main components of . 

\vspace{-0.3em}
\subsubsection{Formal Specification}
The specification is a necessary step prior to the software development and the basis for the system verification. 
Examples of common formal specification methods in software design are temporal logic and regular expressions.
However, in the training of ML algorithms, the training set specifies the model task in the target distribution rather than a list of rules and requirements.  %  learn by the ML models, is aimed to learn the source distribution with the help of available training set and training regularization. 
% model-level
Here we review techniques and experiments in model and environment specifications.

\vspace{-0.3em}
\paragraph{Model Specification:} Specifying the desired model behavior is a design requirement prior to any system development.
However, formal specification of ML algorithms is very challenging for real-world tasks involving high-dimensional data like images. 
Sheshia et al.~\cite{seshia2016towards,seshia2018formal} review open challenges for ML specifications and survey the landscape of formal specification approaches.
For example, ML for semantic feature learning can be specified on multiple levels (e.g., system-level, input distribution level, etc.) to simplify overall specifications.
% Authors point out the importance of formal specification for system verification, semi-formal verification, and adequate testing.  
% system-level
Bartocci et al. \cite{bartocci2018specification} review tools for specifying ML systems in complex dynamic environments. 
They propose specification-based monitoring algorithms that provide qualitative and quantitative satisfaction score for the model using either simulated (online) or existing (offline) inputs.  % check model and...
Additionally, as ML algorithms carry uncertainty in their outputs, the benefits of including prediction uncertainty in the overall system specification is an open and under investigation topic \cite{mcallister2017concrete}. 
Focusing on invariance specifications, Pei et al. \cite{pei2017towards} decompose safety properties for common real-world image distortions into 12 transformation invariance properties that a ML algorithm should maintain. 
Based on these specifications, they verify safety properties of the trained model using samples from the target domain.
In the domain of adversarial perturbations, Dreossi et al. \cite{dreossi2019formalization} propose a unifying formalization to specify adversarial perturbations from formal methods perspective.

\vspace{-0.3em}
\paragraph{Environment Modeling: } Modeling the operational or target environment is a requirement in formal system specification. 
Robust data collection techniques are needed for full coverage of the problem space environment. 
Several techniques such as active learning, semi-supervised learning, and knowledge transfer are proposed to improve the training data by following design specifications and encourage the model to learn more generalizable features in the target environment. 
As a result, a training set with enough coverage can better close the generalization gap between the source training and target operational domains. 
Similarly, in AI-based systems with multiple ML components, a robust specification of the dynamic environment with its active components (e.g., humans actions) enables for better system design \cite{sadigh2016planning}.

\vspace{-0.3em}
\subsubsection{Model Verification}

Formal verification in software development is an assurance process for design and implementation validity. 
There are several approaches in software engineering such as constraint solving and exhaustive search to perform formal verification.
For instance, a constrain solver like Boolean Satisfiability (SAT) provides deterministic guarantee on different verification constraints.
However, verification of ML algorithms with conventional methods is challenging for high-dimensional data domains.
In this section, we review different approaches for system-level and algorithm-level verification on ML systems. 

\vspace{-0.3em}
\paragraph{Formal Verification}
A line of research adapts conventional verification methods for ML algorithms.
For example, Narodytska et al. \cite{narodytska2018verifying} present a SAT solver for Boolean encoded neural networks in which all weights and activations are binary functions. 
Their solution verifies various properties of binary networks such as robustness to adversarial examples; however, these solvers often perform well when problems are represented as a Boolean combination of constraints. 
Recent SMT solvers for neural networks present use cases of efficient solvers for DNN verification on airborne collision avoidance and vehicle collision prediction applications \cite{katz2017reluplex}.
However, the proposed line of solutions is limited to small models with limited number of parameters commonly used on low dimensional data. 
Additionally, given the complexity of DNNs, the efficiency and truthfulness of these verification techniques require sanity checks and comparisons against similar techniques \cite{dutta2017output}. 
Huang et al. \cite{huang2017safety} present an automated verification framework based on SMT theory which applies image manipulations and perturbations such as changing camera angle and lighting conditions. 
Their technique employs region-based exhaustive search and benefits from layer-wise analysis of perturbation propagation.
% Neurify 
Wang et al. \cite{wang2018efficient} combine symbolic interval and linear relaxation that scales to larger networks up to 10000 nodes. 
Their experiments on small image datasets verify trained networks for perturbations such as brightness and contrast.

\vspace{-0.3em}
\paragraph{Quantitative and Semi-formal Verification}
Given the complexity of ML algorithms, quantitative verification assigns quality values to the ML system rather than a Boolean output. 
% However, solution beyond the standard validation test sets are needed for this matter. 
For example, Dvijotham et al. \cite{dvijotham2018training} present a jointly predictor-verifier training framework that simultaneously trains and verifies certain properties of the network. 
Specifically, the predictor and verifier networks are jointly trained on a combination of the main task loss and the upper bound on the worst-case violation of the specification from the verifier.
Another work presents random sample generators within the model and environment specification constraints for quantitative verification \cite{chakraborty2014distribution}.

%%% system-level %%%
Further, system-level simulation tools provide the environment to generate new scenarios to illustrate various safety properties of the intelligent system.
For example, Leike et al. \cite{leike2017ai} present a simulation environment that can decompose safety problems into robustness and specification limitations.
In more complex AI-based systems, Dreossi et al. \cite{dreossi2019compositional} presents a framework to analyze and identify misclassifications leading to system-level property violations. 
Their framework creates an approximation of the model and feature space to provide sets of misclassified feature vectors that can falsify the system.
Another example is VerifAI \cite{dreossi2019verifai}, a simulation-based verification and synthesis toolkit guided by formal models and specifications. 
VerifAI consists of four main modules to model the environment to abstract feature space, search the feature space to find scenarios that violate specifications, monitor the properties and objective functions, and analyze the counterexamples found during simulations.

\vspace{-0.3em}
\subsubsection{Formal Testing}
Testing is the process of evaluating the model or system against an unseen set of samples or scenarios. 
Unlike model verification, testing does not require formal specification of the system or environment but instead focuses only on the set of test samples.
The need for a new and unseen test set is because often model errors happen due to systematic biases in the training data resulting in learning incorrect or incomplete representations of the environment and task. 
Structural coverage metrics such as statement and modified condition/decision coverage (MC/DC) have been used in code-based algorithms to measure and ensure the adequacy of testing of safety-critical applications. 
However, testing in high-dimensional space is expensive as it requires very large number of test scenarios to ensure adequate coverage and an oracle to identify failures \cite{zhang2020machine}.
We review these two aspects in the following.
% Two important aspects of testing are the test coverage to ensure the completeness of the test scenarios and end-to-end system testing. 
% environment and system for semi-formal verification. 

\vspace{-0.3em}
\paragraph{Test Coverage} 
The coverage of test scenarios or samples is a particularly important factor to satisfy the testing quality.
Inspired by the MC/DC test coverage criterion, Sun et al. \cite{sun2018testing} propose a DNN specific test coverage criteria to balance between the computation cost and finding erroneous samples. 
They developed a search algorithm based on gradient descent which looks for satisfiable test cases in an adaptive manner. 
Pei et al. \cite{pei2017deepxplore} introduce the neuron coverage metric as the number of unique activated neurons (for the entire test set) over the total number of neurons in the DNN.
They present DeepXplore framework to systematically test DNN by its neuron coverage and cross-referencing oracles. 
Similar to adversarial training setups, their experiments demonstrate that searching for samples that both trigger diverse outputs and achieving high neuron coverage in a joint optimization fashion can improve model prediction accuracy. 

In guided search for testing, Lakkaraju et al. \cite{lakkaraju2017identifying} present an explore-exploit strategy for discovering unknown-unknown false positive samples in the unseen $D_{test}$. % by partitioning the test space into descriptive patterns. 
Later, Bansal and Weld \cite{bansal2018coverage} present a search algorithm which is formulated as an optimization problem that aims to select samples from $D_{test}$ maximize the utility model subject to a budget of maximum number of oracle calls. 
Differential testing techniques use multiple trained copies of the target algorithm to serve as a correctness oracle for cross-referencing \cite{qin2018syneva}. 

\vspace{-0.3em}
\paragraph{End-to-end Simulations}

End-to-end simulation tools enable testing for complex AI-based systems by generating diverse test samples and evaluating system components together. 
For instance, Yamaguchi et al. \cite{yamaguchi2016combining} present a series of simulations to combine requirement mining and model checking in simulation-based testing for end-to-end systems.
Fremont et al. \cite{fremont2020formal} present a scenario-based testing tool which generates test cases by combining specification of possible scenarios and safety properties. In many cases, samples and scenarios from the simulation tests are used to improve the training set.
For example, Dreossi et al. \cite{Dreossi2018aug} propose a framework for generating counterexamples or edge-cases for ML to improve both the training set and test coverage. 
Their experiment results for simulation-based augmentation show edge-cases have important properties for retraining and improving the model. 
In the application of object detection, Kim et al. \cite{kim2020programmatic} present a framework to identify and characterize misprediction scenarios using high-level semantic representation of the environment. 
Their framework consists of an environment simulator and rule extractor that generates compact rules that help the scene generator to debug and improve the training set.

\vspace{-0.5em}
\subsection{Challenges and Opportunities}

% ---- (1) 
The first main challenge of designing inherently safe ML models lies in the computation complexity and scalability of solutions \cite{huang2020survey,zhang2020machine}. 
As ML models are becoming exponentially more complex, it will become extremely difficult to impose specifications and perform verification mechanisms that are well-adapted for large ML models. 
A practical solution could be the modular approach presented by \citet{dreossi2019compositional} for scaling up formal methods to large ML systems, even when some components (such as perception) do not themselves have precise formal specifications. 

% ---- (2) 
On the other hand, recent advancements in 3D rendering and simulation have introduced promising solutions for end-to-end testing and semi-formal verification in simulated environments.
However, it is challenging to mitigate the gap between simulation and real-world situations, causing questionable transfer of simulated verification and testing results. 
Recent work starts exploring how simulated formal simulation aid in designing real-world tests \cite{fremont2020formal}. 
Additionally, thorough and scenario-based simulations enable system verification in broader terms such as monitoring interactions between ML modules in a complex system.
% For instance, e.g., how would an attack on the perception module affect the control module.

%  and system's mitigation plan.
% current evaluation paradigms of ML safety are also isolated by different stages, and often focus on simplistic measures that accounts for each stage only and may not be relevant to the end security goal. To verify relevance to security and wide applicability, defenses have to be measured in a novel testbed employing scenario-based evaluations, e.g., how an attack on the perception module affect the control module and feedback loop on its top.

% \subsection{Algorithm Robustness}
% \section{Evaluating and Enhancing Model Robustness}
% \section{Model Performance and Robustness}
\vspace{-0.5em}
\section{Enhancing Model Performance and Robustness}
\label{sec:robustness-performance}

Enhancing model performance and robustness is the most common strategy to improve product quality and reduce the safety risk of ML models in the open world. 
Specifically, techniques to enhance model performance and robustness reduce different model error types to gain dependability w.r.t criteria reviewed in Section \ref{sec:error-types}. 
% Intuitively, stronger robustness means making fewer mistakes by taking more ``pre-caution": a ``wisdom" that might count on more to learn. 
In the following, we review and organize ML solutions to improve performance and robustness into three parts focusing on (1) robust network architecture, (2) robust training, and (3) data sampling and manipulation. 
% (1) model capacity and architecture, (2) training and regularization, and (3) data and augmentations for open environments. 

\vspace{-0.5em}
% \subsection{Model Capacity and Architecture} % for Robustness
\subsection{Robust Network Architecture} % for Robustness

Model robustness can be influenced by model capacity and architecture.

\vspace{-0.2em}
\paragraph{Model Capacity}
Djolonga et al. \cite{djolonga2020robustness} 
% studied the impacts of model capacity and training set size on distributional shift robustness. The results 
showed that with enough training data, increasing model capacity (both width and depth) consistently helps model robustness against distribution shifts.
Madry et al. \cite{madry2017towards} showed that increasing model capacity (width alone) could increase model robustness against adversarial attacks. 
Xie et al. \cite{xie2019intriguing} observed that increasing network depth for adversarial training could largely boost adversarial robustness, while the corresponding clean accuracy quickly saturates as the network goes deeper. 
The above empirical findings that larger models lead to better robustness are also consistent with theoretical results \cite{nakkiran2019adversarial,gao2019convergence}.
In contrast, Wu et al. \cite{wu2020does} conducted a thorough study on the impact of model width on adversarial robustness and concluded that wider neural networks may suffer from worse perturbation stability and thus worse overall model robustness.
To accommodate for computational resources while maintaining model robustness, a surge of works have studied robustness-aware model compression \cite{lin2018defensive,gui2019model,hu2020triple,wang2020once,ye2019adversarial,sehwag2020pruning}.
% For instance, Lin et al. \cite{lin2018defensive} observe a non-monotonic relationship between model size and robustness in deep network quantization. 
% In studying how sparsity affects DNN robustness, Guo et al. \cite{guo2018sparse} empirically discovered that an appropriately higher deep model sparsity led to better robustness, whereas over-sparsification could in turn cause fragility. 

%%%% effects of model structure on robustness %%%%
\vspace{-0.3em}
\paragraph{Network Structure and Operator}  
 
Activation functions may play an important role in model robustness \cite{xie2020smooth,tavakoli2021splash}.
For example, Xie et al. \cite{xie2020smooth} observed that using smoother activation functions in the backward pass of training improves both model accuracy and robustness. 
Tavakoli et al. \cite{tavakoli2021splash} proposed a set of learnable activation functions, termed SPLASH, which simultaneously improves model accuracy and robustness.
\citet{zhang2019making} and \cite{vasconcelos2020effective} pointed out the use of down-sampling methods (e.g., pooling, strided convolutions) in modern DNNs leads to aliasing issues and pool input invariance under image shifting.
The authors proposed to use traditional anti-aliasing methods (e.g., by adding low-pass filters after sampling operations) to increase model invariance under image shifting and model robustness against distribution shifts.
% Following that, \cite{vasconcelos2020effective} modified the ResNet \cite{he2016deep} model structure by adding blur filters and using smooth activation functions. Such improved ResNet structure achieves better robustness on multiple out-of-distribution generalization tasks.
Neural Architecture Search (NAS) has also been used to search for robust network structures \cite{guo2020meets,chen2020anti,ning2020multi,mok2021advrush}.
For example, Guo et al. \cite{guo2020meets} first leveraged NAS to discover a family of robust architectures (RobNets) that are resilient to adversarial attacks. They empirically observed that using densely connected patterns and adding convolution operations to direct connection edge improve model robustness.
% Following that, Dong et al. \cite{dong2020adversarially} explored the relationship among adversarial robustness, Lipschitz constant, and architecture parameters and show that an appropriate constraint on architecture parameters could reduce the Lipschitz constant, which can further improve the model robustness. 
% The research trend of NAS for adversarially robust architectures continues to develop \cite{chen2020anti,ning2020multi}, although they still lack a clear comprehension of how an architectural inductive bias could be represented and translated to the robustness analysis.
With the recent success of Visual Transformers (ViTs), some works \cite{bhojanapalli2021understanding,mahmood2021robustness} benchmarked model robustness of ViTs and observed it to have better general robustness than traditional CNNs. 
However, a recent paper overturned such a conclusion by showing that CNNs can be as robust as ViTs if trained properly \cite{bai2021transformers}.

\vspace{-0.5em}
% \subsection{Model Training and Regularization} % Techniques
\subsection{Robust Training} % Techniques
\label{sec:RT}
% \vspace{-0.3em}

Various robust training methods have been proposed to improve model robustness. 

\vspace{-0.3em}
\subsubsection{Training Regularization}
The most representative regularization for robust training is to encourage model smoothness: Similar outputs should be obtained on similar inputs (e.g, two versions of the same images with slightly different rotation angles should lead to similar model predictions) \cite{zheng2016improving,hendrycks2019augmix,miyato2019virtual}.
A typical form of such regularization is the pairwise distance $d(f(x),f(x'))$, where $x$ and $x'$ are two different versions of the same image, and $d(\cdot,\cdot)$ is some distance measure (e.g., $\ell_2$ norm).
\citet{zheng2016improving} stabilized deep networks against small input distortions by regularizing the feature distance between the original image and its corrupted version with additive random Gaussian noises. 
% \citet{zhang2017universum} explored using unlabeled free data to regularize model training for robustness and uncertainty. 
\citet{yuan2020revisiting} showed that model distillation can be taken as a learned label smoothing \cite{muller2019does} regularization which helps in-distribution generalization.
\citet{hendrycks2019augmix} use the JSD divergence among the original image and its augmented versions as a consistency regularizer to improve model robustness against common corruptions. 
Virtual Adversarial Training (VAT) \cite{miyato2019virtual} penalizes the worst-case pairwise distance on unlabeled data, achieving improve robustness in semi-supervised learning.
% Lipschitz continuity is a well studied regularization to enforce both empirical and certified (see Section~\ref{sec:AT}) model robustness.
Besides the above pairwise distance regularization, model smoothness can also be directly regularized by adding Lipschitz continuity constraints on model weights \cite{qian2018l2,singlafantastic}.
% For example, Huster et al. \cite{qian2018l2} propose a methodology of controlling Lipschitz constants to maximize model robustness by regularizing the model to learn a class of well-conditioned neural networks in which a unit amount of change in the inputs spaces only causes at most the same unit amount of change in the outputs. 
For example, \citet{singlafantastic} proposed a differential upper bound on Lipschitz constant of convolutional layers, which can be directly optimized to improve model generalization ability and robustness.

\vspace{-0.5em}
\subsubsection{Pre-training and Transfer Learning} 
Transfer learning is consist of a range of techniques to transfer useful features from 
an auxiliary domain or task to the target domain and task for improved model generalization and robustness \cite{yosinski2014transferable}. 
Pre-training is a common transfer learning approach by first training the model on a large-scale dataset and then fine tuning the model on the target downstream domain and task (e.g., CIFAR10). 
For example, \citet{hendrycks2019using} show pre-training on ImageNet dataset can greatly improve model robustness and uncertainty on smaller target datasets like CIFAR10. % , although not beneficial for in-distribution generalization. 
To benefit from unlabeled data sources, transfer learning can be done by first pre-training the model on a self-supervised task (e.g., rotation angle prediction), and then fine tuning on the target supervised task (e.g., image classification) \cite{Hendrycks2019sv}. % on the same dataset. 
% For example, Pre-training with auxiliary tasks, 
% \citet{Hendrycks2019sv} later showed that another form of pre-training, termed self-supervised pre-training, also benefits model robustness.
Recently, \citet{chen2020adversarial} introduced adversarial training into self-supervised pre-training to provide general-purpose robust pre-trained models.  %  for the first time
% They found these robust pre-trained models can benefit the subsequent fine-tuning to boost final model robustness and reduce the computation cost. 
In another work, \citet{jiang2020robust} leveraged contrastive learning for pre-training which further boosted adversarial robustness.
% As a closely related approach, simultaneous training for multiple tasks encourages the model to learn diverse features to improve robustness. 
Since robust training is usually data-hungry \cite{tsipras2018robustness} and time-consuming \cite{zhang2019you}, it would be economically beneficial if the robustness learned on one task or data distribution can be efficiently transferred to other ones.  
\citet{awais2021adversarial} showed that model robustness can be transferred from adversarially-pretrained ImageNet models to different unlabeled target domains, by aligning the features of different models on the target domain. 
Motivated by the practical constraint in federated learning  that only some resource-rich devices can support robust training, \citet{hong2021federated} proposed the first method to transfer robustness among different devices in federated learning, while preserving the data privacy of each participant.

% \citet{you2020graph} generalized similar ideas to graph neural networks. 
% Shafahi et al. \cite{shafahi2020adversarially} demonstrated the feasibility to transfer not only performance but also robustness from a source model to a target domain. 
% More recently, \cite{chen2020automated,chen2021contrastive} show that pre-trained ImageNet knowledge can be leveraged through distillation to significantly improve zero-shot synthetic-to-real generalization. 
% Another similar work is \cite{yue2019domain} which uses category-specific ImageNet images to randomly stylize the synthetic training images. In some sense, such a strategy can also be viewed as leveraging natural image domain knowledge in addition to their category information for stylization. 

% multi-class vs. multiclass
% open-world vs open world
% test-time vs. test time
% real-time vs. real time

\vspace{-0.5em}
\subsubsection{Adversarial Training}
\label{sec:AT}

Adversarial training (AT) incorporates adversarial examples into training data to increase model robustness against adversarial attacks at test-time. 
State-of-the-art AT methods are arguably top-performers \cite{zhang2019theoretically,madry2017towards} to enhance deep network robustness against adversarial attacks. 
A typical AT algorithm optimizes a hybrid loss consisting of a standard classification loss $\cL_c$ and a adversarial robustness loss term $\cL_a$:
\begin{equation} 
    \min_{\theta} \mathbb{E}_{(x,y) \sim \mathcal{D}} ~ [(1-\lambda) \cL_c + \lambda \cL_a], \, \cL_c =\cL(f(x;\theta),y), 
\cL_a = \max_{\delta \in \mathcal{B}(\epsilon)}\cL(f(x+\delta;\theta),y)
\end{equation}
where $\mathcal{B}(\epsilon) = \{ \delta \mid \|\delta\|_\infty \leq \epsilon\}$ is the allowed perturbation set to keep samples are visually unchanged, $\epsilon$ is the radius of the $\ell_\infty$ ball $\mathcal{B}(\epsilon)$, 
and $\lambda$ is a fixed training weight hyper-parameter. 
For example common AT methods, both Fast Gradient Sign Method (FGSM-AT) \cite{goodfellow2014explaining} and Projected Gradient
Descent (PGD-AT) \cite{madry2017towards} uses an 
$\epsilon$ for the allowed perturbation set that formalizes the manipulative power of the adversary.
% \begin{equation}\label{eq:xentropy}
% \cL_c =\cL(f(x;\theta),y), 
% \cL_a = \max_{\delta \in \mathbb{B}(\epsilon)}\cL(f(x+\delta;\theta),y)
% \end{equation}
% where $\mathbb{B}(\epsilon) = \{ \delta \mid \|\delta\|_\infty \leq \epsilon\}$ is the allowed perturbation set that formalizes the manipulative power of the adversary.
% TRADES \cite{zhang2019theoretically} use the same $\cL_c$ as PGD-AT, but replace $\cL_a$ from cross-entropy to a soft logits-pairing term: $\cL_a = \max_{\delta \in \cB(\eps)}\cL(f(x+\delta;\theta),f(x;\theta))$.
% In MMA training \cite{ding2020mma}, $\cL_a$ is to maximize the margins between correctly classified images. 
Examples of variations of PGD include, TRADES \cite{zhang2019theoretically} which uses the same clean loss as PGD-AT, but replace $\cL_a$ from cross-entropy to a soft logits-pairing term. 
In MMA training~\cite{ding2020mma}, the adversarial $\cL_a$ loss is to maximize the margins between correctly classified images. 
As a novel solution to the trade-off between model accuracy and adversarial robustness, Once-for-all AT (OAT) trains a single model that can be adjusted in-situ during run-time to achieve different desired trade-off levels between accuracy and robustness \cite{wang2020once}. 
\citet{hong2022efficient}~extended OAT to federated learning (FL), achieving run-time adaptation for robustness in practical FL systems where different participant devices have different levels of safety requirements. 
% Additionally, despite the possible trade-off between adversarial robustness and prediction accuracy, some works \cite{tsipras2018robustness} show that both the model accuracy and adversarial robustness may benefit from sufficiently large model capacity, as well as more training data. 

\vspace{-0.5em}
\paragraph{Fast Adversarial Training}
%%% computation cost %%% 
Despite the effectiveness of AT methods against attacks, they suffer from very high computation cost due to multiple extra backward propagations to generate adversarial examples.
The high training cost would make AT impractical in certain domains and large-scale datasets \cite{xie2019feature}.
% For instance, Xie et al. \cite{xie2019feature} leveraged 128 Nvidia V100 GPUs and 38 hours of training to apply a combination of feature denoising with n-step PGD-AT on a ResNet-101 model for ImageNet training set. 
% A major factor in computational cost in AT is introduced by the number of gradient steps used for generating adversarial examples. 
Therefore, a line of work tries to accelerate AT. 
\citet{zhang2019you} restricted most adversarial updates in the first layer to effectively reduce the total number of the forward and backward passes to improve the training efficiency. 
Shafahi et al. \cite{shafahi2019adversarial} used a ``free'' AT algorithm by updating the network parameters and adversarial perturbation simultaneously on a single backward pass.
Wong et al. \cite{wong2020fast} showed that, with proper random initialization, single-step PGD-AT can be as effective as multiple-step ones but much more efficient.

% \vspace{-0.3em}
% \paragraph{Semi-supervised and self-supervised for Adversarial Training}
% % \paragraph{Unlabeled Data for Adversarial Training}
% Schmidt et al. \cite{schmidt2018adversarially} show that sample complexity in adversarial learning can be significantly larger than that of ``standard'' learning and hence requires more samples to achieve non-trivial adversarial robust classifier. 
% However, conventional adversarial training needs class labels and can not be easily applied to unlabeled data. 
% Adversarial training under semi-supervised and self-supervised environments can boost the adversarial robustness \cite{carmon2019unlabeled} to use unlabeled data. % removed uesato2019labels
% To benefit from self-supervised learning, \citet{hendrycks2019using} used the self-supervision to improve the robustness by simply adding a rotation loss into the training pipeline. 
% After it, Chen et al. \cite{chen2020adversarial} study different combinations between pre-train and fine-tune on self-supervised settings and showed that fine-tuning the pre-trained self-supervised learning could contribute to the dominant portion of robustness improvement. 
% Adversarial Self-Supervised Contrastive Learning
% Different from the above methods which still require the labeled data to perform adversarial training, \cite{kim2020adversarial} proposed a contrastive self-supervised learning framework to train an adversarially robust model without any class labels. 
 
\vspace{-0.5em}
%  \cite{robey2020model}
\paragraph{Certified Adversarial Training} 
% Adversarial training commonly considers the simple threat models such as $\ell_2$ or $\ell_\infty$ in which the adversarial examples is constrained to be close to the original images in $\ell_2$ or $\ell_\infty$ distance.
% However, using adversarial training to solve the inner maximal problem ($\cL_a$) for non-convex neural networks is an approximated solution which computes the {\em lower bound} and could not provide a guarantee. 
% Beyond $\ell_p$ based threat model, additionally threat models such as spatial perturbations \cite{xiao2018spatially} and semantic perturbation \cite{qiu2020semanticadv} also proposed which introduce new challenges to design adversarial training based methods for new perturbations. % removed pmlr-v97-engstrom19a
% A simple solution for these perturbations is to apply adversarial training with multiple threat models simultaneously~\cite{maini2020adversarial}. % tramer2019adversarial
% However, this strategy results in a lower robustness compared training on single type of adversarial examples and it is not possible to formalized all perturbations types. 
% In general, using adversarial training to solve the inner maximal problem ($\cL_a$) for non-convex neural networks is an approximated solution which computes the {\em lower bound} and could not provide a guarantee. 
Certified AT aims to obtain networks with \emph{provable} guarantees on robustness under certain assumptions and conditions. 
Certified AT uses a verification method to find an upper bound of the inner max, and then update the parameters based on this upper bound of robust loss. Minimizing an upper bound of the inner max guarantees to minimize the robust loss. 
% Several works  are proposed to give \emph{provable} guarantees on the robustness performance, such as 
Linear relaxations of neural networks \cite{wong2018provable} use the dual of linear programming (or other similar approaches \cite{wang2018mixtrain}) to provide a linear relaxation of the network (referred to as a ``convex adversarial polytope'') and the resulting bounds are tractable for robust optimization. % weng2018towards zhang2018crown dvijotham2018training
However, these methods are both computationally and memory intensive which can increase model training time by a factor of hundreds. 
Interval Bound Propagation (IBP) \cite{gowal2018effectiveness} is a simple and efficient method for training verifiable neural networks which achieved state-of-the-art verified error on many datasets. 
However, the training procedure of IBP is unstable and sensitive to hyperparameters. 
Similarly, Zhang et al. proposed CROWN-IBP \cite{zhang2019towards} to combine the efficiency of IBP and the tightness of a linear relaxation based verification bound.
Other certified adverarial training techniques include ReLU stability regularization \cite{xiao2018training}, distributionally robust optimization \cite{sinha2018certifying} , semi-definite relaxations~\cite{raghunathan2018certified,dvijothamefficient2019} and random smoothing~\cite{cohen2019certified}.
% In addition to techniques focusing the tightness bound on $\ell_\infty$ norm, \cite{cohen2019certified} prove a tightness guarantee under the $\ell_2$ norm via transforming the original classifier into a new smoothed classifier.  
% For example, Salman et al. \cite{salman2019provably} first show the feasibility of getting a certified defense on ImageNet.  Then employ adversarial training into a random smoothing pipeline to further improve the certified robustness. 

%Also, to improve robustness of object detection models to occlusions and deformations, Wang et al.~\cite{wang2017fast} used an adversarial network to generate hard positive examples.  

\vspace{-0.5em}
\subsubsection{Domain Generalization}
\label{sec:DG}

Domain generalization presents an important indicator of ``in-the-wild'' model robustness for open-world applications such as autonomous vehicles and robotics where deployment in different domains is common. 
% We review multiple approaches to achieve domain generalization.

%%%% domain randomization %%%%
\vspace{-0.3em}
\paragraph{Domain Randomization} Utilizing randomized variations of the source training set can improve generalization to the unseen target domain. 
% Data augmentation techniques present a simple yet effective approach to achieve zero-shot generalization without seeing test images. 
Domain randomization with random data augmentation has been a popular baseline in both reinforcement learning \cite{tobin2017domain} and general scene understanding \cite{tremblay2018training}, where the goal is to introduce randomized variations to the input to improve sim-to-real generalization. 
Yue et al. \cite{yue2019domain} use category-specific ImageNet images to randomly stylize the synthetic training images. 
% In some sense, such strategy can also be viewed as leveraging natural image domain knowledge in addition to their category information for stylization.
YOLOv4 \cite{bochkovskiy2020yolov4} benefits from a new random data augmentation method that diversifies training samples by mixing images for detection of objects outside their normal context.
Data augmentation can also be achieved through network randomization \cite{lee2020network,xu2021robust}, which introduces randomized convolutional neural networks for more robust representation. 
Data augmentation for general visual recognition tasks is discussed in Section \ref{par:data-augmentation}.

% Most references here have already described in previous sections.
\paragraph{Robust Representation Learning} 
The ability to generalize across domains also hinges greatly on the quality of learned representations. As a result, introducing inductive bias and regularizations are crucial tools to promote robust representation. For instance, Pan et al. \cite{pan2018two} show that better designed normalization leads to improved generalization. 
In addition, neural networks are prone to overfitting to superficial (domain-specific) representations such as textures and high-frequency components. 
Therefore, preventing overfitting by better capturing the global image gestalt can considerably improve the generalization to unseen domains \cite{wang2019learning,wang2019learning2,wang2020high,huang2020self}. 
For example, \citet{wang2019learning2} prevented the early layers from learning low-level features, such as color and texture, but instead focus on the global structure of the image.
Representation Self-Challenging (RSC) \cite{huang2020self} iteratively discards the dominant features during training and forces the network to utilize remaining features that correlate with labels.

%%%% multi-source training 
\paragraph{Multi-Source Training} Another stream of methods assume multiple source domains during training and target the generalization on the held-out test domains. 
They use multiple domain information during training to learn domain agnostic biases and common knowledge that also apply to unseen target domains \cite{li2017deeper,li2018learning,balaji2018metareg,tang2019pamtri,wu2019delving,gong2019dlow,lambert2020mseg}.
% A representative work is the PACS Dataset \cite{li2017deeper}, which has inspired many follow-up work~\cite{li2018learning,balaji2018metareg}. 
For example,~\citet{gong2019dlow} aimed to bridge multiple source domains by introducing a continuous sequence of intermediate domains and thus be able to capture any test domain that lies between the source domains. More recently, \citet{lambert2020mseg} constructed a composite semantic segmentation dataset from multiple sources to improve the zero-shot generalization ability to unseen domains of semantic segmentation models. 

\vspace{-0.5em}
\subsection{Data Sampling and Augmentation} 
% \vspace{-0.3em}

The quality of training data is an important factor for machine learning with big data \cite{l2017machine}. 
In this section, we review a collection of algorithmic techniques for data sampling, cleansing, and augmentation for improved and robust training.

\vspace{-0.3em}
\paragraph{Active Learning} Active learning is a framework solution to improve training set quality and minimize data labeling costs by selectively labeling valuable samples (i.e., ``edge cases'') from an unlabeled pool. 
The sample acquisition function in active learning frameworks often relies on prediction uncertainty such as Bayesian estimation \cite{gal2017deep} and ensemble-based \cite{beluch2018power} uncertainties. 
% However, due to their high computational cost, other techniques such as ensemble of multiple models more efficient and successful approaches \cite{beluch2018power}.
Differently, ViewAL \cite{siddiqui2020viewal} actively samples hard-cases by measuring prediction inconsistency across different viewpoints for multi-view semantic segmentation tasks. 
\citet{haussmann2020scalable} presents a scalable active learning framework for open-world applications in which both the target and environment can greatly affect model predictions.  
Active learning techniques can also improve training data quality by identifying noisy labels with minimum crowdsourcing overhead \cite{bouguelia2018agreeing}.

\vspace{-0.3em}
\paragraph{Hardness Weighted Sampling} 
A typical data sampling strategy to improve model robustness is to assign larger weights (in the training loss function) to harder training samples \cite{katharopoulos2018not,zhang2020geometry,fidon2020distributionally}.
For instance, \citet{katharopoulos2018not} achieved this by estimating sample hardness by an efficient upper bound of their gradient norms.
\citet{zhang2020geometry} estimated sample hardness by their distance to the classification boundary and assigning larger weights to hard samples. 
\citet{fidon2020distributionally} performed weight sampling by modeling sample uncertainty through minimizing the worst-case expected loss over an uncertainty set of training data distributions.

\vspace{-0.3em}
\paragraph{Data Cleansing} 
Label errors on large-scale annotated training sets induce noisy or even incorrect supervision during model training and evaluation \cite{yun2021re,beyer2020we,wang2020going}. 
A robust learning strategy against label noise is to select training samples with small loss values to update the model, since samples with label noise usually have larger loss values than clean samples \cite{han2018co,yu2019does,wei2020combating}.
ImageNet is commonly used as a representative case study for label noise. Specifically, many images in ImageNet contain multiple objects, while only one of them is labeled as the ground-truth classification target. 
Efforts have been made to correct such label noise on both the training \cite{yun2021re} and validation \cite{beyer2020we} sets of ImageNet, by providing localized annotations for each object in the image using machine or human annotators.
Training on the new cleansed annotations improved both in-distribution accuracy and robustness \cite{yun2021re}.

\vspace{-0.3em}
\paragraph{Data Augmentation}
\label{par:data-augmentation}
Data augmentations are widely used to improve model generalizability and robustness due to their effectiveness and simplicity.
The most popular strategy of data augmentation is to increase the diversity of training samples, for example, by random rotation and scaling, random color jittering, random patch erasing \cite{zhong2017random}, and many others \cite{cubuk2018autoaugment,cubuk2020randaugment,yun2019cutmix,hendrycks2019augmix,hendrycks2020many}. 
% Recently, style transfer augmentations \cite{geirhos2018imagenet} have been shown to improve model robustness against texture bias. 
AugMix \cite{hendrycks2019augmix} utilizes diverse random augmentations by mixing multiple augmented images, significantly improving model robustness against natural visual corruptions.
CutMix \cite{yun2019cutmix} removes image patches and replaces the removed regions with a patch from another image, where the new ground truth labels are also mixed proportionally to the number of pixels of combined images. 
% CutMix is shown to improve the model robustness against input corruptions and its out-of-distribution detection performance. 
DeepAugment \cite{hendrycks2020many} increases robustness to cross-domain shifts by employing image-to-image translation networks for data augmentation rather than conventional data-independent pixel-domain augmentation.
The second type of data augmentation strategy is adversarial data augmentation \cite{volpi2018generalizing,zhao2020maximum,xie2020adversarial,zhang2019dada}, where fictitious target distributions are generated adversarially to resemble ``worst-case" unforeseen data shifts throughout training. 
For example, \citet{xie2020adversarial} observed that adversarial samples could be used as data augmentation to improve both accuracy and robustness with the help of a novel batch normalization layer.
Recently, \citet{gong2020maxup} and \citet{wang2021augmax} proposed MaxUp and AugMax to combine diversity and adversity in data augmentation, further boosting model generalizability and robustness.

\vspace{-0.5em}
\subsection{Challenges and Opportunities}

% ---- (1) 
Despite advances in defending different types of naturally occurring distributional shifts and synthetic adversarial attacks, there lacks systematic efforts to tackle robustness limitations in a unified framework to cover the ``in-between" cases within this spectrum. 
In fact, in many cases, techniques proposed to enhance one type of robustness do not translate to benefiting other types of robustness. 
For example, \citet{li2020shape} showed that top-performing robust training methods for one type of distribution shift may even harm the robustness on other different distribution shifts.

% ---- (2) 
Another less explored direction for ML robustness is to benefit from multi-domain and multi-source training data for improved representation learning. 
The rich contexts captured from sensor sets with diverse orientations and data modality may improve prediction robustness compared to a single input source (e.g., single camera). % against distribution shifts and artifacts.
For example, a recent paper \cite{fort2021exploring} showed that large models trained on multi-modality data, such as CLIP \cite{radford2021learning}, can significantly improve representation learning to detect domain shift. 
Based on the above finding, a promising direction for future research is to design multi-modality training methods which explicitly encourage model robustness. 
Another under-exploited approach for model robustness is run-time self-checking based on various temporal and semantic coherence of data. 

% ---- (3) 
Faithful and effective evaluation of model robustness is another open challenge in real-world applications. 
Traditional evaluation approaches are designed based on the availability of labeled test sets on the target domain. 
However, in a real-world setting, the target domain may be constantly shifting and making the test data collection inefficient and inaccurate. 
To address this issue, recent work propose more practical settings to evaluate model robustness with only unlabeled test data \cite{garg2022leveraging} or selective data labeling \cite{wang2020going}. 

% ---- (3) 
Unlike training datasets and evaluation benchmarks commonly used in research, a safety-aware training set requires extensive data capturing, cleaning, and labeling to increase the coverage of unknown edge cases by collecting them directly from the open-world. 
Technique like active learning \cite{meng2021towards}, object re-sampling \cite{chang2021image}, and self-labeling allow for efficient and targeted dataset improvements which can directly translate to model performance improvements. 
Generative Adversarial Networks (GAN) could be an underway trend for generating effective large-scale vision datasets. 
For example, \citet{zhang2021datasetgan} propose DatasetGAN, an automatic procedure to generate realistic image datasets with semantically segmented labels. 

\vspace{-0.5em}
\section{Run-time Error Detection} 
\label{sec:error-detection}

The third strategy for ML safety is to detect model errors at run-time. 
Although the robust training methods discussed in Section \ref{sec:RT} can significantly improve model robustness, they cannot entirely prevent run-time errors. 
As a result, run-time monitoring to detect any potential prediction errors is necessary from the safety standpoint. 
\textit{Selective prediction}, also known as prediction with reject option, is the main approach for run-time error detection \cite{geifman2017selective,geifman2019selectivenet}.
Specifically, it requires the model to cautiously provide predictions only when they have high confidence in the test samples.
Otherwise, when the model detects potential anomalies, it will trigger fail-safe plans to prevent system failure.
Selective prediction can significantly improve model robustness at the cost of test set coverage. 
% Recent examples \cite{geifman2017selective,geifman2019selectivenet} present simple and effective implementations which guarantee control over the true risk when running on in-domain test samples. 
In this section, we first review methods for model calibration and uncertainty quantification (Sec. \ref{sec:uncertainty-estimation}) and then go over technique to adopt such methods on specific application scenarios: out-of-distribution detection (Sec. \ref{sec:ood-detection}) and adversarial attack detection (Sec. \ref{sec:adv-detection}). 
% Note that although these techniques have overlapping contributions, we separate them by their targeted error types and discuss their relationships and limitations. 

% \vspace{-0.3em}
% \subsection{Model Calibration and Uncertainty Quantification}
\subsection{Prediction Uncertainty}
\label{sec:uncertainty-estimation}

Prediction uncertainty is the key to enabling selective prediction.
The most intuitive uncertainty measure in DNN models is the softmax probability of the predicted class (also known as maximum softmax probability, or MSP) as used in \cite{hendrycks2016baseline,guo2017calibration}. 
However, DNNs are widely known to be susceptible to overconfidence in their prediction in which the MSP of misclassified samples could be as high as the correct predictions, making MSP a poor measure for prediction confidence \cite{guo2017calibration,hein2019relu}.
Here we review two commonly used solutions for models' overconfidence predictions.

% \vspace{-0.3em}
\subsubsection{Model Calibration}
\label{sec:calibration}
Model calibration aims to design robust training methods so that the MSP of the resultant models be aligned with the likelihood of correct prediction. 
Temperature scaling \citet{guo2017calibration} is arguably the most simple model calibration method, without the need to retrain the existing poor-calibrated model.
Specifically, it softens the softmax probability by scaling down the logits by a factor of $T (T>1)$ during test time. 
This technique was later found also very effective in alleviating overconfidence on test samples from a different distribution \cite{li2020improving}. 
Considering that temperature scaling may undesirably clamp down legitimate high confidence predictions, \citet{kumar2018trainable} proposed maximum mean calibration error (MMCE) which is a trainable calibration measure based on a reproducible kernel Hilbert space (RKHS), and is minimized alongside the NLL loss during training. 
Label smoothing~\cite{szegedy2016rethinking} is another simple and effective technique for model calibration by training the model with more uniform target distribution in cross-entropy loss instead of the traditional one-hot labels. It not only gives more calibrated outputs but also leads to improved network generalization. 
Training with MixUp data augmentation \cite{zhang2017mixup} is also found to benefit model calibration \cite{thulasidasan2019mixup}. 
\citet{zhang2019confidence} used structured dropout to promote model diversity and improve model calibration.
Unlike previous implicit regularization methods, \citet{moon2020confidence} proposed a correctness ranking loss to explicitly encourage model calibration in training.
A concurrent work \cite{krishnan2020improving} proposed another explicit model calibration loss function, termed accuracy versus uncertainty calibration (AvUC) loss, for Bayesian neural networks.
Recently, \citet{karandikar2021soft} proposed a softened version of AvUC, termed S-AvUC, together with another soft calibration loss function termed SB-ECE, which are applicable under the more general non-Bayesian network setting and outperforms previous methods.

% \vspace{-0.3em}
\subsubsection{Uncertainty Quantification (UQ)}
Uncertainty quantification aims to design accurate uncertainty or prediction confidence measures for ML models.
% There have been rich studies of uncertainty quantification in many fields, ranging from science and engineering to many real-world applications. 
The sources of uncertainty can be categorized into two types: \textit{aleatoric} uncertainty and \textit{epistemic} uncertainty \cite{der2009aleatory}. 
% Such categorization and definitions are also inherited in deep learning \cite{gal2016uncertainty}, where Bayesian learning has been a classical tool to quantify uncertainties. 
\citet{gal2016uncertainty} first considered modeling aleatoric uncertainties in deep neural networks following a Bayesian modeling framework where one assumes some prior distribution over the space of parameters. 
% The author described an instantiation of this Bayesian modeling framework using Bayesian Neural Networks (BNNs) as well as approximate inference techniques with variational inference. 
The authors derived a practical approximate UQ measure, which essentially equals to the prediction variance of an ensemble of models generated by statistical regularization techniques such as dropout~\cite{srivastava2014dropout} and other more advanced variants~\cite{gal2016dropout,gal2017concrete}. 
\citet{kendall2017uncertainties} further proposed a Bayesian framework that jointly models both aleatoric and epistemic uncertainties. 
Deep ensembling~\cite{lakshminarayanan2017simple,ovadia2019can} have also been popular approaches for UQ.
The common disadvantage of the above methods is additional computation at run-time. 
To overcome such high computation cost in UQ, approaches with single deterministic models have been proposed \cite{chen2020angular,liu2020simple,van2020uncertainty,mukhoti2021deterministic}.  
For instance, Chen et al.~\cite{chen2020angular} propose an angular visual hardness (AVH) distance-based measure which shows a good correlation to human perception of visual hardness. 
% AVH shows the underlying connection to aleatoric uncertainty and significantly improves tasks such as self-training. 
AVH can be computed using regular training with softmax cross-entropy loss, making it a convenient drop-in replacement for MSP as the uncertainty measure. 
% Several other works also improve uncertainty estimation and its computation cost by using only a single model and employing RBF networks \cite{liu2020simple}, distance training \cite{van2020uncertainty}, and inductive biases \cite{mukhoti2021deterministic}.

% other refs: \cite{gal2017deep}
% \TODO{some overlap with discussion section: }
% \cite{loquercio2020general,pires2020towards}
\vspace{-0.5em}
\subsection{Out-of-distribution Detection}
\label{sec:ood-detection}

Out-of-distribution (OOD) or outlier refers to samples that are disjoint from the source training distribution. 
OOD detection is a binary classification task to distinguish OOD samples from in-distribution (ID) samples at test-time. 
Unlike model calibration (Section \ref{sec:calibration}), OOD detection does not require the prediction confidence to align well with the likelihood of correct prediction on ID data. 
In the following, we categorize and review OOD detection techniques into three groups.

\vspace{-0.5em}
\subsubsection{Distance-Based Detection}
Distance-based methods measure the distance between the input sample and source training set in the representation space. 
These techniques involve pre-processing or test-time sampling of the source domain distribution and measuring their averaged distance to the test sample. 
Various distance measures including Mahalanobis distance \cite{lee2018simple}, cosine similarity \cite{techapanurak2020hyperparameter}, and Euclidean distance \cite{pmlr-v119-goyal20c} have been employed.
For example, Ruff et al. \cite{ruff2018deep} present a deep learning one-class classification approach to minimize the representation hypersphere for normal distribution and calculate the detection score by the distance of the outlier sample to the center of the hypersphere. 
They later \cite{ruff2019deep} extended this work using samples labeled as OOD in a semi-supervised manner.  
Bergman and Hoshen \cite{bergman2020classification} presented a technique that learns a feature space such that inter-class distance is larger than the intra-class distance.
Sohn et al. \cite{sohn2020learning} presented a two-stage one-class classification framework that leverages self-supervision and a shallow one-class classifier.
The OOD detection performance of distance-based methods can be improved by ensembling measurements over multiple input augmentations \cite{tack2020csi} and network layers \cite{lee2018simple,sastry2019detecting}. 
Sastry and Oore \cite{sastry2019detecting} used gram matrices to compute pairwise feature correlations between channels of each layer and identify anomalies by comparing inputs values with its respective range observed over the training data. 
% Tack et al. \cite{tack2020csi} applied contrastive learning for OOD detection by ensembling over random augmentations to improve OOD detection performance. 

\vspace{-0.5em}
\subsubsection{Classification-Based Detection}
Classification-based detection techniques seek effective representation learning to encode normality together with OOD detection scores. 
Various OOD detection scores have been proposed including maximum softmax probability~\cite{hendrycks2016baseline}, prediction entropy~\cite{hendrycks2018deep}, KL-divergence and Jensen-Shannon divergence \cite{Hendrycks2019sv} from uniform distribution as detection score.  
% For example, a simple baseline approach for classification-based OOD detection is to use class probabilities as a measure for OOD detection~\cite{hendrycks2016baseline}. 
Further, Lee et al.~\cite{lee2017training} and Hsu et al. \cite{hsu2020generalized} proposed a combination of temperature scaling and adversarial input perturbations to calibrate the model for better OOD detection.
Another line of research proposes using disjoint unlabeled OOD training set to learn normality and hence improve OOD detection. 
Hendrycs el al.~\cite{hendrycks2018deep} present a case for joint training of natural outlier set (from any auxiliary disjoint training set) with the normal training set resulting in fast and memory efficient OOD detection with minimal architectural changes. 
Later, Mohseni et al.~\cite{mohseni2020ood} show this type of training can be more improved by using additional reject classes in the last layer. 
Other classification-based techniques include revising network architecture for learning better prediction confidence during the training \cite{devries2018learning,yu2019unsupervised}.
From a different perspective, Vyas et al.~\cite{vyas2018out} present a framework for employing an ensemble of classifiers that each leave out a subset of the training set as OOD examples and the rest as the normal in-distribution training set. % They combine ensemble output to obtain OOD detection score which could exceed detecting using a single model. 
% Yu and Aizawa \cite{yu2019unsupervised} present a two-head network to maximize prediction disagreements for outlier samples. 
Recent work show that self-supervised learning can further improve OOD detection and surpass prior techniques \cite{golan2018deep,Hendrycks2019sv,winkens2020contrastive,tack2020csi,sehwag2021ssd,mohseni2021multitask}.
For instance, Tack et al.~\cite{tack2020csi} propose using geometric transformations like rotation to shift different samples further away to improve OOD detection performance. 
% However, contrastive learning techniques are sensitive to the choice of data transformations, training source and target OOD distributions. 

\vspace{-0.5em}
\subsubsection{Density-based Detection}
Using density estimates from Deep Generative Models (DGM) is another line of work to detect OOD samples by creating a probability density function from the source distribution. 
% Recent work on Variational Autoencoders (VAE) and Generative Adversarial Networks (GAN) \cite{schlegl2017unsupervised} shows these models often assign higher likelihoods to samples with high similarity to the training set and low likelihood to outliers with significantly different semantics. 
Different scores based on likelihood are proposed for OOD detection \cite{ren2019likelihood} using GANs and based on reconstruction error\cite{zong2018deep} using VAEs. % removed akcay2018ganomaly
However, some recent studies present counterintuitive results that challenge the validity of VAEs and DGMs likelihood ratios for semantic OOD detection \cite{wang2020further} in high-dimensional data and propose ways to improve likelihood-based score such as using natural perturbation \cite{Choi2020Novelty} % removed nalisnick2018deep
For instance, Serr{\`a} et al. \cite{serra2019input} connect the limitations of generative models' likelihood score for OOD detection with the input complexity and use an estimate of input complexity to derive a new efficient detection score.
Energy-based models (EBMs) are another family of DGMs that has shown higher performance in OOD detection. 
\citet{du2019implicit} present EBMs for OOD detection on high-dimensional data and investigate limitations of reconstruction error on VAEs compared to their proposed energy score. 
% A recent line of work reinterprets standard discriminative classifiers of $p(y|x)$ as an energy-based model for the joint distribution $p(x, y)$. 
% For instance, Grathwohl et al.~\cite{grathwohl2019your} present a joint energy-based model (JEM), an architecture that uses logits of a supervised classifier to define the joint density of training points.
\citet{liu2020energy} propose another energy-based framework for model training with a new OOD detection score based on discriminative models.

\vspace{-0.5em}
\subsection{Adversarial Detection and Guards} 
\label{sec:adv-detection}
% ------ test-time defense  ---------

Adversarial detection and adversarial guards are test-time methods to mitigate the risk of adversarial attacks. 
Adversarial detection refers to designing a detector to identify adversarial perturbations, while in adversarial guarding is done based on removing the effects of adversarial perturbations from a given image sample.
Note that neither of these approaches manipulates model parameters or model training process; therefore, these solutions are complementary to adversarial training solutions reviewed in Section \ref{sec:AT}.

\vspace{-0.3em}
\subsubsection{Adversarial Attack Detection}

The most straightforward way to detect adversarial examples is to train a {\em secondary model} as the adversarial sample detector. 
For example, \citet{grosse2017statistical} and \citet{gong2017adversarial} both trained a binary classifier on clean and adversarial training samples as the adversarial sample detector. 
% Instead of using the adversarial data to train the detector model, \citet{gong2017adversarial} leverage the inner convolutions layers of the original model as the input data and train the detector based on them.
% Although such methods have shown promising results in their papers to detect adversarial examples; unfortunately, these methods could be viewed as one-step detection strategies and could be attacked again by adaptive adversarial examples. They are not very effective. 
% Different from training a separate classifier with high dimensional input to distinguish the adversarial examples, 
Besides raw pixel values, adversarial and clean examples have some different intrinsic properties which can be used to detect adversarial examples.  
For example, adversarial images have greater variances in low-ranked principal components \cite{hendrycks2016early} and larger local intrinsic dimensionality (LID) \citet{ma2018characterizing} than clean images.
% For instance,  \citet{hendrycks2016early} apply Principal Component Analysis (PCA) to the natural and adversarial images directly. 
% They find that the adversarial images have consistently greater variances for low-ranked principal components than clean images. 
% Specifically, the coefficients of an adversarial image have consistently greater variances for the later principal components. 
% Therefore, their technique uses the coefficient variance as a sole feature to detect adversarial examples.
% In another work, \citet{ma2018characterizing} show that Local Intrinsic Dimensionality (LID) which assesses the space-filling capability of the region surrounding a reference example, is significantly larger for adversarial examples than normal images. 
% Based on these analysis, they propose using LID as a feature to detect adversarial examples. 
% Rather than exploring new intrinsic features of adversarial examples, 
% MagNet \cite{meng2017magnet} uses the distance between the test-time inputs and training data manifold for detecting adversarial examples. 
% They distinguish the distance between the input and the data manifold via existing distance functions to detect the adversarial examples. 

% \paragraph{Statistical Test}
{\em Statistical Testing} 
% is another way to distinguish the adversarial examples and natural examples via statistical methods.
% This approach 
utilizes the difference in distribution of adversarial examples and natural clean examples for adversarial detection.
For example, \citet{grosse2017statistical} use the Maximum Mean Discrepancy (MMD) test to evaluate whether the clean images and adversarial examples are from the same distribution. They observe that adversarial examples are located in the different output surface regions compared to clean inputs. 
% Therefore, it could be used to detect adversarial examples. 
Similarly, \citet{feinman2017detecting} leverage the kernel density estimations from the last layer and Bayesian uncertainty estimations from the dropout layer to measure the statistical difference between adversarial examples and normal ones. 
% They show that such statistic difference could be used to detect adversarial examples. 

% \paragraph{Transformation and Randomness} 
{\em Applying Transformation and Randomness} is another approach to detect adversarial examples based on the observation that natural images could be resistant to the transformation or random perturbations while the adversarial ones are not.  
Therefore, one can detect adversarial examples with high accuracy based on the model prediction discrepancy due to applying simple transformations and randomness \cite{li2017adversarial,xu2017feature}. 
% For instance, \citet{feinman2017detecting} propose to add dropout to build up a randomized network. 
% Then, they use uncertainty evaluation metrics to detect the natural and adversarial examples. 
For instance, \citet{li2017adversarial} apply a small mean blur filter to the input image before feeding it into the network. 
They show that natural images are resistant to such transformations while adversarial images are not. 
% Similarly, \citet{xu2017feature} apply the feature squeezing methods such as reducing the color bit and spatial smoothing to inputs.
% They observe that feature squeezing reduces the search space of the adversarial examples to make them coalesce to the original samples. 
% Therefore, the model prediction is different for a given adversarial example before and after feature squeezing. 
% Transformation and randomness could also be used to reduce the adversarial effect of the adversarial examples. 
% Such transformation-based methods are also heavily used as test-time adversarial guards to defend against adversarial examples. We will introduce them in the later subsection. 

\vspace{-0.3em}
\subsubsection{Test-Time Adversarial Guard}
Test-Time adversarial guard aims to accurately classify both adversarial and natural inputs without changing the model parameters. 
Various test-time transformations have been proposed to diminish the adversarial effect of the adversarial perturbations by pre-processing inputs before feeding to the model. % To achieve this goal, 
The research investigates the efficiency of applying different basis transformations to input images, including the  JPEG compression \cite{das2017keeping,guo2017countering,shaham2018defending}, bit-depth reduction, image quilting, total variance minimization \cite{guo2017countering}, low-pass filters, PCA, low-resolution wavelet approximations, and soft-thresholding~\cite{shaham2018defending}. 
% Pixel Deflection \cite{prakash2018deflecting} randomly selects pixels from the test image and replaces their values with other randomly selected pixels within their neighborhood.
% It has been shown that the adversarial effect could be canceled out with these transformations without hurting the benign performance. 
Most of these defense strategies prevent adversarial attacks by obfuscating the gradient propagation in adversarial attacks with non-differential or random operations. 
As a result, they have been shown ineffective under stronger adversarial attacks which bypass such gradient obfuscation \cite{athalye2018obfuscated}. 
% Therefore, these could be used as test-time adversarial guard methods. 

% Different from the above transformation-based methods, a separate denoising network or operation can be used on adversarial images to remove the adversarial effects. 
% For instance, \citet{liang2017enhancing} use a noise reduction method to mitigate the adversarial effects and \citet{samangouei2018defense} use a generative adversarial network trained on clean training samples to ``denoise'' adversarial examples. 
% Additionally, randomness has also been used to map the adversarial examples back to the natural image manifold. 
% For example, \citet{xie2017mitigating} propose to apply randomization operations including random resizing and random padding to input images to remove adversarial effects. 
% Input processing is an effective and relatively computationally-affordable way to

\vspace{-0.5em}
\subsection{Challenges and Opportunities}
% ----- (1)
An open challenge in OOD detection is to improve performance on near-OOD samples that are visually similar to ID samples but yet outliers w.r.t. semantic meanings. 
This scenario is very common in fine-grained image classification and analysis domains where target ID samples could be highly similar to OOD samples. % could be some other species or diseases that are visually very similar with the ones we are interested in.
Recent papers have made attempts in this more challenging scenario \cite{zhang2021fine,fort2021exploring}; however, the OOD detection performance on near-OOD samples is still much worse than that performance on far-OOD samples (i.e., visually more distinct samples). % there is still large room for performance improvement: 

% ---- (2)
Another open research direction is to propose techniques for efficient OOD sample selection and training.
In a recent work, \citet{chen2021atom} present ATOM as an empirically verified technique for mining informative auxiliary OOD training data. 
However, this direction remains under-explored, and many useful measures such as gradient norms \cite{katharopoulos2018not} could be investigated for OOD training efficiency and performance. 
% It is well known that not all ID samples play an equally important role in ID classification. %  \cite{katharopoulos2018not}. 

% ---- (3)
Detecting adversarial examples will remain open research as new attacks are introduced to challenge and defeat detection methods \cite{carlini2017adversarial}.
Given the overhead computational costs for both generating and detecting adversarial samples, an efficient way to nullify attacks could be is to benefit from multi-domain inputs, temporal data characteristics, and domain knowledge from a known clean training set. % A potential future direction is by leveraging domain knowledge and data properties to improve attack detection.
A related example is the work by \citet{xiao2018characterizing} that studies the spatial consistency property in the semantic segmentation task by randomly selecting image patches and cross-checking model predictions among the overlap area. %  for natural images while not for adversarial examples.

\vspace{-0.5em}
\section{Discussion and Conclusions}
\label{sec:discussion}

In our survey, we presented a review of fundamental ML limitations in engineering safety methods; followed by a taxonomy of safety-related techniques in ML. 
The impetus of this work was to leverage from both engineering safety strategies and state-of-the-art ML techniques to enhance the dependability of ML components in autonomous systems. 
Here we summarize key takeaways from our survey and continue with recommendations on each item for future research. % directions that can further advance less studied aspects of ML safety

\vspace{-0.5em}
% \subsection*{T1 \& T2: Engineering Safety Standards to Support ML Products} 
\subsection*{T1: Engineering Standards Can Support ML Product Safety} 

Safety needs for design, development, and deployment of ML learning algorithms have subtle distinctions with code-based software. Our analyses are aligned with prior work and indicate that conventional engineering safety standards are not directly applicable on ML algorithms design. Consequently, relevant industrial safety standards suggest enforcing limitations on operation domain of critical ML functions to minimize potential hazards due to ML malfunction. The limitations enforced on ML functions are due to the lack of ML technology readiness and intended to minimize the risk of hazard to an acceptable level.
Additionally, recent regulations mandate data privacy and algorithmic transparency which in turn could encourage new principles in responsible ML development and deployment pipelines. 
Our recommendation is aligned with safety standards to perform thorough risk and hazard assessments for ML components and limit their functionalities to minimize the risk of failure.

\vspace{-0.5em}
% \subsection*{T2: Establishing Taxonomy of ML Safety Strategies}
\subsection*{T2: The Value of ML Safety Taxonomy}

The main contribution of this paper is to establish a meaningful \textit{ML Safety Taxonomy} based on ML characteristics and limitations to directly benefit from engineering safety practices. 
Specifically, our taxonomy of ML safety techniques maps key engineering safety principles into relevant ML safety strategies to understand and emphasize on the impact of each ML solution on model reliability.
The proposed taxonomy is supported with a comprehensive review of related literature and a hierarchical table of representative papers (Table 1) to categorize ML techniques into three major safety strategies and subsequent solutions.

The benefit of the ML safety taxonomy is to break down the problem space into smaller components, help to lay down a road map for safety needs in ML, and identify plausible future research directions. 
We remark existing challenges and plausible directions as a way to gauge technology readiness on each safety strategy within the main body of literature review in Sections 5, 6, and 7. 
However, given the fast pace of developments in the field of ML, a thorough assessment of technology readiness may not be a one size fits all for ML systems. 
On the other hand, the proposed ML safety taxonomy can benefits from emerging technologies concepts such as Responsible AI \cite{arrieta2020explainable,lu2021software} to take social and legal aspects of safety into account. 
% are existing directions any good? are they mature now? what are the cons and pros? 

\vspace{-0.5em}
% \subsection*{T3: Diversification Strategy for ML Safety} 
\subsection*{T3: Recommendations for Choosing ML Safety Strategies} 
A practical way to improve safety of complex ML products is to benefit from diversification in ML safety strategies and hence minimizing the risk of hazards associated with ML malfunction. We recognize multiple reasons to benefit from diversification of safety strategies. To start with, as no ML solutions guarantees absolute error-less performance in open-world environments, a design based on collection of diverse solutions could learn more complete data representation and hence achieve higher performance and robustness.
In other words, a design based on a collection of diverse solution is more likely to maintain robustness at the time of unforeseen distribution shifts as known as edge-cases.

Additionally, the overlaps and interactions between ML solutions boost overall performance and reduce development costs. 
For instance, scenario-based testing for model validation can directly impact data collection and training set quality in design and development cycles. Another example is the positive effects of transfer learning and domain generalization on uncertainty quantification and OOD detection. Lastly, diverse strategies can be applied on different stages of design, development, and deployment of ML lifecycle which benefits from continues monitoring of ML safety across all ML product teams.
% to ensure continues monitoring and prevention of ML . 
% Hence, diversification can minimize design and development costs. 

\vspace{-0.5em}
% \subsection*{T4: Safe AI Development Frameworks}
\subsection*{T4: Recommendations for Safe AI Development Frameworks}

ML system development tools and platforms (MLOps) aim to automate and unify the design, development, and deployment of ML systems with a collection of best practices. 
Prior work have emphasized on MLOps tools to minimize the development and maintenance costs in large scale ML systems \cite{amershi2019software}. % and maximize the reliability of real-world 
We propose existing and emerging MLOps to support and prioritize adaptation and monitoring of ML safety strategies across both system and software level.
% However, MLOps have the opportunity to significantly aid the safety of ML systems by supporting and prioritizing adaptation of ML safety strategies throughout the product lifecycle. 
A safety-oriented ML lifecycle incorporates all aspects of ML development from constructing safety scope and requirements, to data management, model training and evaluation, and open-world deployment and monitoring \cite{ashmore2021assuring,hawkins2021guidance}. 
Industry-oriented efforts in safety-aware MLOps can unify necessary tools, metrics, and increase accessibility for all AI development teams. % safety checklists and rubrics breck2016s

Recent emerging concepts such as Responsible AI \cite{arrieta2020explainable} and Explainable AI \cite{mohseni2018multidisciplinary} aim for building safe AI systems to ensure data privacy, fairness, and human-centered values in AI development.  % wiens2019no in healthcare  falco2021governing for governane 
These new emerging AI concepts can target beyond functional safety of the intelligent system and help to prevent end-users (e.g., driver in the autonomous vehicle) from unintentional misuse of the system due to over-trusting and user unawareness \cite{mohseni2019practical}.

{\small
\setlength{\bibsep}{0pt}
\bibliographystyle{abbrvnat}
\bibliography{Bibliography}

\begin{thebibliography}{300}
\providecommand{\natexlab}[1]{#1}
\providecommand{\url}[1]{\texttt{#1}}
\expandafter\ifx\csname urlstyle\endcsname\relax
  \providecommand{\doi}[1]{doi: #1}\else
  \providecommand{\doi}{doi: \begingroup \urlstyle{rm}\Url}\fi

\bibitem[Adebayo et~al.(2018)Adebayo, Gilmer, Muelly, Goodfellow, Hardt, and
  Kim]{adebayo2018sanity}
J.~Adebayo, J.~Gilmer, M.~Muelly, I.~Goodfellow, M.~Hardt, and B.~Kim.
\newblock Sanity checks for saliency maps.
\newblock In \emph{NIPS}, 2018.

\bibitem[Amershi et~al.(2019)Amershi, Begel, Bird, DeLine, Gall, Kamar,
  Nagappan, Nushi, and Zimmermann]{amershi2019software}
S.~Amershi, A.~Begel, C.~Bird, R.~DeLine, H.~Gall, E.~Kamar, N.~Nagappan,
  B.~Nushi, and T.~Zimmermann.
\newblock Software engineering for machine learning: A case study.
\newblock In \emph{ICSE-SEIP}. IEEE, 2019.

\bibitem[Amodei et~al.(2016)Amodei, Olah, Steinhardt, Christiano, Schulman, and
  Man{\'e}]{amodei2016concrete}
D.~Amodei, C.~Olah, J.~Steinhardt, P.~Christiano, J.~Schulman, and D.~Man{\'e}.
\newblock Concrete problems in ai safety.
\newblock \emph{arXiv preprint arXiv:1606.06565}, 2016.

\bibitem[Arrieta et~al.(2020)Arrieta, D{\'\i}az-Rodr{\'\i}guez, Del~Ser,
  Bennetot, Tabik, Barbado, Garc{\'\i}a, Gil-L{\'o}pez, Molina, Benjamins,
  et~al.]{arrieta2020explainable}
A.~B. Arrieta, N.~D{\'\i}az-Rodr{\'\i}guez, J.~Del~Ser, A.~Bennetot, S.~Tabik,
  A.~Barbado, S.~Garc{\'\i}a, S.~Gil-L{\'o}pez, D.~Molina, R.~Benjamins, et~al.
\newblock Explainable artificial intelligence (xai): Concepts, taxonomies,
  opportunities and challenges toward responsible ai.
\newblock \emph{Information Fusion}, 58:\penalty0 82--115, 2020.

\bibitem[Ashmore et~al.(2021)Ashmore, Calinescu, and
  Paterson]{ashmore2021assuring}
R.~Ashmore, R.~Calinescu, and C.~Paterson.
\newblock Assuring the machine learning lifecycle: Desiderata, methods, and
  challenges.
\newblock \emph{ACM Computing Surveys (CSUR)}, 2021.

\bibitem[Athalye et~al.(2018)Athalye, Carlini, and
  Wagner]{athalye2018obfuscated}
A.~Athalye, N.~Carlini, and D.~Wagner.
\newblock Obfuscated gradients give a false sense of security: Circumventing
  defenses to adversarial examples.
\newblock In \emph{International Conference on Machine Learning}, pages
  274--283. PMLR, 2018.

\bibitem[Awais et~al.(2021)Awais, Zhou, Xu, Hong, Luo, Bae, and
  Li]{awais2021adversarial}
M.~Awais, F.~Zhou, H.~Xu, L.~Hong, P.~Luo, S.-H. Bae, and Z.~Li.
\newblock Adversarial robustness for unsupervised domain adaptation.
\newblock In \emph{Proceedings of the IEEE/CVF International Conference on
  Computer Vision}, pages 8568--8577, 2021.

\bibitem[Bach et~al.(2015)Bach, Binder, Montavon, Klauschen, M{\"u}ller, and
  Samek]{bach2015pixel}
S.~Bach, A.~Binder, G.~Montavon, F.~Klauschen, K.-R. M{\"u}ller, and W.~Samek.
\newblock On pixel-wise explanations for non-linear classifier decisions by
  layer-wise relevance propagation.
\newblock \emph{PloS one}, 10\penalty0 (7):\penalty0 e0130140, 2015.

\bibitem[Bai et~al.(2021)Bai, Mei, Yuille, and Xie]{bai2021transformers}
Y.~Bai, J.~Mei, A.~L. Yuille, and C.~Xie.
\newblock Are transformers more robust than {CNNs}?
\newblock \emph{NeurIPS}, 2021.

\bibitem[Balaji et~al.(2018)Balaji, Sankaranarayanan, and
  Chellappa]{balaji2018metareg}
Y.~Balaji, S.~Sankaranarayanan, and R.~Chellappa.
\newblock Metareg: Towards domain generalization using meta-regularization.
\newblock In \emph{NeurIPS}, pages 1006--1016, 2018.

\bibitem[Bansal and Weld(2018)]{bansal2018coverage}
G.~Bansal and D.~S. Weld.
\newblock A coverage-based utility model for identifying unknown unknowns.
\newblock In \emph{Thirty-Second AAAI Conference on AI}, 2018.

\bibitem[{Bansal} et~al.(2018){Bansal}, {Chen}, and {Wang}]{bansal2018can}
N.~{Bansal}, X.~{Chen}, and Z.~{Wang}.
\newblock Can we gain more from orthogonality regularizations in training deep
  cnns.
\newblock In \emph{Proceedings of the 32nd International Conference on Neural
  Information Processing Systems}, 2018.

\bibitem[Bartocci et~al.(2018)Bartocci, Deshmukh, Donz{\'e}, Fainekos, Maler,
  Ni{\v{c}}kovi{\'c}, and Sankaranarayanan]{bartocci2018specification}
E.~Bartocci, J.~Deshmukh, A.~Donz{\'e}, G.~Fainekos, O.~Maler,
  D.~Ni{\v{c}}kovi{\'c}, and S.~Sankaranarayanan.
\newblock Specification-based monitoring of cyber-physical systems: a survey on
  theory, tools and applications.
\newblock In \emph{Lectures on Runtime Verification}, pages 135--175. Springer,
  2018.

\bibitem[Beluch et~al.(2018)Beluch, Genewein, N{\"u}rnberger, and
  K{\"o}hler]{beluch2018power}
W.~H. Beluch, T.~Genewein, A.~N{\"u}rnberger, and J.~M. K{\"o}hler.
\newblock The power of ensembles for active learning in image classification.
\newblock In \emph{CVPR}, 2018.

\bibitem[Bergman and Hoshen(2020)]{bergman2020classification}
L.~Bergman and Y.~Hoshen.
\newblock Classification-based anomaly detection for general data.
\newblock \emph{arXiv:2005.02359}, 2020.

\bibitem[Beyer et~al.(2020)Beyer, H{\'e}naff, Kolesnikov, Zhai, and
  Oord]{beyer2020we}
L.~Beyer, O.~J. H{\'e}naff, A.~Kolesnikov, X.~Zhai, and A.~v.~d. Oord.
\newblock Are we done with imagenet?
\newblock \emph{arXiv preprint arXiv:2006.07159}, 2020.

\bibitem[Bhojanapalli et~al.(2021)Bhojanapalli, Chakrabarti, Glasner, Li,
  Unterthiner, and Veit]{bhojanapalli2021understanding}
S.~Bhojanapalli, A.~Chakrabarti, D.~Glasner, D.~Li, T.~Unterthiner, and
  A.~Veit.
\newblock Understanding robustness of transformers for image classification.
\newblock \emph{arXiv preprint arXiv:2103.14586}, 2021.

\bibitem[Bochkovskiy et~al.(2020)Bochkovskiy, Wang, and
  Liao]{bochkovskiy2020yolov4}
A.~Bochkovskiy, C.-Y. Wang, and H.-Y.~M. Liao.
\newblock Yolov4: Optimal speed and accuracy of object detection.
\newblock \emph{arXiv preprint arXiv:2004.10934}, 2020.

\bibitem[Bojarski et~al.(2018)Bojarski, Choromanska, Choromanski, Firner,
  Ackel, Muller, Yeres, and Zieba]{bojarski2018visualbackprop}
M.~Bojarski, A.~Choromanska, K.~Choromanski, B.~Firner, L.~J. Ackel, U.~Muller,
  P.~Yeres, and K.~Zieba.
\newblock Visualbackprop: Efficient visualization of cnns for autonomous
  driving.
\newblock In \emph{ICRA}, 2018.

\bibitem[Bouguelia et~al.(2018)Bouguelia, Nowaczyk, Santosh, and
  Verikas]{bouguelia2018agreeing}
M.-R. Bouguelia, S.~Nowaczyk, K.~Santosh, and A.~Verikas.
\newblock Agreeing to disagree: active learning with noisy labels without
  crowdsourcing.
\newblock \emph{International Journal of Machine Learning and Cybernetics},
  9\penalty0 (8):\penalty0 1307--1319, 2018.

\bibitem[Brockman et~al.(2016)Brockman, Cheung, Pettersson, Schneider,
  Schulman, Tang, and Zaremba]{brockman2016openai}
G.~Brockman, V.~Cheung, L.~Pettersson, J.~Schneider, J.~Schulman, J.~Tang, and
  W.~Zaremba.
\newblock Openai gym.
\newblock \emph{arXiv preprint arXiv:1606.01540}, 2016.

\bibitem[Brown et~al.(2017)Brown, Man{\'e}, Roy, Abadi, and
  Gilmer]{brown2017adversarial}
T.~B. Brown, D.~Man{\'e}, A.~Roy, M.~Abadi, and J.~Gilmer.
\newblock Adversarial patch.
\newblock \emph{arXiv preprint arXiv:1712.09665}, 2017.

\bibitem[Carlini and Wagner(2017)]{carlini2017adversarial}
N.~Carlini and D.~Wagner.
\newblock Adversarial examples are not easily detected: Bypassing ten detection
  methods.
\newblock In \emph{10th ACM Workshop on Artificial Intelligence and Security},
  2017.

\bibitem[Chakraborty et~al.(2014)Chakraborty, Fremont, Meel, Seshia, and
  Vardi]{chakraborty2014distribution}
S.~Chakraborty, D.~Fremont, K.~Meel, S.~Seshia, and M.~Vardi.
\newblock Distribution-aware sampling and weighted model counting for sat.
\newblock In \emph{Proceedings of the AAAI Conference on Artificial
  Intelligence}, volume~28, 2014.

\bibitem[Chan et~al.(2021)Chan, Lis, Uhlemeyer, Blum, Honari, Siegwart, Fua,
  Salzmann, and Rottmann]{chan2021segmentmeifyoucan}
R.~Chan, K.~Lis, S.~Uhlemeyer, H.~Blum, S.~Honari, R.~Siegwart, P.~Fua,
  M.~Salzmann, and M.~Rottmann.
\newblock Segmentmeifyoucan: A benchmark for anomaly segmentation.
\newblock In \emph{Advances in Neural Information Processing Systems}, 2021.

\bibitem[Chang et~al.(2021)Chang, Yu, Wang, Anandkumar, Fidler, and
  Alvarez]{chang2021image}
N.~Chang, Z.~Yu, Y.-X. Wang, A.~Anandkumar, S.~Fidler, and J.~M. Alvarez.
\newblock Image-level or object-level? a tale of two resampling strategies for
  long-tailed detection.
\newblock In \emph{International Conference on Machine Learning}, pages
  1463--1472. PMLR, 2021.

\bibitem[Chen et~al.(2020{\natexlab{a}})Chen, Liu, Yu, Kautz, Shrivastava,
  Garg, and Anandkumar]{chen2020angular}
B.~Chen, W.~Liu, Z.~Yu, J.~Kautz, A.~Shrivastava, A.~Garg, and A.~Anandkumar.
\newblock Angular visual hardness.
\newblock In \emph{ICML}, 2020{\natexlab{a}}.

\bibitem[Chen et~al.(2020{\natexlab{b}})Chen, Zhang, Xue, Gong, Liu, Ji, and
  Doermann]{chen2020anti}
H.~Chen, B.~Zhang, S.~Xue, X.~Gong, H.~Liu, R.~Ji, and D.~Doermann.
\newblock Anti-bandit neural architecture search for model defense.
\newblock In \emph{ECCV}, pages 70--85, 2020{\natexlab{b}}.

\bibitem[Chen et~al.(2021)Chen, Li, Wu, Liang, and Jha]{chen2021atom}
J.~Chen, Y.~Li, X.~Wu, Y.~Liang, and S.~Jha.
\newblock {ATOM}: Robustifying out-of-distribution detection using outlier
  mining.
\newblock In \emph{ECML}, pages 430--445, 2021.

\bibitem[Chen et~al.(2020{\natexlab{c}})Chen, Liu, Chang, Cheng, Amini, and
  Wang]{chen2020adversarial}
T.~Chen, S.~Liu, S.~Chang, Y.~Cheng, L.~Amini, and Z.~Wang.
\newblock Adversarial robustness: From self-supervised pre-training to
  fine-tuning.
\newblock In \emph{Proceedings of the IEEE/CVF Conference on Computer Vision
  and Pattern Recognition}, 2020{\natexlab{c}}.

\bibitem[Choi and Chung(2020)]{Choi2020Novelty}
S.~Choi and S.-Y. Chung.
\newblock Novelty detection via blurring.
\newblock In \emph{International Conference on Learning Representations}, 2020.

\bibitem[Cluzeau et~al.(2020)Cluzeau, Henriquel, Rebender, Soudain, van Dijk,
  Gronskiy, Haber, Perret-Gentil, and Polak]{cluzeau2020concepts}
J.~Cluzeau, X.~Henriquel, G.~Rebender, G.~Soudain, L.~van Dijk, A.~Gronskiy,
  D.~Haber, C.~Perret-Gentil, and R.~Polak.
\newblock Concepts of design assurance for neural networks (codann).
\newblock \emph{Public Report Extract Version 1.0}, 2020.

\bibitem[Cohen et~al.(2019)Cohen, Rosenfeld, and Kolter]{cohen2019certified}
J.~M. Cohen, E.~Rosenfeld, and J.~Z. Kolter.
\newblock Certified adversarial robustness via randomized smoothing.
\newblock \emph{arXiv preprint arXiv:1902.02918}, 2019.

\bibitem[Commission(2000)]{IEC-61508}
I.~E. Commission.
\newblock Functional safety of electrical/electronic/programmable electronic
  safety-related systems.
\newblock Technical report, 2000.

\bibitem[Croce and Hein(2020)]{croce2020reliable}
F.~Croce and M.~Hein.
\newblock Reliable evaluation of adversarial robustness with an ensemble of
  diverse parameter-free attacks.
\newblock \emph{arXiv preprint arXiv:2003.01690}, 2020.

\bibitem[Cubuk et~al.(2018)Cubuk, Zoph, Mane, Vasudevan, and
  Le]{cubuk2018autoaugment}
E.~D. Cubuk, B.~Zoph, D.~Mane, V.~Vasudevan, and Q.~V. Le.
\newblock Autoaugment: Learning augmentation policies from data.
\newblock \emph{arXiv preprint arXiv:1805.09501}, 2018.

\bibitem[Cubuk et~al.(2020)Cubuk, Zoph, Shlens, and Le]{cubuk2020randaugment}
E.~D. Cubuk, B.~Zoph, J.~Shlens, and Q.~V. Le.
\newblock Randaugment: Practical automated data augmentation with a reduced
  search space.
\newblock In \emph{Proceedings of the IEEE/CVF Conference on Computer Vision
  and Pattern Recognition Workshops}, 2020.

\bibitem[Das et~al.(2016)Das, Agrawal, Zitnick, Parikh, and Batra]{VQA-HAT}
A.~Das, H.~Agrawal, C.~L. Zitnick, D.~Parikh, and D.~Batra.
\newblock Human attention in visual question answering: Do humans and deep
  networks look at the same regions?
\newblock In \emph{EMNLP}, 2016.

\bibitem[Das et~al.(2017)Das, Shanbhogue, Chen, Hohman, Chen, Kounavis, and
  Chau]{das2017keeping}
N.~Das, M.~Shanbhogue, S.-T. Chen, F.~Hohman, L.~Chen, M.~E. Kounavis, and
  D.~H. Chau.
\newblock Keeping the bad guys out: Protecting and vaccinating deep learning
  with jpeg compression.
\newblock \emph{arXiv preprint arXiv:1705.02900}, 2017.

\bibitem[Deng et~al.(2009)Deng, Dong, Socher, Li, Li, and
  Fei-Fei]{deng2009imagenet}
J.~Deng, W.~Dong, R.~Socher, L.-J. Li, K.~Li, and L.~Fei-Fei.
\newblock Imagenet: A large-scale hierarchical image database.
\newblock In \emph{2009 IEEE conference on computer vision and pattern
  recognition}, pages 248--255. Ieee, 2009.

\bibitem[Der~Kiureghian and Ditlevsen(2009)]{der2009aleatory}
A.~Der~Kiureghian and O.~Ditlevsen.
\newblock Aleatory or epistemic? does it matter?
\newblock \emph{Structural safety}, 2009.

\bibitem[DeVries and Taylor(2018)]{devries2018learning}
T.~DeVries and G.~W. Taylor.
\newblock Learning confidence for out-of-distribution detection in neural
  networks.
\newblock \emph{ICLR}, 2018.

\bibitem[Ding et~al.(2020)Ding, Sharma, Lui, and Huang]{ding2020mma}
G.~W. Ding, Y.~Sharma, K.~Y.~C. Lui, and R.~Huang.
\newblock {MMA} training: {Direct} input space margin maximization through
  adversarial training.
\newblock In \emph{ICLR}, 2020.

\bibitem[Djolonga et~al.(2020)Djolonga, Yung, Tschannen, Romijnders, Beyer,
  Kolesnikov, Puigcerver, Minderer, D'Amour, Moldovan,
  et~al.]{djolonga2020robustness}
J.~Djolonga, J.~Yung, M.~Tschannen, R.~Romijnders, L.~Beyer, A.~Kolesnikov,
  J.~Puigcerver, M.~Minderer, A.~D'Amour, D.~Moldovan, et~al.
\newblock On robustness and transferability of convolutional neural networks.
\newblock \emph{arXiv:2007.08558}, 2020.

\bibitem[Doshi-Velez and Kim(2017)]{doshi2017towards}
F.~Doshi-Velez and B.~Kim.
\newblock Towards a rigorous science of interpretable machine learning.
\newblock \emph{arXiv preprint arXiv:1702.08608}, 2017.

\bibitem[Dreossi et~al.(2018)Dreossi, Ghosh, Yue, Keutzer,
  Sangiovanni-Vincentelli, and Seshia]{Dreossi2018aug}
T.~Dreossi, S.~Ghosh, X.~Yue, K.~Keutzer, A.~Sangiovanni-Vincentelli, and S.~A.
  Seshia.
\newblock Counterexample-guided data augmentation.
\newblock In \emph{Proceedings of the Twenty-Seventh International Joint
  Conference on Artificial Intelligence}, 2018.

\bibitem[Dreossi et~al.(2019{\natexlab{a}})Dreossi, Donz{\'e}, and
  Seshia]{dreossi2019compositional}
T.~Dreossi, A.~Donz{\'e}, and S.~A. Seshia.
\newblock Compositional falsification of cyber-physical systems with machine
  learning components.
\newblock \emph{Journal of Automated Reasoning}, 63\penalty0 (4):\penalty0
  1031--1053, 2019{\natexlab{a}}.

\bibitem[Dreossi et~al.(2019{\natexlab{b}})Dreossi, Fremont, Ghosh, Kim,
  Ravanbakhsh, Vazquez-Chanlatte, and Seshia]{dreossi2019verifai}
T.~Dreossi, D.~J. Fremont, S.~Ghosh, E.~Kim, H.~Ravanbakhsh,
  M.~Vazquez-Chanlatte, and S.~A. Seshia.
\newblock Verifai: A toolkit for the formal design and analysis of artificial
  intelligence-based systems.
\newblock In \emph{International Conference on Computer Aided Verification},
  pages 432--442. Springer, 2019{\natexlab{b}}.

\bibitem[Dreossi et~al.(2019{\natexlab{c}})Dreossi, Ghosh,
  Sangiovanni-Vincentelli, and Seshia]{dreossi2019formalization}
T.~Dreossi, S.~Ghosh, A.~Sangiovanni-Vincentelli, and S.~A. Seshia.
\newblock A formalization of robustness for deep neural networks.
\newblock \emph{arXiv preprint arXiv:1903.10033}, 2019{\natexlab{c}}.

\bibitem[Du and Mordatch(2019)]{du2019implicit}
Y.~Du and I.~Mordatch.
\newblock Implicit generation and modeling with energy based models.
\newblock In \emph{Advances in Neural Information Processing Systems}, pages
  3608--3618, 2019.

\bibitem[Dutta et~al.()Dutta, Jha, Sankaranarayanan, and
  Tiwari]{dutta2017output}
S.~Dutta, S.~Jha, S.~Sankaranarayanan, and A.~Tiwari.
\newblock Output range analysis for deep feedforward neural networks.
\newblock \emph{NASA Formal Methods}, LNCS 10811.

\bibitem[Dvijotham et~al.(2018)Dvijotham, Gowal, Stanforth, Arandjelovic,
  O'Donoghue, Uesato, and Kohli]{dvijotham2018training}
K.~Dvijotham, S.~Gowal, R.~Stanforth, R.~Arandjelovic, B.~O'Donoghue,
  J.~Uesato, and P.~Kohli.
\newblock Training verified learners with learned verifiers.
\newblock \emph{arXiv preprint arXiv:1805.10265}, 2018.

\bibitem[Dvijotham et~al.()Dvijotham, Stanforth, Gowal, Qin, De, and
  Kohli]{dvijothamefficient2019}
K.~D. Dvijotham, R.~Stanforth, S.~Gowal, C.~Qin, S.~De, and P.~Kohli.
\newblock Efficient neural network verification with exactness
  characterization.

\bibitem[Feinman et~al.(2017)Feinman, Curtin, Shintre, and
  Gardner]{feinman2017detecting}
R.~Feinman, R.~R. Curtin, S.~Shintre, and A.~B. Gardner.
\newblock Detecting adversarial samples from artifacts.
\newblock \emph{arXiv preprint arXiv:1703.00410}, 2017.

\bibitem[Fidon et~al.(2020)Fidon, Ourselin, and
  Vercauteren]{fidon2020distributionally}
L.~Fidon, S.~Ourselin, and T.~Vercauteren.
\newblock Distributionally robust deep learning using hardness weighted
  sampling.
\newblock \emph{arXiv preprint arXiv:2001.02658}, 2020.

\bibitem[for Standardization(2011)]{ISO-26262}
I.~O. for Standardization.
\newblock Iso 26262: Road vehicles – functional safety.
\newblock Technical report, 2011.

\bibitem[for Standardization(2019)]{ISO/PAS-21448}
I.~O. for Standardization.
\newblock Iso/pas 21448:: Road vehicles — safety of the intended
  functionality.
\newblock Technical report, 2019.

\bibitem[Fort et~al.(2021)Fort, Ren, and Lakshminarayanan]{fort2021exploring}
S.~Fort, J.~Ren, and B.~Lakshminarayanan.
\newblock Exploring the limits of out-of-distribution detection.
\newblock In \emph{NeurIPS}, 2021.

\bibitem[Fremont et~al.(2020)Fremont, Kim, Pant, Seshia, Acharya, Bruso, Wells,
  Lemke, Lu, and Mehta]{fremont2020formal}
D.~J. Fremont, E.~Kim, Y.~V. Pant, S.~A. Seshia, A.~Acharya, X.~Bruso,
  P.~Wells, S.~Lemke, Q.~Lu, and S.~Mehta.
\newblock Formal scenario-based testing of autonomous vehicles: From simulation
  to the real world.
\newblock In \emph{2020 IEEE 23rd International Conference on Intelligent
  Transportation Systems (ITSC)}, pages 1--8. IEEE, 2020.

\bibitem[Gal(2016)]{gal2016uncertainty}
Y.~Gal.
\newblock \emph{Uncertainty in deep learning}.
\newblock PhD thesis, PhD thesis, University of Cambridge, 2016.

\bibitem[Gal and Ghahramani(2016)]{gal2016dropout}
Y.~Gal and Z.~Ghahramani.
\newblock Dropout as a bayesian approximation: Representing model uncertainty
  in deep learning.
\newblock In \emph{ICML}, 2016.

\bibitem[Gal et~al.(2017{\natexlab{a}})Gal, Hron, and Kendall]{gal2017concrete}
Y.~Gal, J.~Hron, and A.~Kendall.
\newblock Concrete dropout.
\newblock In \emph{NeurIPS}, 2017{\natexlab{a}}.

\bibitem[Gal et~al.(2017{\natexlab{b}})Gal, Islam, and Ghahramani]{gal2017deep}
Y.~Gal, R.~Islam, and Z.~Ghahramani.
\newblock Deep bayesian active learning with image data.
\newblock In \emph{International Conference on Machine Learning}, pages
  1183--1192. PMLR, 2017{\natexlab{b}}.

\bibitem[Ganin and Lempitsky(2014)]{ganin2014unsupervised}
Y.~Ganin and V.~Lempitsky.
\newblock Unsupervised domain adaptation by backpropagation.
\newblock \emph{arXiv:1409.7495}, 2014.

\bibitem[Gao et~al.(2019)Gao, Cai, Li, Hsieh, Wang, and
  Lee]{gao2019convergence}
R.~Gao, T.~Cai, H.~Li, C.-J. Hsieh, L.~Wang, and J.~D. Lee.
\newblock Convergence of adversarial training in overparametrized neural
  networks.
\newblock \emph{Advances in Neural Information Processing Systems},
  32:\penalty0 13029--13040, 2019.

\bibitem[Garg et~al.(2022)Garg, Balakrishnan, Lipton, Neyshabur, and
  Sedghi]{garg2022leveraging}
S.~Garg, S.~Balakrishnan, Z.~C. Lipton, B.~Neyshabur, and H.~Sedghi.
\newblock Leveraging unlabeled data to predict out-of-distribution performance.
\newblock In \emph{ICLR}, 2022.

\bibitem[Geifman and El-Yaniv(2017)]{geifman2017selective}
Y.~Geifman and R.~El-Yaniv.
\newblock Selective classification for deep neural networks.
\newblock In \emph{NIPS}, pages 4878--4887, 2017.

\bibitem[Geifman and El-Yaniv(2019)]{geifman2019selectivenet}
Y.~Geifman and R.~El-Yaniv.
\newblock Selectivenet: A deep neural network with an integrated reject option.
\newblock \emph{arXiv preprint arXiv:1901.09192}, 2019.

\bibitem[Geirhos et~al.(2019)Geirhos, Rubisch, Michaelis, Bethge, Wichmann, and
  Brendel]{geirhos2018imagenet}
R.~Geirhos, P.~Rubisch, C.~Michaelis, M.~Bethge, F.~A. Wichmann, and
  W.~Brendel.
\newblock Imagenet-trained cnns are biased towards texture; increasing shape
  bias improves accuracy and robustness.
\newblock \emph{ICLR}, 2019.

\bibitem[Ghorbani et~al.(2019)Ghorbani, Wexler, Zou, and
  Kim]{NEURIPS2019_77d2afcb}
A.~Ghorbani, J.~Wexler, J.~Y. Zou, and B.~Kim.
\newblock Towards automatic concept-based explanations.
\newblock In \emph{Advances in Neural Information Processing Systems},
  volume~32, pages 9277--9286, 2019.

\bibitem[Golan and El-Yaniv(2018)]{golan2018deep}
I.~Golan and R.~El-Yaniv.
\newblock Deep anomaly detection using geometric transformations.
\newblock In \emph{NIPS}, pages 9758--9769, 2018.

\bibitem[Gong et~al.(2021)Gong, Ren, Ye, and Liu]{gong2020maxup}
C.~Gong, T.~Ren, M.~Ye, and Q.~Liu.
\newblock {MaxUp}: A simple way to improve generalization of neural network
  training.
\newblock In \emph{CVPR}, 2021.

\bibitem[Gong et~al.(2019)Gong, Li, Chen, and Gool]{gong2019dlow}
R.~Gong, W.~Li, Y.~Chen, and L.~V. Gool.
\newblock {DLOW}: Domain flow for adaptation and generalization.
\newblock In \emph{CVPR}, pages 2477--2486, 2019.

\bibitem[Gong et~al.(2017)Gong, Wang, and Ku]{gong2017adversarial}
Z.~Gong, W.~Wang, and W.-S. Ku.
\newblock Adversarial and clean data are not twins.
\newblock \emph{arXiv preprint arXiv:1704.04960}, 2017.

\bibitem[Goodfellow et~al.(2015)Goodfellow, Shlens, and
  Szegedy]{goodfellow2014explaining}
I.~J. Goodfellow, J.~Shlens, and C.~Szegedy.
\newblock Explaining and harnessing adversarial examples.
\newblock \emph{ICLR}, 2015.

\bibitem[Goodman and Flaxman(2017)]{goodman2017european}
B.~Goodman and S.~Flaxman.
\newblock European union regulations on algorithmic decision-making and a
  “right to explanation”.
\newblock \emph{AI magazine}, 38\penalty0 (3):\penalty0 50--57, 2017.

\bibitem[Gowal et~al.(2018)Gowal, Dvijotham, Stanforth, Bunel, Qin, Uesato,
  Mann, and Kohli]{gowal2018effectiveness}
S.~Gowal, K.~Dvijotham, R.~Stanforth, R.~Bunel, C.~Qin, J.~Uesato, T.~Mann, and
  P.~Kohli.
\newblock On the effectiveness of interval bound propagation for training
  verifiably robust models.
\newblock \emph{arXiv preprint arXiv:1810.12715}, 2018.

\bibitem[Goyal et~al.(2020)Goyal, Raghunathan, Jain, Simhadri, and
  Jain]{pmlr-v119-goyal20c}
S.~Goyal, A.~Raghunathan, M.~Jain, H.~V. Simhadri, and P.~Jain.
\newblock {DROCC}: Deep robust one-class classification.
\newblock In \emph{Proceedings of the 37th International Conference on Machine
  Learning}, 2020.

\bibitem[Grathwohl et~al.(2019)Grathwohl, Wang, Jacobsen, Duvenaud, Norouzi,
  and Swersky]{grathwohl2019your}
W.~Grathwohl, K.-C. Wang, J.-H. Jacobsen, D.~Duvenaud, M.~Norouzi, and
  K.~Swersky.
\newblock Your classifier is secretly an energy based model and you should
  treat it like one.
\newblock In \emph{International Conference on Learning Representations}, 2019.

\bibitem[Grosse et~al.(2017)Grosse, Manoharan, Papernot, Backes, and
  McDaniel]{grosse2017statistical}
K.~Grosse, P.~Manoharan, N.~Papernot, M.~Backes, and P.~McDaniel.
\newblock On the (statistical) detection of adversarial examples.
\newblock \emph{arXiv preprint arXiv:1702.06280}, 2017.

\bibitem[Gui et~al.(2019)Gui, Wang, Yang, Yu, Wang, and Liu]{gui2019model}
S.~Gui, H.~Wang, H.~Yang, C.~Yu, Z.~Wang, and J.~Liu.
\newblock Model compression with adversarial robustness: A unified optimization
  framework.
\newblock 2019.

\bibitem[Guidotti et~al.(2018)Guidotti, Monreale, Ruggieri, Pedreschi, Turini,
  and Giannotti]{guidotti2018local}
R.~Guidotti, A.~Monreale, S.~Ruggieri, D.~Pedreschi, F.~Turini, and
  F.~Giannotti.
\newblock Local rule-based explanations of black box decision systems.
\newblock \emph{arXiv preprint arXiv:1805.10820}, 2018.

\bibitem[Guo et~al.(2017)Guo, Pleiss, Sun, and Weinberger]{guo2017calibration}
C.~Guo, G.~Pleiss, Y.~Sun, and K.~Q. Weinberger.
\newblock On calibration of modern neural networks.
\newblock In \emph{ICML}, 2017.

\bibitem[Guo et~al.(2018)Guo, Rana, Cisse, and van~der
  Maaten]{guo2017countering}
C.~Guo, M.~Rana, M.~Cisse, and L.~van~der Maaten.
\newblock Countering adversarial images using input transformations.
\newblock In \emph{ICLR}, 2018.

\bibitem[Guo et~al.(2020)Guo, Yang, Xu, Liu, and Lin]{guo2020meets}
M.~Guo, Y.~Yang, R.~Xu, Z.~Liu, and D.~Lin.
\newblock When {NAS} meets robustness: In search of robust architectures
  against adversarial attacks.
\newblock In \emph{CVPR}, 2020.

\bibitem[Haixiang et~al.(2017)Haixiang, Yijing, Shang, Mingyun, Yuanyue, and
  Bing]{haixiang2017learning}
G.~Haixiang, L.~Yijing, J.~Shang, G.~Mingyun, H.~Yuanyue, and G.~Bing.
\newblock Learning from class-imbalanced data: Review of methods and
  applications.
\newblock \emph{Expert systems with applications}, 73:\penalty0 220--239, 2017.

\bibitem[Han et~al.(2018)Han, Yao, Yu, Niu, Xu, Hu, Tsang, and
  Sugiyama]{han2018co}
B.~Han, Q.~Yao, X.~Yu, G.~Niu, M.~Xu, W.~Hu, I.~Tsang, and M.~Sugiyama.
\newblock Co-teaching: Robust training of deep neural networks with extremely
  noisy labels.
\newblock In \emph{NeurIPS}, 2018.

\bibitem[Han et~al.(2019)Han, Luo, and Wang]{han2019deep}
J.~Han, P.~Luo, and X.~Wang.
\newblock Deep self-learning from noisy labels.
\newblock In \emph{ICCV}, pages 5138--5147, 2019.

\bibitem[Haussmann et~al.(2020)Haussmann, Fenzi, Chitta, Ivanecky, Xu, Roy,
  Mittel, Koumchatzky, Farabet, and Alvarez]{haussmann2020scalable}
E.~Haussmann, M.~Fenzi, K.~Chitta, J.~Ivanecky, H.~Xu, D.~Roy, A.~Mittel,
  N.~Koumchatzky, C.~Farabet, and J.~M. Alvarez.
\newblock Scalable active learning for object detection.
\newblock In \emph{IEEE Intelligent Vehicles Symposium (IV)}, 2020.

\bibitem[Hawkins et~al.(2021)Hawkins, Paterson, Picardi, Jia, Calinescu, and
  Habli]{hawkins2021guidance}
R.~Hawkins, C.~Paterson, C.~Picardi, Y.~Jia, R.~Calinescu, and I.~Habli.
\newblock Guidance on the assurance of machine learning in autonomous systems
  (amlas).
\newblock \emph{arXiv preprint arXiv:2102.01564}, 2021.

\bibitem[Hein et~al.(2019)Hein, Andriushchenko, and Bitterwolf]{hein2019relu}
M.~Hein, M.~Andriushchenko, and J.~Bitterwolf.
\newblock Why relu networks yield high-confidence predictions far away from the
  training data and how to mitigate the problem.
\newblock In \emph{CVPR}, 2019.

\bibitem[Hendrycks and Dietterich(2019)]{hendrycks2019benchmarking}
D.~Hendrycks and T.~Dietterich.
\newblock Benchmarking neural network robustness to common corruptions and
  perturbations.
\newblock \emph{ICLR}, 2019.

\bibitem[Hendrycks and Gimpel(2016{\natexlab{a}})]{hendrycks2016baseline}
D.~Hendrycks and K.~Gimpel.
\newblock A baseline for detecting misclassified and out-of-distribution
  examples in neural networks.
\newblock \emph{arXiv preprint arXiv:1610.02136}, 2016{\natexlab{a}}.

\bibitem[Hendrycks and Gimpel(2016{\natexlab{b}})]{hendrycks2016early}
D.~Hendrycks and K.~Gimpel.
\newblock Early methods for detecting adversarial images.
\newblock \emph{arXiv preprint arXiv:1608.00530}, 2016{\natexlab{b}}.

\bibitem[Hendrycks et~al.(2018)Hendrycks, Mazeika, and
  Dietterich]{hendrycks2018deep}
D.~Hendrycks, M.~Mazeika, and T.~Dietterich.
\newblock Deep anomaly detection with outlier exposure.
\newblock In \emph{International Conference on Learning Representations}, 2018.

\bibitem[Hendrycks et~al.(2019{\natexlab{a}})Hendrycks, Basart, Mazeika,
  Mostajabi, Steinhardt, and Song]{hendrycks2019scaling}
D.~Hendrycks, S.~Basart, M.~Mazeika, M.~Mostajabi, J.~Steinhardt, and D.~Song.
\newblock Scaling out-of-distribution detection for real-world settings.
\newblock \emph{arXiv preprint arXiv:1911.11132}, 2019{\natexlab{a}}.

\bibitem[Hendrycks et~al.(2019{\natexlab{b}})Hendrycks, Lee, and
  Mazeika]{hendrycks2019using}
D.~Hendrycks, K.~Lee, and M.~Mazeika.
\newblock Using pre-training can improve model robustness and uncertainty.
\newblock \emph{arXiv preprint arXiv:1901.09960}, 2019{\natexlab{b}}.

\bibitem[Hendrycks et~al.(2019{\natexlab{c}})Hendrycks, Mazeika, Kadavath, and
  Song]{Hendrycks2019sv}
D.~Hendrycks, M.~Mazeika, S.~Kadavath, and D.~Song.
\newblock Using self-supervised learning can improve model robustness and
  uncertainty.
\newblock In \emph{NeurIPS}, pages 15663--15674, 2019{\natexlab{c}}.

\bibitem[Hendrycks et~al.(2019{\natexlab{d}})Hendrycks, Mu, Cubuk, Zoph,
  Gilmer, and Lakshminarayanan]{hendrycks2019augmix}
D.~Hendrycks, N.~Mu, E.~D. Cubuk, B.~Zoph, J.~Gilmer, and B.~Lakshminarayanan.
\newblock Augmix: A simple data processing method to improve robustness and
  uncertainty.
\newblock \emph{arXiv preprint arXiv:1912.02781}, 2019{\natexlab{d}}.

\bibitem[Hendrycks et~al.(2019{\natexlab{e}})Hendrycks, Zhao, Basart,
  Steinhardt, and Song]{hendrycks2019natural}
D.~Hendrycks, K.~Zhao, S.~Basart, J.~Steinhardt, and D.~Song.
\newblock Natural adversarial examples.
\newblock \emph{arXiv preprint arXiv:1907.07174}, 2019{\natexlab{e}}.

\bibitem[Hendrycks et~al.(2020)Hendrycks, Basart, Mu, Kadavath, Wang, Dorundo,
  Desai, Zhu, Parajuli, Guo, et~al.]{hendrycks2020many}
D.~Hendrycks, S.~Basart, N.~Mu, S.~Kadavath, F.~Wang, E.~Dorundo, R.~Desai,
  T.~Zhu, S.~Parajuli, M.~Guo, et~al.
\newblock The many faces of robustness: A critical analysis of
  out-of-distribution generalization.
\newblock \emph{arXiv:2006.16241}, 2020.

\bibitem[Hendrycks et~al.(2021)Hendrycks, Carlini, Schulman, and
  Steinhardt]{hendrycks2021unsolved}
D.~Hendrycks, N.~Carlini, J.~Schulman, and J.~Steinhardt.
\newblock Unsolved problems in ml safety.
\newblock \emph{arXiv preprint arXiv:2109.13916}, 2021.

\bibitem[Hern{\'a}ndez-Orallo et~al.(2019)Hern{\'a}ndez-Orallo,
  Mart{\'\i}nez-Plumed, Avin, and h{\'E}igeartaigh]{hernandez2019surveying}
J.~Hern{\'a}ndez-Orallo, F.~Mart{\'\i}nez-Plumed, S.~Avin, and S.~{\'O}.
  h{\'E}igeartaigh.
\newblock Surveying safety-relevant ai characteristics.
\newblock In \emph{SafeAI@ AAAI}, 2019.

\bibitem[Hohman et~al.(2019)Hohman, Park, Robinson, and Chau]{hohman2019summit}
F.~Hohman, H.~Park, C.~Robinson, and D.~H.~P. Chau.
\newblock Summit: scaling deep learning interpretability by visualizing
  activation and attribution summarizations.
\newblock \emph{IEEE Transactions on Visualization and Computer Graphics},
  2019.

\bibitem[Hong et~al.(2021)Hong, Wang, Wang, and Zhou]{hong2021federated}
J.~Hong, H.~Wang, Z.~Wang, and J.~Zhou.
\newblock Federated robustness propagation: Sharing adversarial robustness in
  federated learning.
\newblock \emph{arXiv preprint arXiv:2106.10196}, 2021.

\bibitem[Hong et~al.(2022)Hong, Wang, Wang, and Zhou]{hong2022efficient}
J.~Hong, H.~Wang, Z.~Wang, and J.~Zhou.
\newblock Efficient split-mix federated learning for on-demand and in-situ
  customization.
\newblock In \emph{International Conference on Learning Representations}, 2022.

\bibitem[Hsu et~al.(2020)Hsu, Shen, Jin, and Kira]{hsu2020generalized}
Y.-C. Hsu, Y.~Shen, H.~Jin, and Z.~Kira.
\newblock Generalized odin: Detecting out-of-distribution image without
  learning from out-of-distribution data.
\newblock In \emph{Proceedings of the IEEE/CVF Conference on Computer Vision
  and Pattern Recognition}, 2020.

\bibitem[Hu et~al.(2020)Hu, Chen, Wang, and Wang]{hu2020triple}
T.-K. Hu, T.~Chen, H.~Wang, and Z.~Wang.
\newblock Triple wins: Boosting accuracy, robustness and efficiency together by
  enabling input-adaptive inference.
\newblock \emph{arXiv preprint arXiv:2002.10025}, 2020.

\bibitem[Huang et~al.(2017)Huang, Kwiatkowska, Wang, and Wu]{huang2017safety}
X.~Huang, M.~Kwiatkowska, S.~Wang, and M.~Wu.
\newblock Safety verification of deep neural networks.
\newblock In \emph{International conference on computer aided verification},
  pages 3--29. Springer, 2017.

\bibitem[Huang et~al.(2020{\natexlab{a}})Huang, Kroening, Ruan, Sharp, Sun,
  Thamo, Wu, and Yi]{huang2020survey}
X.~Huang, D.~Kroening, W.~Ruan, J.~Sharp, Y.~Sun, E.~Thamo, M.~Wu, and X.~Yi.
\newblock A survey of safety and trustworthiness of deep neural networks:
  Verification, testing, adversarial attack and defence, and interpretability.
\newblock \emph{Computer Science Review}, 37:\penalty0 100270,
  2020{\natexlab{a}}.

\bibitem[Huang et~al.(2020{\natexlab{b}})Huang, Wang, Xing, and
  Huang]{huang2020self}
Z.~Huang, H.~Wang, E.~P. Xing, and D.~Huang.
\newblock Self-challenging improves cross-domain generalization.
\newblock In \emph{ECCV}, 2020{\natexlab{b}}.

\bibitem[Iyyer et~al.(2018)Iyyer, Wieting, Gimpel, and
  Zettlemoyer]{iyyer2018adversarial}
M.~Iyyer, J.~Wieting, K.~Gimpel, and L.~Zettlemoyer.
\newblock Adversarial example generation with syntactically controlled
  paraphrase networks.
\newblock In \emph{NAACL-HLT}, pages 1875--1885, 2018.

\bibitem[Jiang et~al.(2020)Jiang, Chen, Chen, and Wang]{jiang2020robust}
Z.~Jiang, T.~Chen, T.~Chen, and Z.~Wang.
\newblock Robust pre-training by adversarial contrastive learning.
\newblock \emph{arXiv preprint arXiv:2010.13337}, 2020.

\bibitem[Kahng et~al.(2018)Kahng, Andrews, Kalro, and Chau]{kahng2018cti}
M.~Kahng, P.~Y. Andrews, A.~Kalro, and D.~H.~P. Chau.
\newblock Activis: visual exploration of industry-scale deep neural network
  models.
\newblock \emph{IEEE Transactions on Visualization and Computer Graphics},
  24\penalty0 (1):\penalty0 88--97, 2018.

\bibitem[Kang et~al.(2019)Kang, Sun, Hendrycks, Brown, and
  Steinhardt]{kang2019testing}
D.~Kang, Y.~Sun, D.~Hendrycks, T.~Brown, and J.~Steinhardt.
\newblock Testing robustness against unforeseen adversaries.
\newblock \emph{arXiv preprint arXiv:1908.08016}, 2019.

\bibitem[Karandikar et~al.(2021)Karandikar, Cain, Tran, Lakshminarayanan,
  Shlens, Mozer, and Roelofs]{karandikar2021soft}
A.~Karandikar, N.~Cain, D.~Tran, B.~Lakshminarayanan, J.~Shlens, M.~C. Mozer,
  and R.~Roelofs.
\newblock Soft calibration objectives for neural networks.
\newblock In \emph{NeurIPS}, 2021.

\bibitem[Katharopoulos and Fleuret(2018)]{katharopoulos2018not}
A.~Katharopoulos and F.~Fleuret.
\newblock Not all samples are created equal: Deep learning with importance
  sampling.
\newblock In \emph{ICML}, pages 2525--2534, 2018.

\bibitem[Katz et~al.(2017)Katz, Barrett, Dill, Julian, and
  Kochenderfer]{katz2017reluplex}
G.~Katz, C.~Barrett, D.~L. Dill, K.~Julian, and M.~J. Kochenderfer.
\newblock Reluplex: An efficient {SMT} solver for verifying deep neural
  networks.
\newblock In \emph{International Conference on Computer Aided Verification},
  pages 97--117. Springer, 2017.

\bibitem[Kendall and Gal(2017)]{kendall2017uncertainties}
A.~Kendall and Y.~Gal.
\newblock What uncertainties do we need in bayesian deep learning for computer
  vision?
\newblock In \emph{NIPS}, pages 5574--5584, 2017.

\bibitem[Kim et~al.(2018)Kim, Wattenberg, Gilmer, Cai, Wexler, Viegas,
  et~al.]{kim2018interpretability}
B.~Kim, M.~Wattenberg, J.~Gilmer, C.~Cai, J.~Wexler, F.~Viegas, et~al.
\newblock Interpretability beyond feature attribution: Quantitative testing
  with concept activation vectors (tcav).
\newblock In \emph{International conference on machine learning}, 2018.

\bibitem[Kim et~al.(2020)Kim, Gopinath, Pasareanu, and
  Seshia]{kim2020programmatic}
E.~Kim, D.~Gopinath, C.~Pasareanu, and S.~A. Seshia.
\newblock A programmatic and semantic approach to explaining and debugging
  neural network based object detectors.
\newblock In \emph{CVPR}, 2020.

\bibitem[Kindermans et~al.(2019)Kindermans, Hooker, Adebayo, Alber, Sch{\"u}tt,
  D{\"a}hne, Erhan, and Kim]{kindermans2017reliability}
P.-J. Kindermans, S.~Hooker, J.~Adebayo, M.~Alber, K.~T. Sch{\"u}tt,
  S.~D{\"a}hne, D.~Erhan, and B.~Kim.
\newblock The (un) reliability of saliency methods.
\newblock In \emph{Explainable AI: Interpreting, Explaining and Visualizing
  Deep Learning}. Springer, 2019.

\bibitem[Koopman and Wagner(2017)]{koopman2017autonomous}
P.~Koopman and M.~Wagner.
\newblock Autonomous vehicle safety: An interdisciplinary challenge.
\newblock \emph{IEEE Intelligent Transportation Systems Magazine}, 9\penalty0
  (1):\penalty0 90--96, 2017.

\bibitem[Krishnan and Tickoo(2020)]{krishnan2020improving}
R.~Krishnan and O.~Tickoo.
\newblock Improving model calibration with accuracy versus uncertainty
  optimization.
\newblock In \emph{NeurIPS}, pages 18237--18248, 2020.

\bibitem[Kumar et~al.(2018)Kumar, Sarawagi, and Jain]{kumar2018trainable}
A.~Kumar, S.~Sarawagi, and U.~Jain.
\newblock Trainable calibration measures for neural networks from kernel mean
  embeddings.
\newblock In \emph{ICML}, 2018.

\bibitem[Lage et~al.(2018)Lage, Ross, Gershman, Kim, and
  Doshi-Velez]{LageRGKD18}
I.~Lage, A.~S. Ross, S.~J. Gershman, B.~Kim, and F.~Doshi-Velez.
\newblock Human-in-the-loop interpretability prior.
\newblock In \emph{NeurIPS}, pages 10180--10189, 2018.

\bibitem[Lakkaraju et~al.(2016)Lakkaraju, Bach, and
  Leskovec]{lakkaraju2016interpretable}
H.~Lakkaraju, S.~H. Bach, and J.~Leskovec.
\newblock Interpretable decision sets: A joint framework for description and
  prediction.
\newblock In \emph{Proceedings of the 22nd ACM SIGKDD International Conference
  on Knowledge Discovery and Data Mining}, 2016.

\bibitem[Lakkaraju et~al.(2017)Lakkaraju, Kamar, Caruana, and
  Horvitz]{lakkaraju2017identifying}
H.~Lakkaraju, E.~Kamar, R.~Caruana, and E.~Horvitz.
\newblock Identifying unknown unknowns in the open world: Representations and
  policies for guided exploration.
\newblock In \emph{AAAI}, 2017.

\bibitem[Lakkaraju et~al.(2019)Lakkaraju, Kamar, Caruana, and
  Leskovec]{lakkaraju2019faithful}
H.~Lakkaraju, E.~Kamar, R.~Caruana, and J.~Leskovec.
\newblock Faithful and customizable explanations of black box models.
\newblock In \emph{Proceedings of the 2019 AAAI/ACM Conference on AI, Ethics,
  and Society}, pages 131--138, 2019.

\bibitem[Lakkaraju et~al.(2020)Lakkaraju, Arsov, and
  Bastani]{lakkaraju2020robust}
H.~Lakkaraju, N.~Arsov, and O.~Bastani.
\newblock Robust and stable black box explanations.
\newblock In \emph{International Conference on Machine Learning}, pages
  5628--5638. PMLR, 2020.

\bibitem[Lakshminarayanan et~al.(2017)Lakshminarayanan, Pritzel, and
  Blundell]{lakshminarayanan2017simple}
B.~Lakshminarayanan, A.~Pritzel, and C.~Blundell.
\newblock Simple and scalable predictive uncertainty estimation using deep
  ensembles.
\newblock In \emph{NIPS}, 2017.

\bibitem[Lambert et~al.(2020)Lambert, Liu, Sener, Hays, and
  Koltun]{lambert2020mseg}
J.~Lambert, Z.~Liu, O.~Sener, J.~Hays, and V.~Koltun.
\newblock {MSeg}: A composite dataset for multi-domain semantic segmentation.
\newblock In \emph{CVPR}, 2020.

\bibitem[LeCun et~al.(2015)LeCun, Bengio, and Hinton]{lecun2015deep}
Y.~LeCun, Y.~Bengio, and G.~Hinton.
\newblock Deep learning.
\newblock \emph{Nature}, 521\penalty0 (7553):\penalty0 436--444, 2015.

\bibitem[Lee et~al.(2017)Lee, Lee, Lee, and Shin]{lee2017training}
K.~Lee, H.~Lee, K.~Lee, and J.~Shin.
\newblock Training confidence-calibrated classifiers for detecting
  out-of-distribution samples.
\newblock \emph{arXiv preprint arXiv:1711.09325}, 2017.

\bibitem[Lee et~al.(2018)Lee, Lee, Lee, and Shin]{lee2018simple}
K.~Lee, K.~Lee, H.~Lee, and J.~Shin.
\newblock A simple unified framework for detecting out-of-distribution samples
  and adversarial attacks.
\newblock In \emph{Advances in Neural Information Processing Systems}, pages
  7167--7177, 2018.

\bibitem[Lee et~al.(2020)Lee, Lee, Shin, and Lee]{lee2020network}
K.~Lee, K.~Lee, J.~Shin, and H.~Lee.
\newblock Network randomization: A simple technique for generalization in deep
  reinforcement learning.
\newblock In \emph{ICLR}, 2020.

\bibitem[Leike et~al.(2017)Leike, Martic, Krakovna, Ortega, Everitt, Lefrancq,
  Orseau, and Legg]{leike2017ai}
J.~Leike, M.~Martic, V.~Krakovna, P.~A. Ortega, T.~Everitt, A.~Lefrancq,
  L.~Orseau, and S.~Legg.
\newblock Ai safety gridworlds.
\newblock \emph{arXiv preprint arXiv:1711.09883}, 2017.

\bibitem[Lertvittayakumjorn and Toni(2019)]{Lertvittayakumjorn2019human}
P.~Lertvittayakumjorn and F.~Toni.
\newblock Human-grounded evaluations of explanation methods for text
  classification.
\newblock In \emph{EMNLP-IJCNLP}, pages 5198--5208, 2019.

\bibitem[Li et~al.(2017)Li, Yang, Song, and Hospedales]{li2017deeper}
D.~Li, Y.~Yang, Y.-Z. Song, and T.~M. Hospedales.
\newblock Deeper, broader and artier domain generalization.
\newblock In \emph{Proceedings of the IEEE international conference on computer
  vision}, pages 5542--5550, 2017.

\bibitem[Li et~al.(2018)Li, Yang, Song, and Hospedales]{li2018learning}
D.~Li, Y.~Yang, Y.-Z. Song, and T.~Hospedales.
\newblock Learning to generalize: Meta-learning for domain generalization.
\newblock In \emph{Proceedings of the AAAI Conference on Artificial
  Intelligence}, volume~32, 2018.

\bibitem[Li and Li(2017)]{li2017adversarial}
X.~Li and F.~Li.
\newblock Adversarial examples detection in deep networks with convolutional
  filter statistics.
\newblock In \emph{Proceedings of the IEEE International Conference on Computer
  Vision}, pages 5764--5772, 2017.

\bibitem[Li et~al.(2020)Li, Yu, Tan, Mei, Tang, Shen, Yuille,
  et~al.]{li2020shape}
Y.~Li, Q.~Yu, M.~Tan, J.~Mei, P.~Tang, W.~Shen, A.~Yuille, et~al.
\newblock Shape-texture debiased neural network training.
\newblock In \emph{ICLR}, 2020.

\bibitem[Li and Hoiem(2020)]{li2020improving}
Z.~Li and D.~Hoiem.
\newblock Improving confidence estimates for unfamiliar examples.
\newblock In \emph{CVPR}, 2020.

\bibitem[Lin et~al.(2019)Lin, Gan, and Han]{lin2018defensive}
J.~Lin, C.~Gan, and S.~Han.
\newblock Defensive quantization: When efficiency meets robustness.
\newblock In \emph{ICLR}, 2019.

\bibitem[Lipton et~al.(2018)Lipton, Wang, and Smola]{lipton2018detecting}
Z.~Lipton, Y.-X. Wang, and A.~Smola.
\newblock Detecting and correcting for label shift with black box predictors.
\newblock In \emph{International conference on machine learning}, pages
  3122--3130, 2018.

\bibitem[Liu et~al.(2020{\natexlab{a}})Liu, Lin, Padhy, Tran, Bedrax-Weiss, and
  Lakshminarayanan]{liu2020simple}
J.~Z. Liu, Z.~Lin, S.~Padhy, D.~Tran, T.~Bedrax-Weiss, and B.~Lakshminarayanan.
\newblock Simple and principled uncertainty estimation with deterministic deep
  learning via distance awareness.
\newblock \emph{arXiv preprint arXiv:2006.10108}, 2020{\natexlab{a}}.

\bibitem[Liu et~al.(2020{\natexlab{b}})Liu, Wang, Owens, and Li]{liu2020energy}
W.~Liu, X.~Wang, J.~D. Owens, and Y.~Li.
\newblock Energy-based out-of-distribution detection.
\newblock \emph{arXiv:2010.03759}, 2020{\natexlab{b}}.

\bibitem[Lu et~al.(2021)Lu, Zhu, Xu, Whittle, Douglas, and
  Sanderson]{lu2021software}
Q.~Lu, L.~Zhu, X.~Xu, J.~Whittle, D.~Douglas, and C.~Sanderson.
\newblock Software engineering for responsible ai: An empirical study and
  operationalised patterns.
\newblock \emph{arXiv preprint arXiv:2111.09478}, 2021.

\bibitem[Lundberg and Lee(2017)]{lundberg2017unified}
S.~M. Lundberg and S.-I. Lee.
\newblock A unified approach to interpreting model predictions.
\newblock In \emph{Advances in Neural Information Processing Systems}, pages
  4765--4774, 2017.

\bibitem[L’heureux et~al.(2017)L’heureux, Grolinger, Elyamany, and
  Capretz]{l2017machine}
A.~L’heureux, K.~Grolinger, H.~F. Elyamany, and M.~A. Capretz.
\newblock Machine learning with big data: Challenges and approaches.
\newblock \emph{IEEE Access}, 5:\penalty0 7776--7797, 2017.

\bibitem[Ma et~al.(2018)Ma, Li, Wang, Erfani, Wijewickrema, Schoenebeck, Song,
  Houle, and Bailey]{ma2018characterizing}
X.~Ma, B.~Li, Y.~Wang, S.~M. Erfani, S.~Wijewickrema, G.~Schoenebeck, D.~Song,
  M.~E. Houle, and J.~Bailey.
\newblock Characterizing adversarial subspaces using local intrinsic
  dimensionality.
\newblock \emph{arXiv preprint arXiv:1801.02613}, 2018.

\bibitem[Maaten and Hinton(2008)]{maaten2008visualizing}
L.~v.~d. Maaten and G.~Hinton.
\newblock Visualizing data using t-sne.
\newblock \emph{Journal of Machine Learning Research}, 2008.

\bibitem[Madry et~al.(2017)Madry, Makelov, Schmidt, Tsipras, and
  Vladu]{madry2017towards}
A.~Madry, A.~Makelov, L.~Schmidt, D.~Tsipras, and A.~Vladu.
\newblock Towards deep learning models resistant to adversarial attacks.
\newblock \emph{arXiv preprint arXiv:1706.06083}, 2017.

\bibitem[Mahajan et~al.(2018)Mahajan, Girshick, Ramanathan, He, Paluri, Li,
  Bharambe, and Van Der~Maaten]{mahajan2018exploring}
D.~Mahajan, R.~Girshick, V.~Ramanathan, K.~He, M.~Paluri, Y.~Li, A.~Bharambe,
  and L.~Van Der~Maaten.
\newblock Exploring the limits of weakly supervised pretraining.
\newblock In \emph{Proceedings of the European conference on computer vision
  (ECCV)}, pages 181--196, 2018.

\bibitem[Mahmood et~al.(2021)Mahmood, Mahmood, and
  Van~Dijk]{mahmood2021robustness}
K.~Mahmood, R.~Mahmood, and M.~Van~Dijk.
\newblock On the robustness of vision transformers to adversarial examples.
\newblock \emph{arXiv preprint arXiv:2104.02610}, 2021.

\bibitem[Malinin et~al.(2021)Malinin, Band, Gal, Gales, Ganshin, Chesnokov,
  Noskov, Ploskonosov, Prokhorenkova, Provilkov, et~al.]{malinin2021shifts}
A.~Malinin, N.~Band, Y.~Gal, M.~Gales, A.~Ganshin, G.~Chesnokov, A.~Noskov,
  A.~Ploskonosov, L.~Prokhorenkova, I.~Provilkov, et~al.
\newblock Shifts: A dataset of real distributional shift across multiple
  large-scale tasks.
\newblock In \emph{Advances in Neural Information Processing Systems}, 2021.

\bibitem[McAllister et~al.(2017)McAllister, Gal, Kendall, Van Der~Wilk, Shah,
  Cipolla, and Weller]{mcallister2017concrete}
R.~McAllister, Y.~Gal, A.~Kendall, M.~Van Der~Wilk, A.~Shah, R.~Cipolla, and
  A.~V. Weller.
\newblock Concrete problems for autonomous vehicle safety: Advantages of
  bayesian deep learning.
\newblock IJCAI, 2017.

\bibitem[Meinke and Hein(2019)]{meinke2019towards}
A.~Meinke and M.~Hein.
\newblock Towards neural networks that provably know when they don't know.
\newblock In \emph{International Conference on Learning Representations}, 2019.

\bibitem[Meng and Chen(2017)]{meng2017magnet}
D.~Meng and H.~Chen.
\newblock Magnet: a two-pronged defense against adversarial examples.
\newblock In \emph{Proceedings of the 2017 ACM SIGSAC Conference on Computer
  and Communications Security}, pages 135--147. ACM, 2017.

\bibitem[Meng et~al.(2021)Meng, Wang, Zhou, Shen, Jia, and
  Van~Gool]{meng2021towards}
Q.~Meng, W.~Wang, T.~Zhou, J.~Shen, Y.~Jia, and L.~Van~Gool.
\newblock Towards a weakly supervised framework for 3d point cloud object
  detection and annotation.
\newblock \emph{IEEE Transactions on Pattern Analysis and Machine
  Intelligence}, 2021.

\bibitem[{Miyato} et~al.(2019){Miyato}, {Maeda}, {Koyama}, and
  {Ishii}]{miyato2019virtual}
T.~{Miyato}, S.-I. {Maeda}, M.~{Koyama}, and S.~{Ishii}.
\newblock Virtual adversarial training: A regularization method for supervised
  and semi-supervised learning.
\newblock \emph{IEEE Transactions on Pattern Analysis and Machine
  Intelligence}, 41\penalty0 (8):\penalty0 1979--1993, 2019.

\bibitem[Mohri et~al.(2018)Mohri, Rostamizadeh, and
  Talwalkar]{mohri2018foundations}
M.~Mohri, A.~Rostamizadeh, and A.~Talwalkar.
\newblock \emph{Foundations of machine learning}.
\newblock MIT press, 2018.

\bibitem[Mohseni et~al.(2018)Mohseni, Zarei, and
  Ragan]{mohseni2018multidisciplinary}
S.~Mohseni, N.~Zarei, and E.~D. Ragan.
\newblock A multidisciplinary survey and framework for design and evaluation of
  explainable ai systems.
\newblock \emph{arXiv}, pages arXiv--1811, 2018.

\bibitem[Mohseni et~al.(2019)Mohseni, Pitale, Singh, and
  Wang]{mohseni2019practical}
S.~Mohseni, M.~Pitale, V.~Singh, and Z.~Wang.
\newblock Practical solutions for machine learning safety in autonomous
  vehicles.
\newblock \emph{arXiv preprint arXiv:1912.09630}, 2019.

\bibitem[Mohseni et~al.(2020)Mohseni, Pitale, Yadawa, and Wang]{mohseni2020ood}
S.~Mohseni, M.~Pitale, J.~Yadawa, and Z.~Wang.
\newblock Self-supervised learning for generalizable out-of-distribution
  detection.
\newblock In \emph{AAAI Conference on Artificial Intelligence}, 2020.

\bibitem[Mohseni et~al.(2021{\natexlab{a}})Mohseni, Block, and
  Ragan]{mohseni2018human}
S.~Mohseni, J.~E. Block, and E.~Ragan.
\newblock Quantitative evaluation of machine learning explanations: A
  human-grounded benchmark.
\newblock In \emph{26th International Conference on Intelligent User
  Interfaces}, 2021{\natexlab{a}}.
\newblock \doi{10.1145/3397481.3450689}.

\bibitem[Mohseni et~al.(2021{\natexlab{b}})Mohseni, Vahdat, and
  Yadawa]{mohseni2021multitask}
S.~Mohseni, A.~Vahdat, and J.~Yadawa.
\newblock Shifting transformation learning for out-of-distribution detection.
\newblock \emph{arXiv preprint arXiv:2106.03899}, 2021{\natexlab{b}}.

\bibitem[Mok et~al.(2021)Mok, Na, Choe, and Yoon]{mok2021advrush}
J.~Mok, B.~Na, H.~Choe, and S.~Yoon.
\newblock {AdvRush}: Searching for adversarially robust neural architectures.
\newblock In \emph{ICCV}, pages 12322--12332, 2021.

\bibitem[Moon et~al.(2020)Moon, Kim, Shin, and Hwang]{moon2020confidence}
J.~Moon, J.~Kim, Y.~Shin, and S.~Hwang.
\newblock Confidence-aware learning for deep neural networks.
\newblock In \emph{ICML}, pages 7034--7044, 2020.

\bibitem[Mukhoti et~al.(2021)Mukhoti, Kirsch, van Amersfoort, Torr, and
  Gal]{mukhoti2021deterministic}
J.~Mukhoti, A.~Kirsch, J.~van Amersfoort, P.~H. Torr, and Y.~Gal.
\newblock Deterministic neural networks with appropriate inductive biases
  capture epistemic and aleatoric uncertainty.
\newblock \emph{arXiv preprint arXiv:2102.11582}, 2021.

\bibitem[M{\"u}ller et~al.(2019)M{\"u}ller, Kornblith, and
  Hinton]{muller2019does}
R.~M{\"u}ller, S.~Kornblith, and G.~Hinton.
\newblock When does label smoothing help?
\newblock \emph{arXiv preprint arXiv:1906.02629}, 2019.

\bibitem[Nakkiran(2019)]{nakkiran2019adversarial}
P.~Nakkiran.
\newblock Adversarial robustness may be at odds with simplicity.
\newblock \emph{arXiv preprint arXiv:1901.00532}, 2019.

\bibitem[Narodytska et~al.(2018)Narodytska, Kasiviswanathan, Ryzhyk, Sagiv, and
  Walsh]{narodytska2018verifying}
N.~Narodytska, S.~Kasiviswanathan, L.~Ryzhyk, M.~Sagiv, and T.~Walsh.
\newblock Verifying properties of binarized deep neural networks.
\newblock In \emph{Proceedings of the AAAI Conference on Artificial
  Intelligence}, volume~32, 2018.

\bibitem[Ning et~al.(2020)Ning, Zhao, Li, Zhao, Yang, and Wang]{ning2020multi}
X.~Ning, J.~Zhao, W.~Li, T.~Zhao, H.~Yang, and Y.~Wang.
\newblock Multi-shot {NAS} for discovering adversarially robust convolutional
  neural architectures at targeted capacities.
\newblock \emph{arXiv preprint arXiv:2012.11835}, 2020.

\bibitem[Olah et~al.(2018)Olah, Satyanarayan, Johnson, Carter, Schubert, Ye,
  and Mordvintsev]{olah2018the}
C.~Olah, A.~Satyanarayan, I.~Johnson, S.~Carter, L.~Schubert, K.~Ye, and
  A.~Mordvintsev.
\newblock The building blocks of interpretability.
\newblock \emph{Distill}, 2018.
\newblock \doi{10.23915/distill.00010}.

\bibitem[Ovadia et~al.(2019)Ovadia, Fertig, Ren, Nado, Sculley, Nowozin,
  Dillon, Lakshminarayanan, and Snoek]{ovadia2019can}
Y.~Ovadia, E.~Fertig, J.~Ren, Z.~Nado, D.~Sculley, S.~Nowozin, J.~V. Dillon,
  B.~Lakshminarayanan, and J.~Snoek.
\newblock Can you trust your model's uncertainty? evaluating predictive
  uncertainty under dataset shift.
\newblock \emph{arXiv:1906.02530}, 2019.

\bibitem[Pan et~al.(2018)Pan, Luo, Shi, and Tang]{pan2018two}
X.~Pan, P.~Luo, J.~Shi, and X.~Tang.
\newblock Two at once: Enhancing learning and generalization capacities via
  ibn-net.
\newblock In \emph{Proceedings of the European Conference on Computer Vision
  (ECCV)}, pages 464--479, 2018.

\bibitem[Pei et~al.(2017{\natexlab{a}})Pei, Cao, Yang, and
  Jana]{pei2017deepxplore}
K.~Pei, Y.~Cao, J.~Yang, and S.~Jana.
\newblock Deepxplore: Automated whitebox testing of deep learning systems.
\newblock In \emph{proceedings of the 26th Symposium on Operating Systems
  Principles}, pages 1--18, 2017{\natexlab{a}}.

\bibitem[Pei et~al.(2017{\natexlab{b}})Pei, Cao, Yang, and
  Jana]{pei2017towards}
K.~Pei, Y.~Cao, J.~Yang, and S.~Jana.
\newblock Towards practical verification of machine learning: The case of
  computer vision systems.
\newblock \emph{arXiv preprint arXiv:1712.01785}, 2017{\natexlab{b}}.

\bibitem[{Qian} and {Wegman}(2018)]{qian2018l2}
H.~{Qian} and M.~N. {Wegman}.
\newblock L2-nonexpansive neural networks, 2018.

\bibitem[Qin et~al.(2018)Qin, Wang, Xu, Ma, and Lu]{qin2018syneva}
Y.~Qin, H.~Wang, C.~Xu, X.~Ma, and J.~Lu.
\newblock Syneva: Evaluating ml programs by mirror program synthesis.
\newblock In \emph{2018 IEEE International Conference on Software Quality,
  Reliability and Security (QRS)}, pages 171--182. IEEE, 2018.

\bibitem[Qiu et~al.(2020)Qiu, Xiao, Yang, Yan, Lee, and Li]{qiu2020semanticadv}
H.~Qiu, C.~Xiao, L.~Yang, X.~Yan, H.~Lee, and B.~Li.
\newblock Semanticadv: Generating adversarial examples via
  attribute-conditioned image editing.
\newblock In \emph{European Conference on Computer Vision}, pages 19--37.
  Springer, 2020.

\bibitem[Qui{\~n}onero-Candela et~al.(2009)Qui{\~n}onero-Candela, Sugiyama,
  Lawrence, and Schwaighofer]{quinonero2009dataset}
J.~Qui{\~n}onero-Candela, M.~Sugiyama, N.~D. Lawrence, and A.~Schwaighofer.
\newblock \emph{Dataset shift in machine learning}.
\newblock Mit Press, 2009.

\bibitem[Radford et~al.(2021)Radford, Kim, Hallacy, Ramesh, Goh, Agarwal,
  Sastry, Askell, Mishkin, Clark, et~al.]{radford2021learning}
A.~Radford, J.~W. Kim, C.~Hallacy, A.~Ramesh, G.~Goh, S.~Agarwal, G.~Sastry,
  A.~Askell, P.~Mishkin, J.~Clark, et~al.
\newblock Learning transferable visual models from natural language
  supervision.
\newblock In \emph{ICML}, pages 8748--8763, 2021.

\bibitem[Raghunathan et~al.(2018)Raghunathan, Steinhardt, and
  Liang]{raghunathan2018certified}
A.~Raghunathan, J.~Steinhardt, and P.~Liang.
\newblock Certified defenses against adversarial examples.
\newblock \emph{International Conference on Learning Representations (ICLR),
  arXiv preprint arXiv:1801.09344}, 2018.

\bibitem[Recht et~al.(2019)Recht, Roelofs, Schmidt, and
  Shankar]{recht2019imagenet}
B.~Recht, R.~Roelofs, L.~Schmidt, and V.~Shankar.
\newblock Do imagenet classifiers generalize to imagenet?
\newblock In \emph{International Conference on Machine Learning}, pages
  5389--5400, 2019.

\bibitem[Ren et~al.(2019{\natexlab{a}})Ren, Liu, Fertig, Snoek, Poplin,
  Depristo, Dillon, and Lakshminarayanan]{ren2019likelihood}
J.~Ren, P.~J. Liu, E.~Fertig, J.~Snoek, R.~Poplin, M.~Depristo, J.~Dillon, and
  B.~Lakshminarayanan.
\newblock Likelihood ratios for out-of-distribution detection.
\newblock In \emph{Advances in Neural Information Processing Systems}, pages
  14707--14718, 2019{\natexlab{a}}.

\bibitem[Ren et~al.(2019{\natexlab{b}})Ren, Deng, He, and
  Che]{ren2019generating}
S.~Ren, Y.~Deng, K.~He, and W.~Che.
\newblock Generating natural language adversarial examples through probability
  weighted word saliency.
\newblock In \emph{ACL}, pages 1085--1097, 2019{\natexlab{b}}.

\bibitem[Ribeiro et~al.(2016)Ribeiro, Singh, and Guestrin]{ribeiro2016should}
M.~T. Ribeiro, S.~Singh, and C.~Guestrin.
\newblock Why should i trust you? explaining the predictions of any classifier.
\newblock In \emph{Proceedings of the 22nd ACM SIGKDD International Conference
  on Knowledge Discovery and Data Mining}, 2016.

\bibitem[Ribeiro et~al.(2018)Ribeiro, Singh, and Guestrin]{ribeiro2018anchors}
M.~T. Ribeiro, S.~Singh, and C.~Guestrin.
\newblock Anchors: High-precision model-agnostic explanations.
\newblock In \emph{AAAI Conference on Artificial Intelligence}, 2018.

\bibitem[Ruff et~al.(2018)Ruff, Vandermeulen, Goernitz, Deecke, Siddiqui,
  Binder, M{\"u}ller, and Kloft]{ruff2018deep}
L.~Ruff, R.~Vandermeulen, N.~Goernitz, L.~Deecke, S.~A. Siddiqui, A.~Binder,
  E.~M{\"u}ller, and M.~Kloft.
\newblock Deep one-class classification.
\newblock In \emph{ICML}, pages 4393--4402, 2018.

\bibitem[Ruff et~al.(2019)Ruff, Vandermeulen, G{\"o}rnitz, Binder, M{\"u}ller,
  M{\"u}ller, and Kloft]{ruff2019deep}
L.~Ruff, R.~A. Vandermeulen, N.~G{\"o}rnitz, A.~Binder, E.~M{\"u}ller, K.-R.
  M{\"u}ller, and M.~Kloft.
\newblock Deep semi-supervised anomaly detection.
\newblock In \emph{International Conference on Learning Representations}, 2019.

\bibitem[Sadigh et~al.(2016{\natexlab{a}})Sadigh, Sastry, Seshia, and
  Dragan]{sadigh2016planning}
D.~Sadigh, S.~Sastry, S.~A. Seshia, and A.~D. Dragan.
\newblock Planning for autonomous cars that leverage effects on human actions.
\newblock In \emph{Robotics: Science and Systems}, volume~2. Ann Arbor, MI,
  USA, 2016{\natexlab{a}}.

\bibitem[Sadigh et~al.(2016{\natexlab{b}})Sadigh, Sastry, Seshia, and
  Dragan]{sadigh2016information}
D.~Sadigh, S.~S. Sastry, S.~A. Seshia, and A.~Dragan.
\newblock Information gathering actions over human internal state.
\newblock In \emph{2016 IEEE/RSJ International Conference on Intelligent Robots
  and Systems (IROS)}, pages 66--73. IEEE, 2016{\natexlab{b}}.

\bibitem[Salay et~al.(2017)Salay, Queiroz, and Czarnecki]{salay2017analysis}
R.~Salay, R.~Queiroz, and K.~Czarnecki.
\newblock An analysis of iso 26262: Using machine learning safely in automotive
  software.
\newblock \emph{arXiv preprint arXiv:1709.02435}, 2017.

\bibitem[Salehi et~al.(2021)Salehi, Mirzaei, Hendrycks, Li, Rohban, and
  Sabokrou]{salehi2021unified}
M.~Salehi, H.~Mirzaei, D.~Hendrycks, Y.~Li, M.~H. Rohban, and M.~Sabokrou.
\newblock A unified survey on anomaly, novelty, open-set, and
  out-of-distribution detection: Solutions and future challenges.
\newblock \emph{arXiv preprint arXiv:2110.14051}, 2021.

\bibitem[Samangouei et~al.(2018)Samangouei, Kabkab, and
  Chellappa]{samangouei2018defense}
P.~Samangouei, M.~Kabkab, and R.~Chellappa.
\newblock Defense-{GAN}: Protecting classifiers against adversarial attacks
  using generative models.
\newblock \emph{arXiv preprint arXiv:1805.06605}, 2018.

\bibitem[Samek et~al.(2017)Samek, Binder, Montavon, Lapuschkin, and
  M{\"u}ller]{samek2017evaluating}
W.~Samek, A.~Binder, G.~Montavon, S.~Lapuschkin, and K.-R. M{\"u}ller.
\newblock Evaluating the visualization of what a deep neural network has
  learned.
\newblock \emph{IEEE Transactions on Neural Networks and Learning Systems},
  28\penalty0 (11):\penalty0 2660--2673, 2017.

\bibitem[Sastry and Oore(2020)]{sastry2019detecting}
C.~S. Sastry and S.~Oore.
\newblock Detecting out-of-distribution examples with in-distribution examples
  and {Gram} matrices.
\newblock In \emph{ICML}, pages 8491--8501, 2020.

\bibitem[Schlegl et~al.(2017)Schlegl, Seeb{\"o}ck, Waldstein, Schmidt-Erfurth,
  and Langs]{schlegl2017unsupervised}
T.~Schlegl, P.~Seeb{\"o}ck, S.~M. Waldstein, U.~Schmidt-Erfurth, and G.~Langs.
\newblock Unsupervised anomaly detection with generative adversarial networks
  to guide marker discovery.
\newblock In \emph{IPMI}. Springer, 2017.

\bibitem[Sehwag et~al.(2020)Sehwag, Wang, Mittal, and Jana]{sehwag2020pruning}
V.~Sehwag, S.~Wang, P.~Mittal, and S.~Jana.
\newblock On pruning adversarially robust neural networks.
\newblock \emph{arXiv preprint arXiv:2002.10509}, 2020.

\bibitem[Sehwag et~al.(2021)Sehwag, Chiang, and Mittal]{sehwag2021ssd}
V.~Sehwag, M.~Chiang, and P.~Mittal.
\newblock Ssd: A unified framework for self-supervised outlier detection.
\newblock In \emph{International Conference on Learning Representations}, 2021.

\bibitem[Selvaraju et~al.(2017)Selvaraju, Cogswell, Das, Vedantam, Parikh, and
  Batra]{selvaraju2017grad}
R.~R. Selvaraju, M.~Cogswell, A.~Das, R.~Vedantam, D.~Parikh, and D.~Batra.
\newblock Grad-cam: Visual explanations from deep networks via gradient-based
  localization.
\newblock In \emph{Proceedings of the IEEE international conference on computer
  vision}, 2017.

\bibitem[Serr{\`a} et~al.(2019)Serr{\`a}, {\'A}lvarez, G{\'o}mez, Slizovskaia,
  N{\'u}{\~n}ez, and Luque]{serra2019input}
J.~Serr{\`a}, D.~{\'A}lvarez, V.~G{\'o}mez, O.~Slizovskaia, J.~F.
  N{\'u}{\~n}ez, and J.~Luque.
\newblock Input complexity and out-of-distribution detection with
  likelihood-based generative models.
\newblock \emph{arXiv preprint arXiv:1909.11480}, 2019.

\bibitem[Seshia et~al.(2016)Seshia, Sadigh, and Sastry]{seshia2016towards}
S.~A. Seshia, D.~Sadigh, and S.~S. Sastry.
\newblock Towards verified artificial intelligence.
\newblock \emph{arXiv preprint arXiv:1606.08514}, 2016.

\bibitem[Seshia et~al.(2018)Seshia, Desai, Dreossi, Fremont, Ghosh, Kim,
  Shivakumar, Vazquez-Chanlatte, and Yue]{seshia2018formal}
S.~A. Seshia, A.~Desai, T.~Dreossi, D.~J. Fremont, S.~Ghosh, E.~Kim,
  S.~Shivakumar, M.~Vazquez-Chanlatte, and X.~Yue.
\newblock Formal specification for deep neural networks.
\newblock In \emph{ATVA}. Springer, 2018.

\bibitem[Shafahi et~al.(2019)Shafahi, Najibi, Ghiasi, Xu, Dickerson, Studer,
  Davis, Taylor, and Goldstein]{shafahi2019adversarial}
A.~Shafahi, M.~Najibi, A.~Ghiasi, Z.~Xu, J.~Dickerson, C.~Studer, L.~S. Davis,
  G.~Taylor, and T.~Goldstein.
\newblock Adversarial training for free!
\newblock \emph{arXiv preprint arXiv:1904.12843}, 2019.

\bibitem[Shaham et~al.(2018)Shaham, Garritano, Yamada, Weinberger, Cloninger,
  Cheng, Stanton, and Kluger]{shaham2018defending}
U.~Shaham, J.~Garritano, Y.~Yamada, E.~Weinberger, A.~Cloninger, X.~Cheng,
  K.~Stanton, and Y.~Kluger.
\newblock Defending against adversarial images using basis functions
  transformations.
\newblock \emph{arXiv preprint arXiv:1803.10840}, 2018.

\bibitem[Shneiderman(2020)]{shneiderman2020bridging}
B.~Shneiderman.
\newblock Bridging the gap between ethics and practice: Guidelines for
  reliable, safe, and trustworthy human-centered ai systems.
\newblock \emph{ACM Transactions on Interactive Intelligent Systems (TiiS)},
  10\penalty0 (4):\penalty0 1--31, 2020.

\bibitem[Shrikumar et~al.(2017)Shrikumar, Greenside, and
  Kundaje]{shrikumar2017learning}
A.~Shrikumar, P.~Greenside, and A.~Kundaje.
\newblock Learning important features through propagating activation
  differences.
\newblock In \emph{Proceedings of the 34th International Conference on Machine
  Learning-Volume 70}, pages 3145--3153. JMLR. org, 2017.

\bibitem[Siddiqui et~al.(2020)Siddiqui, Valentin, and
  Nie{\ss}ner]{siddiqui2020viewal}
Y.~Siddiqui, J.~Valentin, and M.~Nie{\ss}ner.
\newblock Viewal: Active learning with viewpoint entropy for semantic
  segmentation.
\newblock In \emph{Proceedings of the IEEE/CVF Conference on Computer Vision
  and Pattern Recognition}, pages 9433--9443, 2020.

\bibitem[Siebert et~al.(2020)Siebert, Joeckel, Heidrich, Nakamichi, Ohashi,
  Namba, Yamamoto, and Aoyama]{siebert2020towards}
J.~Siebert, L.~Joeckel, J.~Heidrich, K.~Nakamichi, K.~Ohashi, I.~Namba,
  R.~Yamamoto, and M.~Aoyama.
\newblock Towards guidelines for assessing qualities of machine learning
  systems.
\newblock In \emph{QUATIC}, pages 17--31. Springer, 2020.

\bibitem[Simonyan et~al.(2013)Simonyan, Vedaldi, and
  Zisserman]{simonyan2013deep}
K.~Simonyan, A.~Vedaldi, and A.~Zisserman.
\newblock Deep inside convolutional networks: Visualising image classification
  models and saliency maps.
\newblock \emph{arXiv preprint arXiv:1312.6034}, 2013.

\bibitem[Singla and Feizi()]{singlafantastic}
S.~Singla and S.~Feizi.
\newblock Fantastic four: Differentiable bounds on singular values of
  convolution layers.

\bibitem[Sinha et~al.(2018)Sinha, Namkoong, and Duchi]{sinha2018certifying}
A.~Sinha, H.~Namkoong, and J.~Duchi.
\newblock Certifying some distributional robustness with principled adversarial
  training.
\newblock In \emph{ICLR}, 2018.

\bibitem[Smilkov et~al.(2017)Smilkov, Thorat, Kim, Vi{\'e}gas, and
  Wattenberg]{smilkov2017smoothgrad}
D.~Smilkov, N.~Thorat, B.~Kim, F.~Vi{\'e}gas, and M.~Wattenberg.
\newblock Smoothgrad: removing noise by adding noise.
\newblock \emph{arXiv preprint arXiv:1706.03825}, 2017.

\bibitem[Smuha(2019)]{smuha2019eu}
N.~A. Smuha.
\newblock The eu approach to ethics guidelines for trustworthy artificial
  intelligence.
\newblock \emph{CRi-Computer Law Review International}, 2019.

\bibitem[Sohn et~al.(2021)Sohn, Li, Yoon, Jin, and Pfister]{sohn2020learning}
K.~Sohn, C.-L. Li, J.~Yoon, M.~Jin, and T.~Pfister.
\newblock Learning and evaluating representations for deep one-class
  classification.
\newblock In \emph{International Conference on Learning Representations}, 2021.

\bibitem[Springenberg et~al.(2014)Springenberg, Dosovitskiy, Brox, and
  Riedmiller]{springenberg2014striving}
J.~T. Springenberg, A.~Dosovitskiy, T.~Brox, and M.~Riedmiller.
\newblock Striving for simplicity: The all convolutional net.
\newblock \emph{arXiv preprint arXiv:1412.6806}, 2014.

\bibitem[Srivastava et~al.(2014)Srivastava, Hinton, Krizhevsky, Sutskever, and
  Salakhutdinov]{srivastava2014dropout}
N.~Srivastava, G.~Hinton, A.~Krizhevsky, I.~Sutskever, and R.~Salakhutdinov.
\newblock Dropout: a simple way to prevent neural networks from overfitting.
\newblock \emph{Journal of Machine Learning Research}, 15\penalty0
  (1):\penalty0 1929--1958, 2014.

\bibitem[Strobelt et~al.(2018)Strobelt, Gehrmann, Pfister, and
  Rush]{strobelt2018lstmvis}
H.~Strobelt, S.~Gehrmann, H.~Pfister, and A.~M. Rush.
\newblock Lstmvis: A tool for visual analysis of hidden state dynamics in
  recurrent neural networks.
\newblock \emph{IEEE Transactions on Visualization and Computer Graphics},
  24\penalty0 (1):\penalty0 667--676, 2018.

\bibitem[Sun et~al.(2018)Sun, Huang, Kroening, Sharp, Hill, and
  Ashmore]{sun2018testing}
Y.~Sun, X.~Huang, D.~Kroening, J.~Sharp, M.~Hill, and R.~Ashmore.
\newblock Testing deep neural networks.
\newblock \emph{arXiv preprint arXiv:1803.04792}, 2018.

\bibitem[Szegedy et~al.(2016)Szegedy, Vanhoucke, Ioffe, Shlens, and
  Wojna]{szegedy2016rethinking}
C.~Szegedy, V.~Vanhoucke, S.~Ioffe, J.~Shlens, and Z.~Wojna.
\newblock Rethinking the inception architecture for computer vision.
\newblock In \emph{CVPR}, 2016.

\bibitem[Tack et~al.(2020)Tack, Mo, Jeong, and Shin]{tack2020csi}
J.~Tack, S.~Mo, J.~Jeong, and J.~Shin.
\newblock {CSI}: Novelty detection via contrastive learning on distributionally
  shifted instances.
\newblock In \emph{NeurIPS}, 2020.

\bibitem[Tang et~al.(2019)Tang, Naphade, Birchfield, Tremblay, Hodge, Kumar,
  Wang, and Yang]{tang2019pamtri}
Z.~Tang, M.~Naphade, S.~Birchfield, J.~Tremblay, W.~Hodge, R.~Kumar, S.~Wang,
  and X.~Yang.
\newblock Pamtri: Pose-aware multi-task learning for vehicle re-identification
  using highly randomized synthetic data.
\newblock In \emph{ICCV}, 2019.

\bibitem[Tavakoli et~al.(2021)Tavakoli, Agostinelli, and
  Baldi]{tavakoli2021splash}
M.~Tavakoli, F.~Agostinelli, and P.~Baldi.
\newblock {SPLASH}: Learnable activation functions for improving accuracy and
  adversarial robustness.
\newblock \emph{Neural Networks}, 140:\penalty0 1--12, 2021.

\bibitem[Techapanurak et~al.(2020)Techapanurak, Suganuma, and
  Okatani]{techapanurak2020hyperparameter}
E.~Techapanurak, M.~Suganuma, and T.~Okatani.
\newblock Hyperparameter-free out-of-distribution detection using cosine
  similarity.
\newblock In \emph{ACCV}, 2020.

\bibitem[Thulasidasan et~al.(2019)Thulasidasan, Chennupati, Bilmes,
  Bhattacharya, and Michalak]{thulasidasan2019mixup}
S.~Thulasidasan, G.~Chennupati, J.~A. Bilmes, T.~Bhattacharya, and S.~Michalak.
\newblock On {MixUp} training: Improved calibration and predictive uncertainty
  for deep neural networks.
\newblock In \emph{NeurIPS}, 2019.

\bibitem[Tobin et~al.(2017)Tobin, Fong, Ray, Schneider, Zaremba, and
  Abbeel]{tobin2017domain}
J.~Tobin, R.~Fong, A.~Ray, J.~Schneider, W.~Zaremba, and P.~Abbeel.
\newblock Domain randomization for transferring deep neural networks from
  simulation to the real world.
\newblock In \emph{IROS}, pages 23--30. IEEE, 2017.

\bibitem[Tramer et~al.(2020)Tramer, Carlini, Brendel, and
  Madry]{tramer2020adaptive}
F.~Tramer, N.~Carlini, W.~Brendel, and A.~Madry.
\newblock On adaptive attacks to adversarial example defenses.
\newblock \emph{Advances in Neural Information Processing Systems},
  33:\penalty0 1633--1645, 2020.

\bibitem[Tremblay et~al.(2018)Tremblay, Prakash, Acuna, Brophy, Jampani, Anil,
  To, Cameracci, Boochoon, and Birchfield]{tremblay2018training}
J.~Tremblay, A.~Prakash, D.~Acuna, M.~Brophy, V.~Jampani, C.~Anil, T.~To,
  E.~Cameracci, S.~Boochoon, and S.~Birchfield.
\newblock Training deep networks with synthetic data: Bridging the reality gap
  by domain randomization.
\newblock In \emph{Proceedings of the IEEE Conference on Computer Vision and
  Pattern Recognition Workshops}, pages 969--977, 2018.

\bibitem[Tsipras et~al.(2018)Tsipras, Santurkar, Engstrom, Turner, and
  Madry]{tsipras2018robustness}
D.~Tsipras, S.~Santurkar, L.~Engstrom, A.~Turner, and A.~Madry.
\newblock Robustness may be at odds with accuracy.
\newblock \emph{arXiv preprint arXiv:1805.12152}, 2018.

\bibitem[Tsipras et~al.(2020)Tsipras, Santurkar, Engstrom, Ilyas, and
  Madry]{tsipras2020imagenet}
D.~Tsipras, S.~Santurkar, L.~Engstrom, A.~Ilyas, and A.~Madry.
\newblock From imagenet to image classification: Contextualizing progress on
  benchmarks.
\newblock \emph{arXiv preprint arXiv:2005.11295}, 2020.

\bibitem[Van~Amersfoort et~al.(2020)Van~Amersfoort, Smith, Teh, and
  Gal]{van2020uncertainty}
J.~Van~Amersfoort, L.~Smith, Y.~W. Teh, and Y.~Gal.
\newblock Uncertainty estimation using a single deep deterministic neural
  network.
\newblock In \emph{International Conference on Machine Learning}, pages
  9690--9700. PMLR, 2020.

\bibitem[Van~Horn and Perona(2017)]{van2017devil}
G.~Van~Horn and P.~Perona.
\newblock The devil is in the tails: Fine-grained classification in the wild.
\newblock \emph{arXiv preprint arXiv:1709.01450}, 2017.

\bibitem[Varshney(2016)]{varshney2016engineering}
K.~R. Varshney.
\newblock Engineering safety in machine learning.
\newblock In \emph{Information Theory and Applications Workshop}, 2016.

\bibitem[Vasconcelos et~al.(2020)Vasconcelos, Larochelle, Dumoulin, Roux, and
  Goroshin]{vasconcelos2020effective}
C.~Vasconcelos, H.~Larochelle, V.~Dumoulin, N.~L. Roux, and R.~Goroshin.
\newblock An effective anti-aliasing approach for residual networks.
\newblock \emph{arXiv preprint arXiv:2011.10675}, 2020.

\bibitem[{Volpi} et~al.(2018){Volpi}, {Namkoong}, {Sener}, {Duchi}, {Murino},
  and {Savarese}]{volpi2018generalizing}
R.~{Volpi}, H.~{Namkoong}, O.~{Sener}, J.~C. {Duchi}, V.~{Murino}, and
  S.~{Savarese}.
\newblock Generalizing to unseen domains via adversarial data augmentation.
\newblock In \emph{Advances in Neural Information Processing Systems}, 2018.

\bibitem[Vyas et~al.(2018)Vyas, Jammalamadaka, Zhu, Das, Kaul, and
  Willke]{vyas2018out}
A.~Vyas, N.~Jammalamadaka, X.~Zhu, D.~Das, B.~Kaul, and T.~L. Willke.
\newblock Out-of-distribution detection using an ensemble of self supervised
  leave-out classifiers.
\newblock In \emph{ECCV}, pages 550--564, 2018.

\bibitem[Wang et~al.(2019{\natexlab{a}})Wang, Ge, Xing, and
  Lipton]{wang2019learning2}
H.~Wang, S.~Ge, E.~P. Xing, and Z.~C. Lipton.
\newblock Learning robust global representations by penalizing local predictive
  power.
\newblock 2019{\natexlab{a}}.

\bibitem[Wang et~al.(2019{\natexlab{b}})Wang, He, Lipton, and
  Xing]{wang2019learning}
H.~Wang, Z.~He, Z.~C. Lipton, and E.~P. Xing.
\newblock Learning robust representations by projecting superficial statistics
  out.
\newblock In \emph{ICLR}, 2019{\natexlab{b}}.

\bibitem[Wang et~al.(2020{\natexlab{a}})Wang, Chen, Gui, Hu, Liu, and
  Wang]{wang2020once}
H.~Wang, T.~Chen, S.~Gui, T.-K. Hu, J.~Liu, and Z.~Wang.
\newblock Once-for-all adversarial training: In-situ tradeoff between
  robustness and accuracy for free.
\newblock \emph{arXiv preprint arXiv:2010.11828}, 2020{\natexlab{a}}.

\bibitem[Wang et~al.(2020{\natexlab{b}})Wang, Chen, Wang, and
  Ma]{wang2020going}
H.~Wang, T.~Chen, Z.~Wang, and K.~Ma.
\newblock I am going mad: Maximum discrepancy competition for comparing
  classifiers adaptively.
\newblock In \emph{International Conference on Learning Representations},
  2020{\natexlab{b}}.

\bibitem[Wang et~al.(2020{\natexlab{c}})Wang, Wu, Huang, and
  Xing]{wang2020high}
H.~Wang, X.~Wu, Z.~Huang, and E.~P. Xing.
\newblock High-frequency component helps explain the generalization of
  convolutional neural networks.
\newblock In \emph{CVPR}, 2020{\natexlab{c}}.

\bibitem[Wang et~al.(2021)Wang, Xiao, Kossaifi, Yu, Anandkumar, and
  Wang]{wang2021augmax}
H.~Wang, C.~Xiao, J.~Kossaifi, Z.~Yu, A.~Anandkumar, and Z.~Wang.
\newblock {AugMax}: Adversarial composition of random augmentations for robust
  training.
\newblock In \emph{NeurIPS}, 2021.

\bibitem[Wang et~al.(2018{\natexlab{a}})Wang, Chen, Abdou, and
  Jana]{wang2018mixtrain}
S.~Wang, Y.~Chen, A.~Abdou, and S.~Jana.
\newblock Mixtrain: Scalable training of formally robust neural networks.
\newblock \emph{arXiv preprint arXiv:1811.02625}, 2018{\natexlab{a}}.

\bibitem[Wang et~al.(2018{\natexlab{b}})Wang, Pei, Whitehouse, Yang, and
  Jana]{wang2018efficient}
S.~Wang, K.~Pei, J.~Whitehouse, J.~Yang, and S.~Jana.
\newblock Efficient formal safety analysis of neural networks.
\newblock In \emph{Advances in Neural Information Processing Systems}, pages
  6369--6379, 2018{\natexlab{b}}.

\bibitem[Wang et~al.(2020{\natexlab{d}})Wang, Dai, Wipf, and
  Zhu]{wang2020further}
Z.~Wang, B.~Dai, D.~Wipf, and J.~Zhu.
\newblock Further analysis of outlier detection with deep generative models.
\newblock \emph{arXiv preprint arXiv:2010.13064}, 2020{\natexlab{d}}.

\bibitem[Wei et~al.(2020)Wei, Feng, Chen, and An]{wei2020combating}
H.~Wei, L.~Feng, X.~Chen, and B.~An.
\newblock Combating noisy labels by agreement: A joint training method with
  co-regularization.
\newblock In \emph{CVPR}, pages 13726--13735, 2020.

\bibitem[Wexler(2017)]{wexler2017facets}
J.~Wexler.
\newblock Facets: An open source visualization tool for machine learning
  training data.
\newblock \emph{Google Open Source Blog}, 2017.

\bibitem[Winkens et~al.(2020)Winkens, Bunel, Roy, Stanforth, Natarajan, Ledsam,
  MacWilliams, Kohli, Karthikesalingam, Kohl, et~al.]{winkens2020contrastive}
J.~Winkens, R.~Bunel, A.~G. Roy, R.~Stanforth, V.~Natarajan, J.~R. Ledsam,
  P.~MacWilliams, P.~Kohli, A.~Karthikesalingam, S.~Kohl, et~al.
\newblock Contrastive training for improved out-of-distribution detection.
\newblock \emph{arXiv preprint arXiv:2007.05566}, 2020.

\bibitem[Wong and Kolter(2018)]{wong2018provable}
E.~Wong and Z.~Kolter.
\newblock Provable defenses against adversarial examples via the convex outer
  adversarial polytope.
\newblock In \emph{International Conference on Machine Learning}, pages
  5283--5292, 2018.

\bibitem[Wong et~al.(2020)Wong, Rice, and Kolter]{wong2020fast}
E.~Wong, L.~Rice, and J.~Z. Kolter.
\newblock Fast is better than free: Revisiting adversarial training.
\newblock \emph{arXiv:2001.03994}, 2020.

\bibitem[Wongsuphasawat et~al.(2017)Wongsuphasawat, Smilkov, Wexler, Wilson,
  Mane, Fritz, Krishnan, Vi{\'e}gas, and
  Wattenberg]{wongsuphasawat2017visualizing}
K.~Wongsuphasawat, D.~Smilkov, J.~Wexler, J.~Wilson, D.~Mane, D.~Fritz,
  D.~Krishnan, F.~B. Vi{\'e}gas, and M.~Wattenberg.
\newblock Visualizing dataflow graphs of deep learning models in tensorflow.
\newblock \emph{TVCG}, 2017.

\bibitem[Wu et~al.(2020)Wu, Chen, Cai, He, and Gu]{wu2020does}
B.~Wu, J.~Chen, D.~Cai, X.~He, and Q.~Gu.
\newblock Does network width really help adversarial robustness?
\newblock \emph{arXiv preprint arXiv:2010.01279}, 2020.

\bibitem[Wu et~al.(2018)Wu, Hughes, Parbhoo, Zazzi, Roth, and
  Doshi-Velez]{wu2018beyond}
M.~Wu, M.~C. Hughes, S.~Parbhoo, M.~Zazzi, V.~Roth, and F.~Doshi-Velez.
\newblock Beyond sparsity: Tree regularization of deep models for
  interpretability.
\newblock In \emph{Thirty-Second AAAI Conference on Artificial Intelligence},
  2018.

\bibitem[Wu et~al.(2021)Wu, Guo, Su, and Weinberger]{wu2021online}
R.~Wu, C.~Guo, Y.~Su, and K.~Q. Weinberger.
\newblock Online adaptation to label distribution shift.
\newblock In \emph{Advances in Neural Information Processing Systems}, 2021.

\bibitem[Wu et~al.(2019)Wu, Suresh, Narayanan, Xu, Kwon, and
  Wang]{wu2019delving}
Z.~Wu, K.~Suresh, P.~Narayanan, H.~Xu, H.~Kwon, and Z.~Wang.
\newblock Delving into robust object detection from unmanned aerial vehicles: A
  deep nuisance disentanglement approach.
\newblock In \emph{ICCV}, 2019.

\bibitem[Xiao et~al.(2018{\natexlab{a}})Xiao, Deng, Li, Yu, Liu, and
  Song]{xiao2018characterizing}
C.~Xiao, R.~Deng, B.~Li, F.~Yu, M.~Liu, and D.~Song.
\newblock Characterizing adversarial examples based on spatial consistency
  information for semantic segmentation.
\newblock In \emph{Proceedings of the European Conference on Computer Vision
  (ECCV)}, 2018{\natexlab{a}}.

\bibitem[Xiao et~al.(2018{\natexlab{b}})Xiao, Zhu, Li, He, Liu, and
  Song]{xiao2018spatially}
C.~Xiao, J.-Y. Zhu, B.~Li, W.~He, M.~Liu, and D.~Song.
\newblock Spatially transformed adversarial examples.
\newblock \emph{arXiv preprint arXiv:1801.02612}, 2018{\natexlab{b}}.

\bibitem[Xiao et~al.(2019{\natexlab{a}})Xiao, Yang, Li, Deng, and
  Liu]{xiao2019meshadv}
C.~Xiao, D.~Yang, B.~Li, J.~Deng, and M.~Liu.
\newblock Meshadv: Adversarial meshes for visual recognition.
\newblock In \emph{Proceedings of the IEEE Conference on Computer Vision and
  Pattern Recognition}, pages 6898--6907, 2019{\natexlab{a}}.

\bibitem[Xiao et~al.(2019{\natexlab{b}})Xiao, Tjeng, Shafiullah, and
  Madry]{xiao2018training}
K.~Y. Xiao, V.~Tjeng, N.~M. Shafiullah, and A.~Madry.
\newblock Training for faster adversarial robustness verification via inducing
  {ReLU} stability.
\newblock \emph{ICLR}, 2019{\natexlab{b}}.

\bibitem[Xie and Yuille(2019)]{xie2019intriguing}
C.~Xie and A.~Yuille.
\newblock Intriguing properties of adversarial training at scale.
\newblock \emph{arXiv preprint arXiv:1906.03787}, 2019.

\bibitem[Xie et~al.(2018)Xie, Wang, Zhang, Ren, and Yuille]{xie2017mitigating}
C.~Xie, J.~Wang, Z.~Zhang, Z.~Ren, and A.~Yuille.
\newblock Mitigating adversarial effects through randomization.
\newblock In \emph{ICLR}, 2018.

\bibitem[Xie et~al.(2019)Xie, Wu, Maaten, Yuille, and He]{xie2019feature}
C.~Xie, Y.~Wu, L.~v.~d. Maaten, A.~L. Yuille, and K.~He.
\newblock Feature denoising for improving adversarial robustness.
\newblock In \emph{Proceedings of the IEEE/CVF Conference on Computer Vision
  and Pattern Recognition}, pages 501--509, 2019.

\bibitem[Xie et~al.(2020{\natexlab{a}})Xie, Tan, Gong, Wang, Yuille, and
  Le]{xie2020adversarial}
C.~Xie, M.~Tan, B.~Gong, J.~Wang, A.~L. Yuille, and Q.~V. Le.
\newblock Adversarial examples improve image recognition.
\newblock In \emph{CVPR}, pages 819--828, 2020{\natexlab{a}}.

\bibitem[Xie et~al.(2020{\natexlab{b}})Xie, Tan, Gong, Yuille, and
  Le]{xie2020smooth}
C.~Xie, M.~Tan, B.~Gong, A.~Yuille, and Q.~V. Le.
\newblock Smooth adversarial training.
\newblock \emph{arXiv preprint arXiv:2006.14536}, 2020{\natexlab{b}}.

\bibitem[Xu et~al.(2017)Xu, Evans, and Qi]{xu2017feature}
W.~Xu, D.~Evans, and Y.~Qi.
\newblock Feature squeezing: Detecting adversarial examples in deep neural
  networks.
\newblock \emph{arXiv preprint arXiv:1704.01155}, 2017.

\bibitem[Xu et~al.(2021)Xu, Liu, Yang, and Niethammer]{xu2021robust}
Z.~Xu, D.~Liu, J.~Yang, and M.~Niethammer.
\newblock Robust and generalizable visual representation learning via random
  convolutions.
\newblock 2021.

\bibitem[Yamaguchi et~al.(2016)Yamaguchi, Kaga, Donz{\'e}, and
  Seshia]{yamaguchi2016combining}
T.~Yamaguchi, T.~Kaga, A.~Donz{\'e}, and S.~A. Seshia.
\newblock Combining requirement mining, software model checking and
  simulation-based verification for industrial automotive systems.
\newblock In \emph{FMCAD}, pages 201--204. IEEE, 2016.

\bibitem[Yang et~al.(2021)Yang, Wang, Feng, Yan, Zheng, Zhang, and
  Liu]{yang2021semantically}
J.~Yang, H.~Wang, L.~Feng, X.~Yan, H.~Zheng, W.~Zhang, and Z.~Liu.
\newblock Semantically coherent out-of-distribution detection.
\newblock In \emph{ICCV}, pages 8301--8309, 2021.

\bibitem[Ye et~al.(2019)Ye, Xu, Liu, Cheng, Lambrechts, Zhang, Zhou, Ma, Wang,
  and Lin]{ye2019adversarial}
S.~Ye, K.~Xu, S.~Liu, H.~Cheng, J.-H. Lambrechts, H.~Zhang, A.~Zhou, K.~Ma,
  Y.~Wang, and X.~Lin.
\newblock Adversarial robustness vs. model compression, or both?
\newblock In \emph{ICCV}, 2019.

\bibitem[Yeh et~al.(2020)Yeh, Kim, Arik, Li, Pfister, and
  Ravikumar]{yeh2020completeness}
C.-K. Yeh, B.~Kim, S.~Arik, C.-L. Li, T.~Pfister, and P.~Ravikumar.
\newblock On completeness-aware concept-based explanations in deep neural
  networks.
\newblock \emph{Advances in Neural Information Processing Systems}, 33, 2020.

\bibitem[Yosinski et~al.(2014)Yosinski, Clune, Bengio, and
  Lipson]{yosinski2014transferable}
J.~Yosinski, J.~Clune, Y.~Bengio, and H.~Lipson.
\newblock How transferable are features in deep neural networks?
\newblock In \emph{Advances in neural information processing systems}, pages
  3320--3328, 2014.

\bibitem[You et~al.(2020)You, Chen, Wang, and Shen]{you2020does}
Y.~You, T.~Chen, Z.~Wang, and Y.~Shen.
\newblock When does self-supervision help graph convolutional networks?
\newblock In \emph{International Conference on Machine Learning}, pages
  10871--10880. PMLR, 2020.

\bibitem[Yu and Aizawa(2019)]{yu2019unsupervised}
Q.~Yu and K.~Aizawa.
\newblock Unsupervised out-of-distribution detection by maximum classifier
  discrepancy.
\newblock In \emph{ICCV}, pages 9518--9526, 2019.

\bibitem[Yu et~al.(2019)Yu, Han, Yao, Niu, Tsang, and Sugiyama]{yu2019does}
X.~Yu, B.~Han, J.~Yao, G.~Niu, I.~Tsang, and M.~Sugiyama.
\newblock How does disagreement help generalization against label corruption?
\newblock In \emph{ICML}, pages 7164--7173, 2019.

\bibitem[Yuan et~al.(2020)Yuan, Tay, Li, Wang, and Feng]{yuan2020revisiting}
L.~Yuan, F.~E. Tay, G.~Li, T.~Wang, and J.~Feng.
\newblock Revisiting knowledge distillation via label smoothing regularization.
\newblock In \emph{Proceedings of the IEEE/CVF Conference on Computer Vision
  and Pattern Recognition}, pages 3903--3911, 2020.

\bibitem[Yue et~al.(2019)Yue, Zhang, Zhao, Sangiovanni-Vincentelli, Keutzer,
  and Gong]{yue2019domain}
X.~Yue, Y.~Zhang, S.~Zhao, A.~Sangiovanni-Vincentelli, K.~Keutzer, and B.~Gong.
\newblock Domain randomization and pyramid consistency: Simulation-to-real
  generalization without accessing target domain data.
\newblock In \emph{ICCV}, pages 2100--2110, 2019.

\bibitem[Yun et~al.(2019)Yun, Han, Oh, Chun, Choe, and Yoo]{yun2019cutmix}
S.~Yun, D.~Han, S.~J. Oh, S.~Chun, J.~Choe, and Y.~Yoo.
\newblock Cutmix: Regularization strategy to train strong classifiers with
  localizable features.
\newblock In \emph{Proceedings of the IEEE International Conference on Computer
  Vision}, pages 6023--6032, 2019.

\bibitem[Yun et~al.(2021)Yun, Oh, Heo, Han, Choe, and Chun]{yun2021re}
S.~Yun, S.~J. Oh, B.~Heo, D.~Han, J.~Choe, and S.~Chun.
\newblock Re-labeling {ImageNet}: From single to multi-labels, from global to
  localized labels.
\newblock In \emph{CVPR}, 2021.

\bibitem[Zeiler and Fergus(2014)]{zeiler2014visualizing}
M.~D. Zeiler and R.~Fergus.
\newblock Visualizing and understanding convolutional networks.
\newblock In \emph{ECCV}, 2014.

\bibitem[Zhang et~al.(2019{\natexlab{a}})Zhang, Zhang, Lu, Zhu, and
  Dong]{zhang2019you}
D.~Zhang, T.~Zhang, Y.~Lu, Z.~Zhu, and B.~Dong.
\newblock You only propagate once: Accelerating adversarial training via
  maximal principle.
\newblock \emph{arXiv preprint arXiv:1905.00877}, 2019{\natexlab{a}}.

\bibitem[Zhang et~al.(2017)Zhang, Cisse, Dauphin, and
  Lopez-Paz]{zhang2017mixup}
H.~Zhang, M.~Cisse, Y.~N. Dauphin, and D.~Lopez-Paz.
\newblock {MixUp}: Beyond empirical risk minimization.
\newblock \emph{arXiv preprint arXiv:1710.09412}, 2017.

\bibitem[Zhang et~al.(2019{\natexlab{b}})Zhang, Chen, Xiao, Gowal, Stanforth,
  Li, Boning, and Hsieh]{zhang2019towards}
H.~Zhang, H.~Chen, C.~Xiao, S.~Gowal, R.~Stanforth, B.~Li, D.~Boning, and C.-J.
  Hsieh.
\newblock Towards stable and efficient training of verifiably robust neural
  networks.
\newblock \emph{arXiv preprint arXiv:1906.06316}, 2019{\natexlab{b}}.

\bibitem[Zhang et~al.(2019{\natexlab{c}})Zhang, Yu, Jiao, Xing, Ghaoui, and
  Jordan]{zhang2019theoretically}
H.~Zhang, Y.~Yu, J.~Jiao, E.~Xing, L.~E. Ghaoui, and M.~Jordan.
\newblock Theoretically principled trade-off between robustness and accuracy.
\newblock In \emph{Proceedings of the 36th International Conference on Machine
  Learning}, 2019{\natexlab{c}}.

\bibitem[Zhang et~al.(2020{\natexlab{a}})Zhang, Zhu, Niu, Han, Sugiyama, and
  Kankanhalli]{zhang2020geometry}
J.~Zhang, J.~Zhu, G.~Niu, B.~Han, M.~Sugiyama, and M.~Kankanhalli.
\newblock Geometry-aware instance-reweighted adversarial training.
\newblock \emph{arXiv preprint arXiv:2010.01736}, 2020{\natexlab{a}}.

\bibitem[Zhang et~al.(2021{\natexlab{a}})Zhang, Inkawhich, Chen, and
  Li]{zhang2021fine}
J.~Zhang, N.~Inkawhich, Y.~Chen, and H.~Li.
\newblock Fine-grained out-of-distribution detection with mixup outlier
  exposure.
\newblock \emph{arXiv preprint arXiv:2106.03917}, 2021{\natexlab{a}}.

\bibitem[Zhang et~al.(2020{\natexlab{b}})Zhang, Harman, Ma, and
  Liu]{zhang2020machine}
J.~M. Zhang, M.~Harman, L.~Ma, and Y.~Liu.
\newblock Machine learning testing: Survey, landscapes and horizons.
\newblock \emph{IEEE Transactions on Software Engineering}, 2020{\natexlab{b}}.

\bibitem[Zhang et~al.(2018)Zhang, Wang, and Zhu]{zhang2018examining}
Q.~Zhang, W.~Wang, and S.-C. Zhu.
\newblock Examining cnn representations with respect to dataset bias.
\newblock In \emph{Thirty-Second AAAI Conference on Artificial Intelligence},
  2018.

\bibitem[Zhang(2019)]{zhang2019making}
R.~Zhang.
\newblock Making convolutional networks shift-invariant again.
\newblock In \emph{International Conference on Machine Learning}, pages
  7324--7334. PMLR, 2019.

\bibitem[Zhang and LeCun(2017)]{zhang2017universum}
X.~Zhang and Y.~LeCun.
\newblock Universum prescription: Regularization using unlabeled data.
\newblock In \emph{AAAI}, 2017.

\bibitem[Zhang et~al.(2019{\natexlab{d}})Zhang, Wang, Liu, and
  Ling]{zhang2019dada}
X.~Zhang, Z.~Wang, D.~Liu, and Q.~Ling.
\newblock Dada: Deep adversarial data augmentation for extremely low data
  regime classification.
\newblock In \emph{ICASSP}. IEEE, 2019{\natexlab{d}}.

\bibitem[Zhang et~al.(2021{\natexlab{b}})Zhang, Ling, Gao, Yin, Lafleche,
  Barriuso, Torralba, and Fidler]{zhang2021datasetgan}
Y.~Zhang, H.~Ling, J.~Gao, K.~Yin, J.-F. Lafleche, A.~Barriuso, A.~Torralba,
  and S.~Fidler.
\newblock Datasetgan: Efficient labeled data factory with minimal human effort.
\newblock In \emph{Proceedings of the IEEE/CVF Conference on Computer Vision
  and Pattern Recognition}, pages 10145--10155, 2021{\natexlab{b}}.

\bibitem[Zhang et~al.(2019{\natexlab{e}})Zhang, Dalca, and
  Sabuncu]{zhang2019confidence}
Z.~Zhang, A.~V. Dalca, and M.~R. Sabuncu.
\newblock Confidence calibration for convolutional neural networks using
  structured dropout.
\newblock \emph{arXiv preprint arXiv:1906.09551}, 2019{\natexlab{e}}.

\bibitem[{Zhao} et~al.(2020){Zhao}, {Liu}, {Peng}, and
  {Metaxas}]{zhao2020maximum}
L.~{Zhao}, T.~{Liu}, X.~{Peng}, and D.~{Metaxas}.
\newblock Maximum-entropy adversarial data augmentation for improved
  generalization and robustness.
\newblock In \emph{Advances in Neural Information Processing Systems},
  volume~33, 2020.

\bibitem[{Zheng} et~al.(2016){Zheng}, {Song}, {Leung}, and
  {Goodfellow}]{zheng2016improving}
S.~{Zheng}, Y.~{Song}, T.~{Leung}, and I.~{Goodfellow}.
\newblock Improving the robustness of deep neural networks via stability
  training.
\newblock In \emph{CVPR}, 2016.

\bibitem[Zhong et~al.(2020)Zhong, Zheng, Kang, Li, and Yang]{zhong2017random}
Z.~Zhong, L.~Zheng, G.~Kang, S.~Li, and Y.~Yang.
\newblock Random erasing data augmentation.
\newblock In \emph{Proceedings of the AAAI conference on artificial
  intelligence}, 2020.

\bibitem[Zhou et~al.(2018)Zhou, Sun, Bau, and Torralba]{zhou2018interpretable}
B.~Zhou, Y.~Sun, D.~Bau, and A.~Torralba.
\newblock Interpretable basis decomposition for visual explanation.
\newblock In \emph{ECCV}, 2018.

\bibitem[Zong et~al.(2018)Zong, Song, Min, Cheng, Lumezanu, Cho, and
  Chen]{zong2018deep}
B.~Zong, Q.~Song, M.~R. Min, W.~Cheng, C.~Lumezanu, D.~Cho, and H.~Chen.
\newblock Deep autoencoding gaussian mixture model for unsupervised anomaly
  detection.
\newblock In \emph{International Conference on Learning Representations}, 2018.

\end{thebibliography}
}
% % \bibliographystyle{spbasic}      % basic style, author-year citations
% \bibliographystyle{spmpsci}      % mathematics and physical sciences
% %\bibliographystyle{spphys}       % APS-like style for physics
% %\bibliography{}   % name your BibTeX data base
% \bibliography{Bibliography}

% \begin{thebibliography}{}
% %
% % and use \bibitem to create references. Consult the Instructions
% % for authors for reference list style.
% % Format for Journal Reference
% Author, Article title, Journal, Volume, page numbers (year)
% Author, Book title, page numbers. Publisher, place (year)
% \end{thebibliography}

\end{document}